\documentclass[sigconf,nonacm]{acmart}

\usepackage{libertine}
\usepackage{url}
\usepackage{algorithm}  
\usepackage{algorithmic}  

\usepackage{amssymb}
\usepackage{amsmath}
\usepackage{subfigure}
\usepackage{graphicx}
\usepackage{threeparttable}
\usepackage{multirow}

\AtBeginDocument{%
  \providecommand\BibTeX{{%
    \normalfont B\kern-0.5em{\scshape i\kern-0.25em b}\kern-0.8em\TeX}}}

\begin{document}

\title{Taylor Genetic Programming for Symbolic Regression}
\makeatletter
\patchcmd{\ps@pprintTitle}
  {Preprint submitted}
  {To be submitted}
  {}{}
\makeatother

\author{Baihe He}
\orcid{0000-0003-1060-2673}
\affiliation{%
  \institution{Beijing Key Laboratory of Petroleum Data Mining, China University of Petroleum}
  \city{Beijing} 
  \country{China} 
  \postcode{102249}
}
\email{hebaihe@hotmail.com}

\author{Qiang Lu}
\authornote{Corresponding author.}
\orcid{0000-0001-8217-2305}
\affiliation{%
  \institution{Beijing Key Laboratory of Petroleum Data Mining, China University of Petroleum}
  \city{Beijing} 
  \country{China} 
  \postcode{102249}
}
\email{luqiang@cup.edu.cn}

\author{Qingyun Yang}
\affiliation{%
  \institution{Beijing Key Laboratory of Petroleum Data Mining, China University of Petroleum}
  \city{Beijing} 
  \country{China} 
  \postcode{102249}
}
\email{yangqingyun.cup@hotmail.com}

\author{Jake Luo}
\orcid{0000-0002-3900-643X}
\affiliation{%
  \institution{Department of Health Informatics and Administration, University of Wisconsin Milwaukee}
  \city{Milwaukee} 
  \country{United States}
 }
\email{jakeluo@uwm.edu}

\author{Zhiguang Wang}
\affiliation{%
  \institution{Beijing Key Laboratory of Petroleum Data Mining, China University of Petroleum}
  \city{Beijing} 
  \country{China} 
  \postcode{102249}
}
\email{cwangzg@cup.edu.cn}


\begin{abstract}
Genetic programming (GP) is a commonly used approach to solve symbolic regression (SR) problems. Compared with the machine learning or deep learning methods that depend on the pre-defined model and the training dataset for solving SR problems, GP is more focused on finding the solution in a search space. Although GP has good performance on large-scale benchmarks, it randomly transforms individuals to search results without taking advantage of the characteristics of the dataset. 
So, the search process of GP is usually slow, and the final results could be unstable. 
To guide GP by these characteristics, we propose a new method for SR, called Taylor genetic programming (TaylorGP)\footnote{Code and appendix at https://kgae-cup.github.io/TaylorGP/}. TaylorGP leverages a Taylor polynomial to approximate the symbolic equation that fits the dataset. It also utilizes the Taylor polynomial to extract the features of the symbolic equation: low order polynomial discrimination, variable separability, boundary, monotonic, and parity. GP is enhanced by these Taylor polynomial techniques. Experiments are conducted on three kinds of benchmarks: classical SR, machine learning, and physics. The experimental results show that TaylorGP not only has higher accuracy than the nine baseline methods, but also is faster in finding stable results.
\end{abstract}




\maketitle

\section{Introduction}
Symbolic regression (\textbf{SR}) refers to finding a symbolic equation $f_\theta$ fitted to a given dataset $(X,Y)$ from the mathematical expression space, i.e., $f_\theta(X)=Y$. The mathematical expression space is huge, even for rather simple symbolic equations. For example, if a symbolic equation is represented by a binary tree with a maximum depth of 4, with 20 variables ($x_1,x_2,...,x_{20}$) and 18 basic functions (such as $+$, $-$ and sqrt), the size of the space is $8.2\times 10^{162}$ \cite{korns2013baseline}. Therefore, finding a good symbolic equation for the data from the huge possible search space is a challenging task.  

Research communities of evolutionary computation (\textbf{EC}) and machine learning (\textbf{ML}) have been trying to solve the SR problems from their perspectives. The EC methods, especially genetic programming (\textbf{GP}) methods  \cite{koza_genetic_1994,schmidt_distilling_2009,miller_cartesian_2019,ferreira2001gene}, are designed to search the mathematical expression space by evolving the encoding of each individual in a population. 
The main advantage of the GP approach is that the algorithm components are generalizable and adaptive, including processes such as selection, crossover, mutation, and fitness evaluation. Using these components, the GP algorithm randomly searches for a model $f_\theta$ that fits the dataset in the mathematical expression space, unlike machine learning methods (e.g., neural networks) that try to find and optimize a set of parameters $\theta$ under known models $f$. However, the GP search process usually does not consider the features of the given dataset, which could be an opportunity for improvement.
The ML (including neural networks) methods  \cite{mcconaghy2011ffx,kusner2017grammar,kim2020integration,petersen2021deep,biggio2021neural,udrescu2020ai_a} find parameters $\theta$ in pre-defined models $f$ so that the models fit the dataset, i.e., $f_\theta(X)=Y$. Machine learning methods heavily utilize the features of a dataset to guide the search and optimization for parameters. Therefore, the parameter search process is effective. However, when solving SR problems, though the machine learning approaches could quickly recover some correct symbolic equations $f_\theta$s , the results are commonly biased by the pre-defined regression model $f$s as well as the training dataset. For example, in a recent evaluation \cite{cava2021contemporary} of algorithms for SR problems using large-scale benchmarks, the results show that the top three approaches are still GP-based approaches, and GP approaches still have a substantial advantage over machine learning-based approaches.

\begin{figure}[h]
  \centering
  \includegraphics[width=\linewidth]{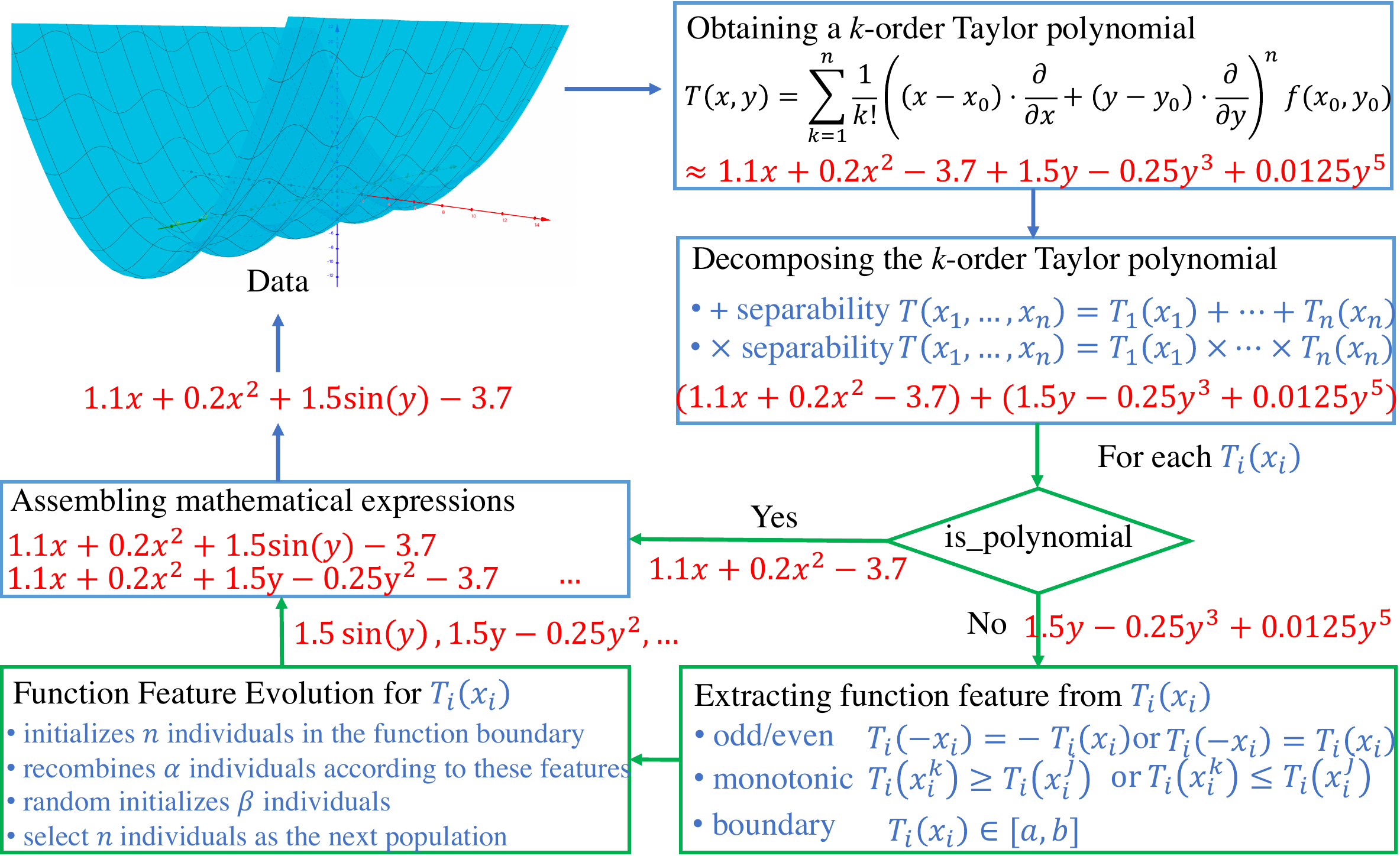}
  \caption{Taylor genetic programming.}
  \label{fig:taylorGP}
\end{figure}

In this paper, we propose a new GP approach called the Taylor Genetic Programming (TaylorGP). To overcome the common GP's drawback -- a search process without considering the data features, we embed Taylor features into GP and leverage the features to guide GP to search for solutions. 
TaylorGP, as shown in Figure \ref{fig:taylorGP}, first obtains a $k$-order Taylor polynomial at a point from the given dataset. According to the Taylor's theorem, the $k$-order Taylor polynomial approximates a smooth function that fits the given dataset near the point. Moreover, the $k$-order Taylor polynomial can show the local features of the smooth function (called \textbf{Taylor features}). Taylor features include three key components: variable separability, low order polynomial discrimination, and function feature. Using the variable separability, TaylorGP can decompose the multivariate Taylor polynomial $T$ into multiple univariate Taylor polynomials $T_is$. 
For example, the two-variable Taylor polynomial "$1.1x+0.2x^2-3.7+1.5y-0.25y^3+0.0125y^5$" can be decomposed into two univariate Taylor polynomials "$1.1x+0.2x^2-3.7$" and "$1.5y-0.25y^3+0.0125y^5$" according the variable separability. TaylorGP then applies the polynomial discrimination to determine whether each $T_i$ is a polynomial. If $T_i$ is not a polynomial, TaylorGP runs the function feature evolution method to find symbolic equations to fit $T_i$. The function feature evolution method creates a set of rules based on the function features to recombine $\lambda$ individuals. The method also employs the individual initialization method to randomly generate $\beta$ individuals to prevent premature convergence. TaylorGP finally assembles mathematical expressions of all $T_i$s to generate the final symbolic equation.

As the Taylor features are directly induced from the coefficients in the $k$-order Taylor polynomial, obtaining the features are simple and quick. For example, the coefficient of each two-variable product term "$x^my^n$" in "$1.1x+0.2x^2-3.7+1.5y-0.25y^3+0.0125y^5$" is zero, meaning the Taylor polynomial is variable separable. So, embedding these features into GP does not increase the GP's computing time complexity. Moreover, the search process guided by the Taylor features enables TaylorGP to find a correct symbolic equation quicker than without using the features.

The main contributions of this paper are the following:

(1) We propose a simple yet powerful Taylor genetic programming (TaylorGP) method for symbolic regression. TaylorGP combines the general characteristics of GP's solution search strategies (e.g., mutation, crossover) with the ML (NN)'s feature-directed search.

(2) We design a new Taylor feature extraction method. Using a Taylor polynomial obtained from a dataset, the method can map the dataset into features that can represent the properties of a target symbolic equation. Moreover, we create a function feature evolution method to transform individuals according to these features.

(3) We demonstrate that TaylorGP significantly outperforms state-of-the-art approaches, such as FFX\cite{mcconaghy2011ffx}, GSGP \cite{hutchison_geometric_2012}, BSR \cite{jin2019bayesian}, SVM, and XGBoost, on the three types of benchmarks: classical SR \cite{mcdermott12benchmark}, AIFeynman \cite{udrescu2020ai_a}, and Penn machine learning benchmarks \cite{Olson2017PMLB}. 


The remainder of this paper is organized as follows. In Section 2, we detail related work. Section 3 presents the Taylor features. Then, we propose the Taylor genetic programming in Section 4. Section 5 and 6 report the experimental results and discussion. Finally, we conclude the paper in Section 6.

\section{Related Work}
\subsection{Machine learning for symbolic regression}
The regression analysis in machine learning, such as linear regression \cite{weisberg2005applied}, SVM \cite{cortes1995support}, XGBoost \cite{chen2016xgboost}, and neural network(NN) \cite{lecun2015deep}, can be viewed as a special case of SR. Different from SR that needs to find both the model and its parameter values, machine learning aims to find the values ($\theta$) of parameters in a pre-defined model $f$, so that $f_{\theta}(X)=Y$. Many ML methods, such as deep neural networks, usually applies the gradient descent method to obtain the target parameter set $\theta$. Neural networks \cite{kusner2017grammar,kingma2013auto,petersen2021deep,biggio2021neural,sahoo2018learning} are trained to learn from a dataset $(X,Y)$ to generate a mathematical equation $f(X)=Y$ according to the features of the training dataset $\{(X,Y),\mathcal{F}\}$. 
For example, GrammarVAE (GVAE) \cite{kusner2017grammar} trains a variational autoencoder \cite{kingma2013auto} to directly encode from and decode to the parse trees using a context-free grammar.
Recently, DSR \cite{petersen2021deep} employed a recurrent neural network trained by a reinforcement learning algorithm (RL) for SR. The algorithm uses a risk-seeking policy gradient to emit a distribution over tractable mathematical expressions. According to the distribution, DSR samples the mathematical expressions with constant placeholders and obtains these constants with the nonlinear optimization algorithm -- BFGS \cite{fletcher2013practical}. 
Inspired by the recent successes of pre-trained models on large datasets, such as BERT \cite{devlin-etal-2019-bert} and GPT \cite{brown2020language}, NeSymReS \cite{biggio2021neural} pre-trains a model called Set Transformer on hundreds of millions of equations to generate a distribution of mathematical expressions according to the given dataset $(X,Y)$. In addition, NeSymReS samples mathematical expressions with constant placeholders by the beam search on the distributions and uses BFGS to optimize these constants.

Besides the above neural networks for SR, EQL \cite{martius2016extrapolation,sahoo2018learning} designs a shallow fully-connected neural network where standard activation functions (e.g., $tanh$, ReLU) are replaced with basic functions (e.g., "$+$", "$\times$, "$sin$). Once the neural network is trained, it can represent a symbolic equation fitted to the given dataset. AIFeynman \cite{udrescu2020ai_a,udrescu2020ai_b} employs neural networks to map the given dataset into simplifying properties (e.g., symmetries, separability, and compositionality). The method then uses a brute-force recursive algorithm guided by these simplifying properties and a suite of physics-inspired techniques to search possible symbolic expressions. 

The above machine learning methods, especially deep learning methods, have an excellent ability to discover mathematical equations on some specific benchmarks. However, experiments on the large-scale benchmarks \cite{cava2021contemporary} indicate that the mathematical equations found by ML or DL are less accurate than those found by GP-based methods. For example, four of the top five methods and six of the top ten methods are GP-based methods, and the other top methods are ensemble tree-based methods, such as XGBoost and LightGBM \cite{ke2017lightgbm}. Surprisingly, the top methods do not include the two neural network methods, DSR and AIFeynman. Furthermore, the neural network methods seem to be more dependent on the training dataset. Neural networks could not discover a correct mathematical equation if their structures (layer by layer) could not extract valid features from the dataset.

\subsection{Genetic programming for symbolic regression}
GP \cite{koza_genetic_1994} is still a commonly used approach to deal with SR.
GP uses evolutionary operators -- crossover, mutation, and selection, to change the individual encoding and generate better offspring for searching a solution in the mathematical expression space.
Various GPs employ different individual encodings to represent mathematical equations, such as tree-encoded GPs \cite{koza_genetic_1994,mckay_grammar-based_2010,hutchison_geometric_2012, trujillo2016neat, chen_improving_2019}, graph-encoded GPs \cite{CGP2008Miller,schmidt_distilling_2009}, and linearly encoded GPs \cite{ferreira2001gene,brameier2007linear,SPJ-GEP2021Lu}. 

For the mathematical expression space, the presence of real constants accounts for a significant portion of the size of the space. For example, the size of the aforementioned problem in the Introduction section is $8.2\times 10^{162}$. In comparison, without real constants, its size is $1.054\times 10^{19}$ \cite{korns2013baseline}. So, some GP methods \cite{korns2011accuracy,worm2013prioritized,krawiec2002genetic} with a constant optimizer are proposed to search the space. These methods represent the skeleton of a mathematical expression by using constant placeholders. And a constant optimizer is used to find the values in these constant placeholders. AEG-GP \cite{korns2011accuracy,korns2013baseline} uses the abstract expression grammar to represent the skeleton of a mathematical expression and utilizes PSO \cite{kennedy1995particle} to find constant values. Unlike AEG-GP, PGE \cite{worm2013prioritized} is a deterministic SR algorithm that offers reliable and reproducible results. While PGE maintains a tree-based representation and Pareto non-dominated sorting from GP, it replaces the genetic operators and random numbers with grammar rules. The method also uses nonlinear regression to fit the constants of a mathematical equation.   
The approaches to GP-based feature engineering, such as GP-based feature construction \cite{krawiec2002genetic}, MRGP \cite{arnaldo2014multiple}, FEW \cite{la2017general}, M3GP \cite{munoz2019evolving}, and FEAT \cite{cava2018learning}, utilize EC to search for possible representations and couple with an ML model to handle the parameters of the representations. 
Different from GP-based feature engineering approaches, FFX \cite{mcconaghy2011ffx} is a deterministic SR algorithm. It enumerates a massive set of basic features (basis functions--$B_i(x)$) by a production rule. It then find coefficient values($a$) in "$y=a_0+ \sum_{i=1}^{N} a_i \times B_i(x)$" by using pathwise regularized learning. 


The other research line uses the hybrid of a neural network and a GP, called \textbf{DL-GP}. DL-GPs \cite{zhong2018deep,cranmer2020discovering,mundhenk2021symbolic,xing2021automated} leverage a neural network to obtain features from the given dataset and apply these features to guide GP. For example, Xing et al. \cite{xing2021automated} design an encoder-decoder neural network based on super-resolution ResNet to predicate the importance of each mathematical operator from the given data; and utilize the importance of each mathematical operator to guide GP. Cranmer et al. \cite{cranmer2020discovering} train a graph neural network to represent sparse latent features of the given dataset, and employ GP to generate symbolic expressions fitted to these latent features. 
The DL-GPs still depend on the time-consuming training work and the training dataset. 

Like AIFeynman, TaylorGP also needs to extract the symbolic equation's properties (e.g., separability and low-order polynomial) from the given dataset. However, the difference is that AIFeynman employs a neural network to obtain these properties while TaylorGP achieves the goal using the coefficients in the Taylor series on the given dataset. Therefore, TaylorGP does not need to train a model and does not depend on the training data. 

\section{Taylor Features Analysis} \label{sec:tfa}

\subsection{Obtaining a Taylor polynomial}
Taylor's theorem \cite{jeffreys1999methods} states that if a function $f$ has $n+1$ continuous derivatives on an open interval containing $a$, for each $x$ in the interval,
\begin{equation}
    f(x) = \left[\sum_{k=0}^n \frac{f^{(k)}(a)}{k!} (x-a)^k\right]+ R_{n+1}(x).
\end{equation}
So, the $k$-order Taylor polynomial ($\sum_{k=0}^n \frac{f^{(k)}(a)}{k!} (x-a)^k$) approximates to $f$ around $a$. 

Given a dataset $(X,Y)$, for any point $(x_0,y_0) \in (X,Y)$, the $k$-order Taylor polynomial around the point can be obtained by the following three steps. First, select $k$ points ($\{(x_1,y_1),...,(x_k,y_k)\}$) around $(x_0,y_0)$ from the dataset. 
Next, according to the selected $k$ points, gather $k$ $k$-order Taylor polynomials by Equation \ref{eq:derivatives}. 
\begin{equation} 
\label{eq:derivatives}
\left\{\begin{array}{c}
\left(x_{1}-x_{0}\right) f^{\prime}\left(x_{0}\right)+\ldots+\frac{\left(x_{1}-x_{0}\right)^{k}}{k !} f^{(k)}\left(x_{0}\right)\approx f\left(x_{1}\right)-f\left(x_{0}\right) \\
\left(x_{2}-x_{0}\right) f^{\prime}\left(x_{0}\right)+\ldots +\frac{\left(x_{2}-x_{0}\right)^{k}}{k !} f^{(k)}\left(x_{0}\right)\approx f\left(x_{2}\right)-f\left(x_{0}\right) \\
\ldots \\
\left(x_{k}-x_{0}\right) f^{\prime}\left(x_{0}\right)+\ldots+\frac{\left(x_{k}-x_{0}\right)^{k}}{k !} f^{(k)}\left(x_{0}\right)\approx f\left(x_{k}\right)-f\left(x_{0}\right)
\end{array}\right.
\end{equation}
, where $f(x_i)=y_i$. 
The final step is to obtain the $k$ derivatives ($F$) by Equation \ref{eq:F}
\begin{equation} \label{eq:F}
 F \approx DA^{-1}    
\end{equation}
, where $F=[f^{'}(a),f^{''}(a),...,f^{(k)}(a)]^{T}$, $D=[y_1-y_0,y_2-y_0,...,y_k-y_0]^{T}$. For each $a_{ij} \in A$, $a_{ij}=\frac{\left(x_{i}-x_{0}\right)^{j}}{j !}$. $F$ can generate the $k$-order Taylor polynomial.

\subsubsection{Scaling to a high dimensional dataset}
\label{sec:highdimension}

Mathematically, the higher the order of $k$, the more accurate the Taylor polynomial is. However, in practice, $k$ can not be too high in a high dimensional dataset $D$ because of the two following limitations. 
According to the Taylor's theorem for multivariate functions \cite{holmes2009introduction}, the $n$-variable $k$-order Taylor polynomial has $C_{n+k}^{n}$ terms.
So, $C_{n+k}^{n}-1$ points need to be sampled around the point to obtain the Taylor polynomial. However, if $k$ is too high, $C_{n+k}^{n}-1$ will be greater than all points in $D$. This is impossible. Therefore, $k$ must be limited so that $C_{n+k}^{n}-1 < |D|$. The other limitation is that, if $k$ is too high, the inverse of the matrix $A$ in Equation \ref{eq:F} would be a big challenge to compute since $A$ is an ultra-large-scale ($(C_{n+k}^{n}-1)\times (C_{n+k}^{n}-1)$) matrix. So, we usually set $k=1 \ or \ 2$ in a high-dimensional dataset because of the two limitations.  

\subsection{Extracting Taylor features} \label{Taylorfeature}
Since the above $k$-order Taylor polynomial is generated from a given dataset $(X,Y)$, it can approximate a function $f$ that fits the dataset (i.e., $f(X)=Y$) around a point $(x_0,y_0)$. It can also represent some $f$'s local features, called Taylor features. This paper discusses the Taylor features, low order polynomial, variable separability, function boundary, monotony and parity. 

\subsubsection{Low order polynomial discrimination}
For SR, a key problem is to discriminate whether there is (or only) a low-order polynomial that can represent the given dataset. 
If it exists, a linear regression algorithm can be used to get it and the algorithm could solve SR quickly. The $k$-order Taylor polynomial can easily solve the discrimination problem owing to its coefficients. If a function is a $k$-order polynomial, its Taylor expansion at a point is also a $k$-order polynomial. For example, for "$1.1x+0.2x^2-3.7$", its Taylor expansion at $x=1$ is also '$1.1x+0.2x^2-3.7$'. While, if a function is not a $k$-order polynomial, its Taylor expansion at a point is an infinite-order polynomial. For $1.5sin(x)$, its Taylor expansion is an infinite order polynomial. So, for the $k$-order Taylor polynomial obtained from a dataset, in each term whose degree is greater than $i$ ($i < k$), if the coefficient is zero, the function that $k$-order Taylor polynomial approximates is a low $i$-order polynomial.        

\subsubsection{Variable separability}
For a multivariate function $f(x_1,...,x_n)$, if there is an operator "$\circ$" that lets $f(x_1,...,x_n)= f_1(x_i,..., x_k)$ $\circ f_2(x_m,...,x_p)$ 
where the two variable sets, $\{x_i,...,x_k\}$ and $\{x_m$,$...$,$x_p$ $\}$, both belong to $\{x_1,...,x_n\}$, and $\{x_i,...,x_k\} \cap \{x_m,...,x_p\} = \phi$, 
it is called "$\circ$" separability. The separability property can decompose a complex multivariate function into multiple simple functions.  

If "$\circ$" is addition or multiplication, it is called \textbf{addition separability} or \textbf{multiplication separability}, respectively. 
The $n$-variable $k$-order Taylor polynomial can represent the two separability properties, respectively. For the Taylor polynomial, if the coefficient in each multi-variable term is zero, the function that the Taylor polynomial approximates is addition separability. As shown in Figure \ref{fig:taylorGP}, the Taylor expansion of "$1.1x+0.2x^2+1.5sin(y)-3.7$" is "$1.1x+0.2x^2-3.7+1.5y -0.25y^3+0.0125y^5$", where the coefficient in each multi-variable term is zero, i.e., $c$ in each $cx^iy^j$ is 0. So, according to the addition separability, the Taylor polynomial is decomposed into multiple polynomials, such as "$1.1x+0.2x^2+1.5sin(y)-3.7 \approx (1.1x+0.2x^2-3.7)+(1.5y -0.25y^3+0.0125y^5)$".   

The above method about addition separability also can be used to discriminate multiplication separability. Because, if a function is multiplication separability, i.e., $f(x_1,...,x_n)= f_1(x_i,..., x_k)\times f_2(x_m,...,x_p)$, then $\log f(x_1,...,x_n)= \log f_1(x_i,..., x_k)+ \log f_2(x_m, $ $..., x_p)$. So, if the $n$-variable $k$-order Taylor polynomial obtained after computing the log of the dataset is addition separability, then the function is the multiplication separability.

\subsubsection{Boundary}
\label{Boundedness}

The $k$-order Taylor polynomial can be used to evaluate the boundary of the function $f$ that it approximates to at the interval $[x_a, x_b]$. Since it is a polynomial, its boundary is computed by interval arithmetic \cite{dawood2011theories}. For example, the boundary of the Taylor polynomial "$1.1x+0.2x^2-3.7$", where $x \in [-1,1]$, is "$[-1.1,1.1]+0.2\times[0,1]-3.7 = [-4.8,-2.4]$". 

\subsubsection{Monotonic}
For all points $(x_i,y_i)$ in a dataset (X,Y), if $y_i \geq y_j$ and $x_i \geq x_j$, the function that the dataset represents is a monotonic-increasing function. Otherwise, if $y_i \geq y_j$ and $x_i \leq x_j$, the function is a monotonic-decreasing function. 

\subsubsection{Parity}
If the $k$-order Taylor polynomial $T(x)$ is an odd or even function, i.e., $T(-x)=-T(x)$ or $T(-x)=T(x)$, then the function that it approximates is also an odd or even function. While the method is simple, testing all points is time-consuming.

Another method is to count odd-order terms and even-order terms in the $k$-order Taylor polynomial except for the $0$-order term (constant term). If the $k$-order Taylor polynomial only contains odd (or even) order terms, it is an odd (or even) function. For example, given that $f(x)=1.5sin(x)-3.7$ and its Taylor polynomial is $1.5x-0.25x^{3}+0.0125x^{5}+...+ 4.217e-15x^{17}-3.7$ at the point $x=0$, after removing the $0$-order term "$-3.7$", the Taylor polynomial only contain odd-order terms, such as "$0.25x^{3}$"and "$0.0125x^{5}$". So, it is an odd function.

\section{Taylor Genetic Programming}

TaylorGP, as shown in Figure \ref{fig:taylorGP}, includes the following six steps: 1) obtaining a Taylor polynomial $T$ from a given dataset, 2) decomposing the Taylor polynomial into multiple simple Taylor polynomials ($\{T_1,T_2,...,T_n\}$), 3) discriminating the low order polynomial, 4) extracting function features, 5) running the function feature evolution method, and 6) assembling mathematical expressions. How to execute the steps: 1), 2), 3), and 4) has been introduced in Section \ref{sec:tfa}. Step 6) is simple, which only composes the mathematical expressions found by each simple Taylor polynomials into various complete mathematical equations and evaluates them. So, the following content details step 5).

The function feature evolution method (FFEM), as shown in Algorithm \ref{alg:FFEM}, evolves individuals based on the \textbf{function feature} $F$ that includes boundary, monotonic, and parity. FFEM mainly contains two evolvable operators, individual initialization (initIndividualByFeatures) and individual recombination (recombineByFeatures). The individual initialization operator randomly generates individuals that satisfy the function feature. The individual recombination operator transforms individuals to ensure that the generated individuals satisfy the function feature. In each generation, FFEM leverages individual recombination to produce offspring with the probability $\alpha$; utilizes the individual initialization to produce offspring with the probability $\beta$; saves individuals as other offspring with the probability $(1-\alpha-\beta)$.   

\floatname{algorithm}{Algorithm}  
\renewcommand{\algorithmicrequire}{\textbf{Input:}}  
\renewcommand{\algorithmicensure}{\textbf{Output:}}  
\begin{algorithm}  
    \caption{function feature evolution method}  
    \label{alg:FFEM}
    \begin{algorithmic}[1] 
        \REQUIRE $(X_i,Y_i)$, $\alpha$, $\beta$, $\mathit{threshold}$, $\mathit{maxGen}$,$F$
        \ENSURE $\mathit{best}$
        \STATE $P \gets$ initIndividualByFeatures($F$,popsize=$N$)
        \STATE $\mathit{best} \gets$ selectBestIndividual($P$)
        \WHILE{$\mathit{best.fitness} \le \mathit{threshold} $ \AND $g<\mathit{maxGen}$}
        \FORALL {$i=1$ \TO $N$}
        \STATE $p_{1},p_{2}$ $\gets$ randomSelectTwoIndividuals($P$)
        \IF {rand() < $\alpha$}
        \STATE $\mathit{child}$ $\gets$ recombineByFeatures($p_{1},p_{2}$,$F$)
        \ELSIF {rand() < $\alpha + \beta$} 
            \STATE $\mathit{child}$ $\gets$ initIndividualByFeatures($F$,popSize=$1$)
        \ELSE
        \STATE $\mathit{child}$ $\gets$ $p_{1}$
            \ENDIF
            \STATE $\mathit{nextP}[i]$ $\gets$ $\mathit{child}$
        \ENDFOR
        \STATE $P$ $\gets$ $\mathit{nextP}$
        \STATE $\mathit{best}$ $\gets$ min(best, selectBestIndividual($P$,$(X_i,Y_i)$)) 
        \STATE $g++$
        \ENDWHILE
        \RETURN $\mathit{best}$
    \end{algorithmic}  
\end{algorithm}    

\subsection{Individual initialization}
The probability of randomly generating an individual that satisfies the function feature $F$ is very small. And the process for obtaining $N$ number of these candidate individuals is very time-consuming. To speed up the process, the individual initialization operator first segments the mathematical expression space into many sub-spaces. It then evaluates the function features of these sub-spaces. Finally, it randomly selects the sub-spaces that satisfy $F$ and randomly generates individuals in these sub-spaces until they satisfy $F$.

\subsubsection{Segmenting mathematical expression space }
A tree can be used to represent a mathematical expression. A tree with depth $h$ also shows a sub-space that contains all mathematical expressions expanded from the tree. Moreover, all trees with depth $h$ represent a segment of the mathematical expression space. For example, given a basic function set $\{ +, sin \}$ and a variable set $\{ x, c\}$, the mathematical expression space is divided into the sub-spaces encoded by trees with depth 3, such as "$+ + sin c x x$", "$+++ x c c$", "$+sin sin xx$", and "$sin+xc$". The sub-space "$++sin c x x $" is a mathematical expression "$c+x+sin(x)$", which contains all mathematical expressions expanded from "$+ + sin c x x$", such as "$++sin+cxxx$"="$(c+x)+sin(x)+x$".

\subsubsection{Evaluating sub-space}
Interval arithmetic \cite{dawood2011theories} can be used to compute the boundary of a sub-space. In the tree that represents the sub-space, if there is a path from a leaf node to the root node consisting of unbound functions ($x, +,-,\times, /, a^x, \ln, ...$), the boundary of the sub-space is $[-\infty, \infty]$. Otherwise, it is computed by interval arithmetic. For example, for the sub-space "$+ + sin c x x$", there is a path "$x++$", so its boundary is $[-\infty, \infty]$. There is no such path for the sub-space "$+sin sin xx$", so its boundary is [-2,2] obtained by interval arithmetic. 

A sub-space's non-monotonic or monotonic increasing/decrease is determined by its derivative $d$. If $d \geq 0$ (or $d \leq 0$) in all variable values, it is a monotone increasing (decreasing) function. Otherwise, it is a non-monotone function. The sub-space is an odd/even function according to "$f(-x)=-f(x)$" or "$f(-x)=f(x)$" for all values in $x$. 
For the sub-space "$+ + sin c x x$=$c+x+sin(x)$", its derivative is "$1+cos(x)$", meaning that it is a monotone-increasing function. 
Owing to "$c$" in "$c+x+sin(x)$", it is a non-odd and non-even function.

\subsubsection{Generating individual}
 
The method "generating individual" obtains the segmented sub-spaces whose boundaries contain the given boundary. It then randomly selects a sub-space from these sub-spaces. If the sub-space does not satisfy the given monotony and parity requirements, the method randomly generates a new individual from the sub-space until it satisfies the given function features. Otherwise, the method randomly generates an individual based on the following rules, as listed in Table \ref{tab:rule}. 
For example, given the function features $\{[-10,10], odd\}$, for the selected sub-space "$++sin c x x $=$c+x+sin(x)$", it is a non-odd and non-even function. So, the method randomly generates individuals until an individual is an odd function whose boundary contains $[-10,10]$. For the selected sub-space "$++sin x x x $=$2x+sin(x)$", it is an odd function. So, the method randomly constructs an odd function (e.g., $sin(x)$). It then combines the odd function and the selected sub-space according to an operator randomly selected (e.g. "$+$") from $\{+,-,\times, /,f(g(x)\}$. With these steps the method finally generates an individual "$2x+2sin(x)$" whose boundary contains $[-10,10]$.

\begin{table}[htbp]
  \caption{Function Combination Rules}
    \label{tab:rule}
    \scalebox{0.89}{
    \begin{tabular}{cccc}
    \toprule
    op & $f(x)$ &  $g(x)$ & results \\
    \midrule
    $+$ & odd     &  odd    &  odd  \\
    $+$ & even    &  even   &  even  \\
    $-$ & odd     &  odd    &  odd  \\
    $-$ & even    &  even   &  even  \\
    $\times, /$ & odd     &  odd    &  even  \\
    $\times, /$ & even    &  even   &  even  \\
    $\times, /$ & odd    &  even   &  odd  \\
    $f(g(x))$ & even    &  even   &  even  \\
    $f(g(x))$ & odd    &  odd   &  odd  \\
    $f(g(x))$ & even    &  odd   &  even  \\
    $f(g(x))$ & odd    &  even   &  even  \\
    $+$      & $\nearrow$  &  $\nearrow$   &  $\nearrow$  \\
    $\times$      & $\nearrow$  &  $\nearrow$   &  $\nearrow$  \\
    $f(g(x))$     & $\nearrow$  &  $\nearrow$   &  $\nearrow$  \\
  \bottomrule
  \end{tabular}
  }
\begin{tablenotes}
\footnotesize
    \item "$\nearrow$" represents a monotone-increasing function. The monotone-decreasing function has similar properties to the monotone-increasing function. 
\end{tablenotes}
\end{table}

\subsection{Individual recombination}
According to the rules in Table \ref{tab:rule}, the individual recombination operator recombines two individuals from the population to construct an individual that satisfies the given function feature. Meanwhile, if the recombined individual exceeds the limit length, the operator prunes it to avoid the individual bloating. For example, given a function feature -- odd function, the operator recombines the two individuals "$2x+2sin(x)$" and $x+x^3$ with "$+$". The recombined individual "$2x+2sin(x)+x+x^3$" exceeds the limited length 12. It then prunes "$sin(x)$" in the individual and replaces "$sin(x)$" with "$x$" according to "$f(g(x))$".  The method in the end generates the individual "$5x+x^3$" whose length is 9.

\section{Experiment}

\subsection{Datasets}

We evaluate the performance of TaylorGP on three kinds of benchmarks:  classical Symbolic Regression Benchmarks (\textbf{SRB}) \cite{mcdermott12benchmark},  Penn Machine Learning Benchmarks (\textbf{PMLB}) \cite{Olson2017PMLB}, and Feynman Symbolic Regression Benchmarks (\textbf{FSRB}) \cite{udrescu2020ai_a}. SRB consists of twenty-three SR problems derived from the five canonical symbolic regression benchmarks,  Nguyen \cite{uy2011semantically}, Korns \cite{korns2011accuracy}, Koza \cite{koza2002benchmark}, Keijzer \cite{keijzer2003improving}, and Vladislavleva \cite{vladislavleva2008order}. PMLB includes seven regression tasks and three classification tasks. FSRB contains forty-eight Feynman equations in \cite{udrescu2020ai_a}. The distribution of the total 81 benchmark sizes by samples and features is shown in Figure \ref{fig:datasets_size}. The details of these benchmarks are listed in the appendix.  

\begin{figure}[h]
  \centering
  \includegraphics[width=0.75\linewidth]{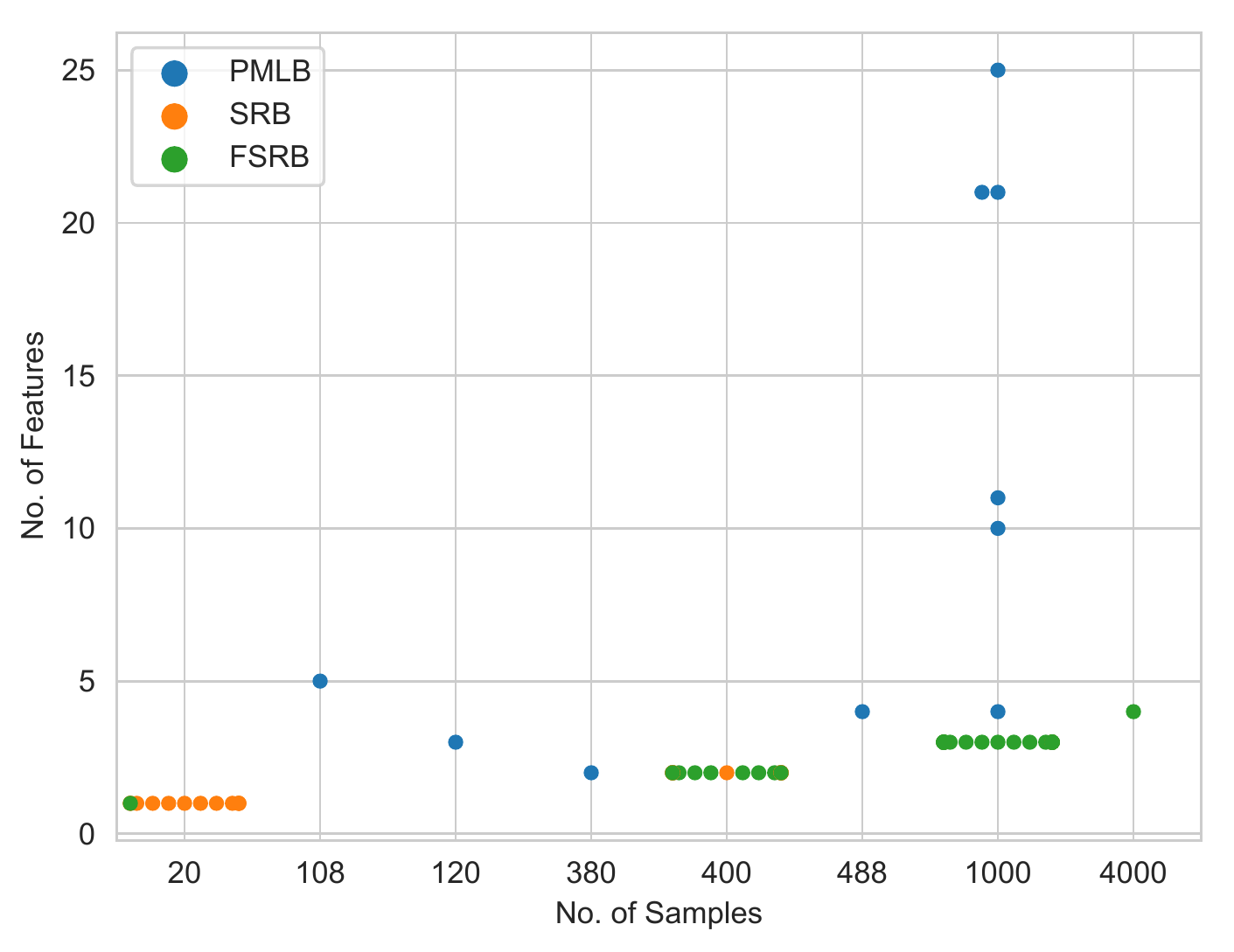} 
  \caption{Properties of Benchmarks.}
  \label{fig:datasets_size}
  \Description{Properties of Benchmarks.}
\end{figure}

\subsection{Algorithm Parameter Settings}
We compare TaylorGP with two kinds of baseline algorithms \footnote{The nine baseline algorithms are implemented in SRBench \cite{cava2021contemporary}}: four symbolic regression methods and five machine learning methods. The symbolic regression methods include \textbf{GPlearn}\footnote{https://github.com/trevorstephens/gplearn}, \textbf{FFX} \cite{mcconaghy2011ffx}, geometric semantic genetic programming (\textbf{GSGP})\cite{hutchison_geometric_2012} and bayesian symbolic regression (\textbf{BSR}) \cite{jin2019bayesian}. The machine learning methods include linear regression (\textbf{LR}), kernel ridge regression (\textbf{KR}), random forest regression (\textbf{RF}), support vector machines (\textbf{SVM}), and \textbf{XGBoost} \cite{chen2016xgboost}.  
The detailed parameters of each algorithm are tuned according to Table \ref{tab:parameters}. 
\renewcommand{\arraystretch}{1}
\begin{table}
	\centering
	\setlength{\abovecaptionskip}{0.1cm}
	\caption{Algorithm parameters}
	\label{tab:parameters}
	\scalebox{0.69}{
		\begin{tabular}{ccccccc}
			\toprule
			Name&&Parameter&Value\\
			\midrule
			TaylorGP&&Function Set&$+$,$-$,$\times$,$\div$,$\sin$,$\cos$,$ln(\left| n \right|)$,$exp$,$sqrt$ \\
			&&Max Generations&10000\\
			&&Population Size&1000\\
			&&Crossover Rate&0.7\\
			&&Mutation Rate&0.2\\
			&&Copy Rate&0.1\\
			&&Stopping Threshold&1e-5\\
		
			\midrule
			GPLearn&&Function Set&$+$,$-$,$\times$,$\div$,$\sin$,$\cos$,$ln(\left| n \right|)$,$exp$,$sqrt$ \\
			&&max generations&10000\\
			&&Population Size&1000\\
			&&Crossover Rate&0.7\\
			&&Mutation Rate&0.2\\
			&&Copy Rate&0.1\\
			&&Stopping Threshold&1e-5\\
			\midrule		
		    GSGP&&Function Set&$+$,$-$,$\times$,$\div$ \\
		    &&Max Generations&10000\\
			&&Population Size&1000\\
			&&Crossover Rate&0.7\\
			&&Mutation Rate&0.2\\
		    &&Stopping Threshold&1e-5\\
			
			\midrule		
		    BSR&&Function Set&$+$,$-$,$\times$,$\div$,$\sin$,$\cos$,$ln(\left| n \right|)$,$exp$,$sqrt$ \\
		    &&MM&10000\\
			&&k&2\\
		    &&Stopping Threshold&1e-5\\
			\midrule		
		    FFX&&None\\
			\midrule		
            LR&&Normalize&FALSE\\
			
			\midrule		
            KR&&Kernal&{'linear', 'poly', 'rbf', 'sigmoid'}\\
			&&Gamma&{0.01,0.1,1,10}\\
		    &&Regularization&{0.001,0.1,1}\\
			\midrule
			
            RF&&Number of Estimators&{10, 100, 1000}\\
			&&Max Features&{'sqrt','log2',None}\\
			\midrule		
			
            SVM&&Kernal&{‘linear’, ‘poly’, ‘rbf’, ‘sigmoid’, ‘precomputed’}\\
			\midrule		
            XGBoost&&Learning Rate&{0.0001,0.01, 0.05, 0.1, 0.2}\\
			&&Gamma&{0,0.1,0.2,0.3,0.4}\\
			\bottomrule
		\end{tabular}
	}

\end{table}

\section{Results and Discussion}
\subsection{Performance Metrics}
TaylorGP and nine baseline algorithms run 30 times on each benchmark. Their fitness results are listed in the appendix. In addition, the following $R^2$ test \cite{cava2021contemporary} is introduced to evaluate the performance of these algorithms on these benchmarks. 
\begin{equation}
    \operatorname{R^{2}} = 1-\frac{\sum_{i=1}^{n}\left(\hat{y}_{i}-y_{i}\right)^{2}}{\sum_{i=1}^{n}\left(\bar{y}-y_{i}\right)^{2}},
\end{equation}
, where $y_{i}$ is the value in the dataset, $\bar{y}$ is mean and $\hat{y}_{i}$ is the output value of the best solution. 

\begin{figure}[h]
  \centering
  	\setlength{\abovecaptionskip}{0.1cm}
  \includegraphics[width=0.9\linewidth]{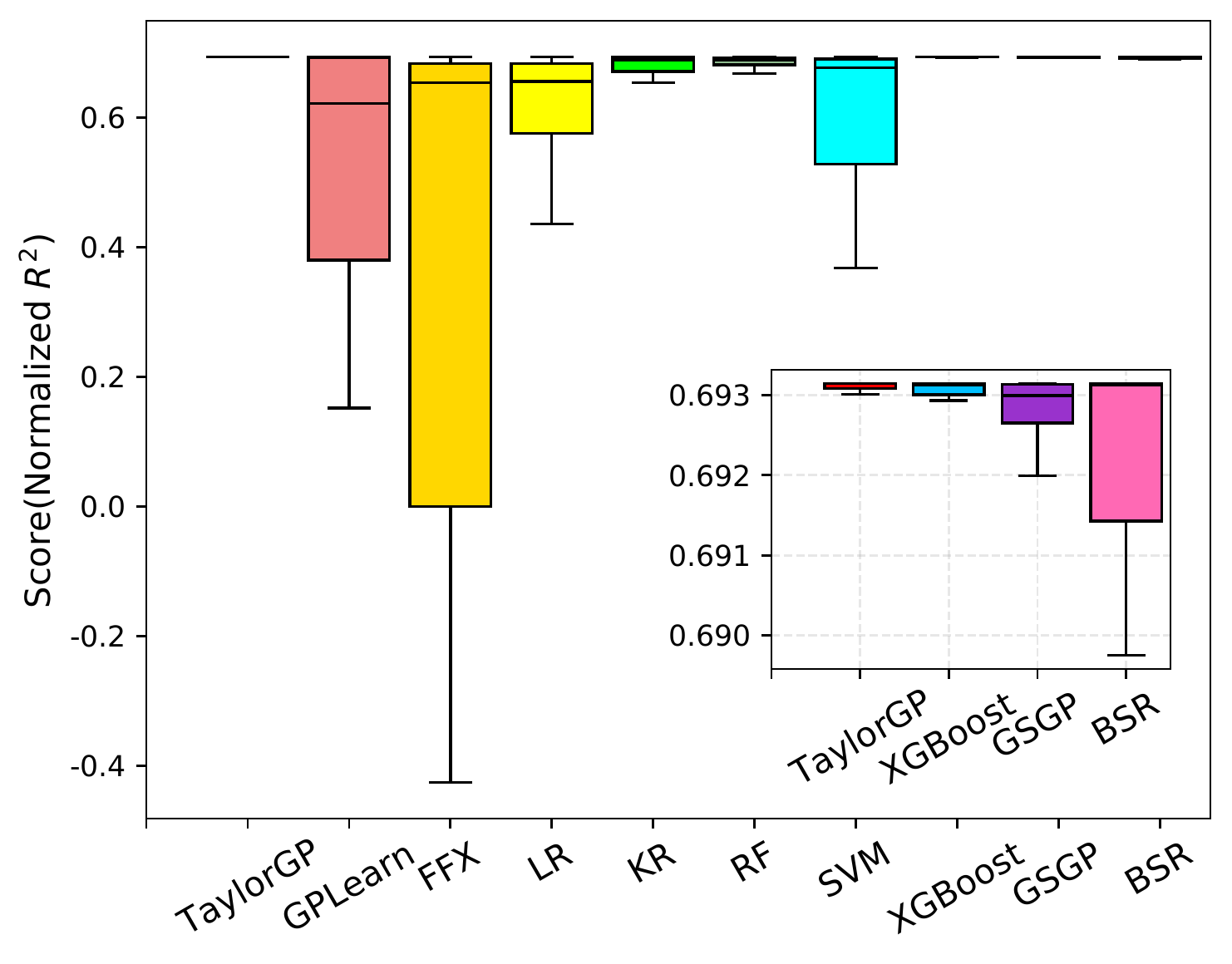} 
  \caption{Normalized $R^{2}$ scores of the ten algorithms.}
  \Description{R$^{2}$ Performance.}
  \label{fig:normalizedR}
  \vspace{-0.2cm}
\end{figure}

Figure \ref{fig:normalizedR} illustrates the normalized $R^2$ scores of the ten algorithms running 30 times on all benchmarks. Since the normalized $R^2$ closer to 1 indicates better results, overall TaylorGP can find more accurate results than other algorithms. Moreover, TaylorGP's results are more stable. The normalized $R^2$ scores of the ten algorithms on each benchmark (in the appendix) show that TaylorGP can outperform the nine baseline algorithms on most benchmarks.

Table \ref{tab:wilcoxonR2} shows that TaylorGP still outperforms the nine baseline algorithms on the pairwise statistical comparisons with the Wilcoxon signed-rank test. Except for TaylorGP,  it is not easy to find one algorithm outperform all other algorithms consistently on the benchmarks.  
\renewcommand{\arraystretch}{1}
\begin{table*}[htp]  
\setlength{\abovecaptionskip}{0.1cm}
	\centering  
	\begin{threeparttable}  
  \caption{Wilcoxon signed-rank test of normalized $R^{2}$ scores for pairwise statistical comparisons.}
		\label{tab:wilcoxonR2} 
		\begin{tabular}{cccccccccc}  
			\toprule
			& TaylorGP& GPLearn & FFX & LR & KR & RF & SVM & XGBoost & GSGP
			\cr 
			\midrule  
            GPLearn&\textbf{5.73e-13}&&&&&&&&\cr
        	FFX&\textbf{8.87e-11}&\textbf{6.55e-03}&&&&&&&\cr
        	LR&\textbf{5.86e-15}&8.70e-01&9.82e-01&&&&&& \cr
        	KR&\textbf{2.73e-15}&1.00e+00&1.00e+00&1.00e+00&&&&&\cr
        	RF&\textbf{4.31e-14}&1.00e+00&1.00e+00&1.00e+00&8.04e-01&&&&\cr
        	SVM&\textbf{2.84e-15}&5.34e-01&1.00e+00&7.74e-01&\textbf{2.28e-05}&\textbf{1.36e-05}&&&\cr
        	XGBoost&\textbf{6.72e-05}&1.00e+00&1.00e+00&1.00e+00&1.00e+00&1.00e+00&1.00e+00&&\cr
        	GSGP&\textbf{1.08e-04}&1.00e+00&1.00e+00&1.00e+00&1.00e+00&1.00e+00&1.00e+00&\textbf{2.43e-03}&\cr
        	BSR&\textbf{3.09e-03}&1.00e+00&1.00e+00&1.00e+00&1.00e+00&1.00e+00&1.00e+00&\textbf{3.62e-02}&9.84e-02\cr
			\bottomrule
		\end{tabular} 
 		\begin{tablenotes}
 		\footnotesize
          \item[1] bold number means that $p<0.05$.
 		\end{tablenotes}
	\end{threeparttable}  
\end{table*} 

\subsection{Discussion}
Why does TaylorGP outperform the nine baseline algorithms on most benchmarks? The main reason is that the Taylor features can guide TaylorGP to search the problem space more effectively than the baselines. Compared with the other five ML methods, TaylorGP does not need to construct a predefined model to find a model that fits the given dataset. Therefore, on large-scale benchmarks, it can find better results. Compared with other GPs that need to search the whole mathematical expression space, TaylorGP's search space is smaller. So, it can find the correct results faster.

\subsubsection{Convergence Analysis}
\begin{figure}[h]
	\setlength{\abovecaptionskip}{0.1cm}
  \centering
  \includegraphics[width=0.9\linewidth]{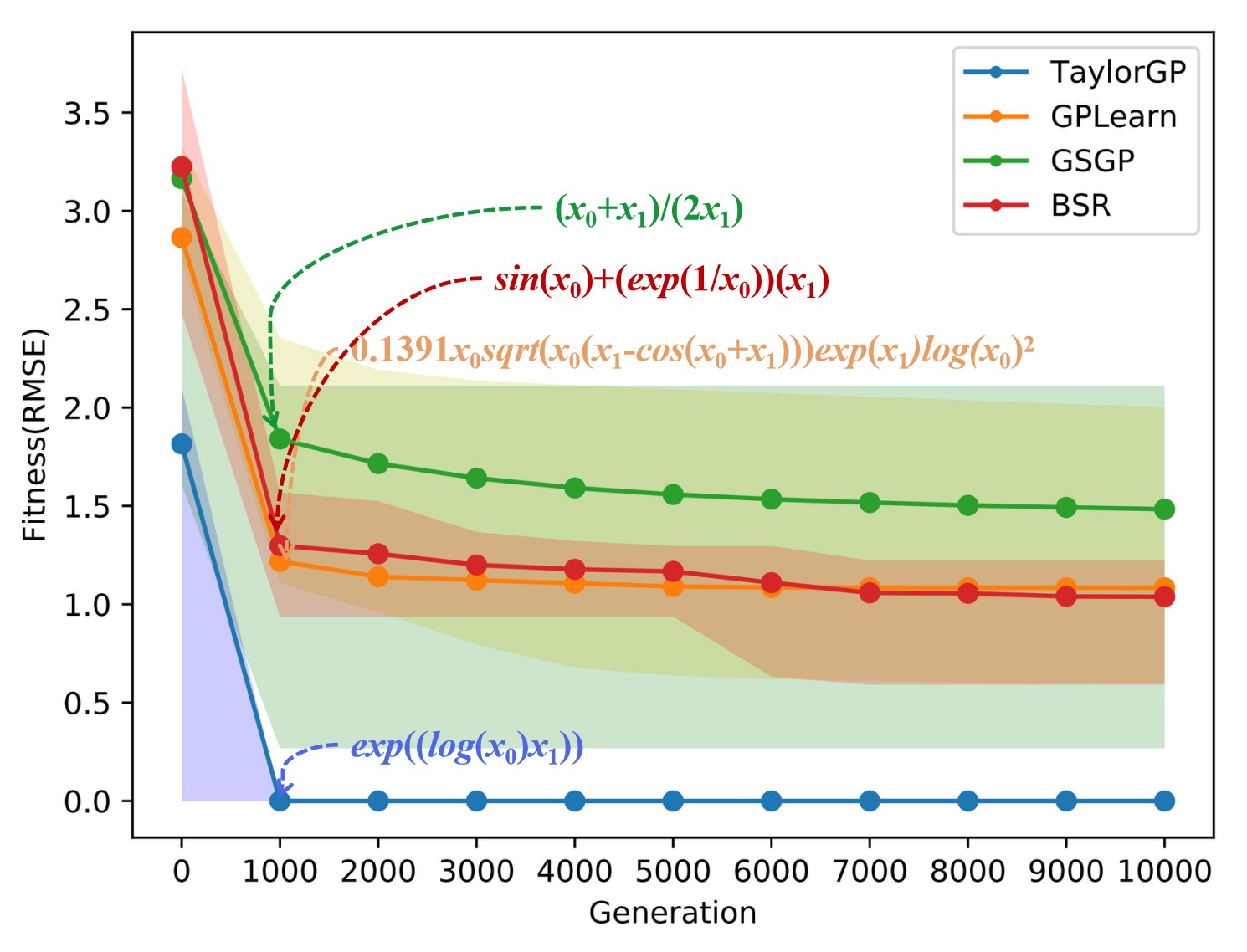} 
  \caption{Convergence Comparison for "$x_{0}^{x_{1}}$".}
	\label{fig:convergence1}
\end{figure}

\begin{figure}[h]
  \centering
  \setlength{\abovecaptionskip}{0.1cm}
  \includegraphics[width=0.9\linewidth]{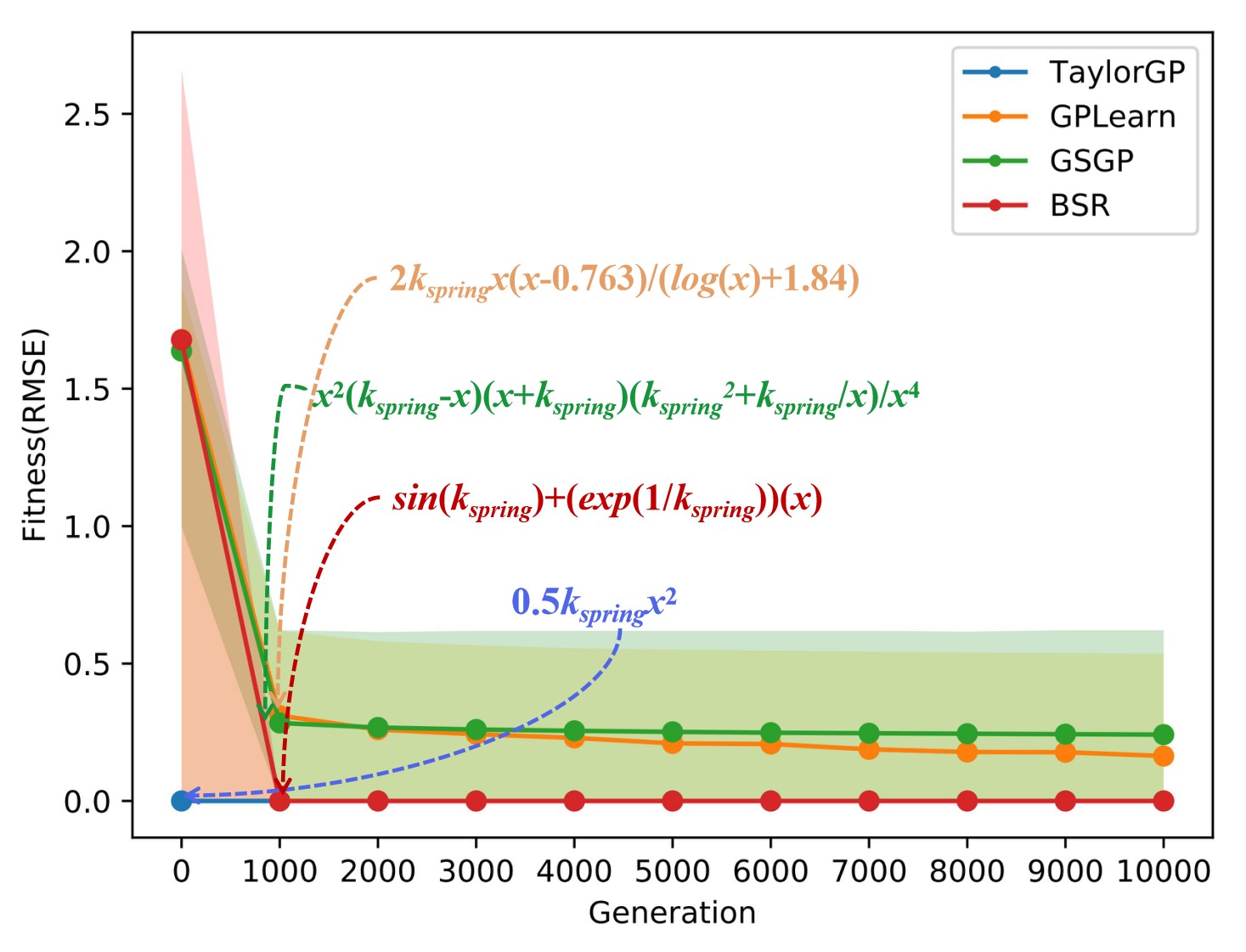} 
  \caption{Convergence Comparison for 
  "$ U = \frac{1}{2} k_{spring} x^{2}$", where $k_{spring}$ and $x$ are variables.}
	\label{fig:convergence2}
	\vspace{-0.2cm}
\end{figure}

We compare TaylorGP with the other three SR methods, GPLearn, GSGP, and BSR. Two benchmarks are used in the following evaluation. One benchmark is the "$x_0^{x_1}$" from SRB. The other is the "$U=\frac{1}{2}k_{spring}x^2$" from FSRB. We illustrate how the Taylor features help TaylorGP quickly find the correct results through the two benchmarks. Figures \ref{fig:convergence1} and \ref{fig:convergence2} show the processes of the four methods running on the two benchmarks, respectively. 

For the benchmark "$x_0^{x_1}$" where $x_0$ and $x1$ both are in $[2,4]$, compared with GPLearn, GSGP and BSR, TaylorGP can find the optimal result "$exp(log(x_0)x_1)=x_0^{x_1}$" at the $1000$th generation. While the other three algorithms still can not find the optimal results until they run for 10,000 generations. 
TaylorGP first generates the Taylor polynomial "$(2.596e-3)x_0^4+2.932x_0^3x_1-7.828x_0^3+15.171x_0^2x_1^2-100.026x_0^2x_1+163.688x_0^2+11.398x_0x_1^3-167.424x_0x_1^2+710.31x_0x_1-932.144x_0+1.636x_1^4-47.859x_1^3+416.67x_1^2-1393.536x_1+1590.457$" at the point(2,2). 
According to the Taylor polynomial, the function boundary is $[4.233,230.513]$, and the function is monotonically increasing. The two Taylor features can reduce the space used to initialize individuals. As "$\log(x_0)$", "$x_0\times x_1$", and "$exp(x_0)$" all are monotone increasing functions at the range [2,4], they are very likely to be selected as initialized. 
The individual recombination that recursively merges the three functions using the operator $f(g(x))$ in Table \ref{tab:rule} may generate "$exp(log(x_0)x_1)$". So, TaylorGP, compared with the other three algorithms, can initialize better individuals and get the optimal result earlier, as shown in Figure \ref{fig:convergence1}. 

\begin{figure*}[h]
	\centering
    \hspace{-0.5cm} \quad
	\subfigure[SRB]{
\includegraphics[width=5.5cm]{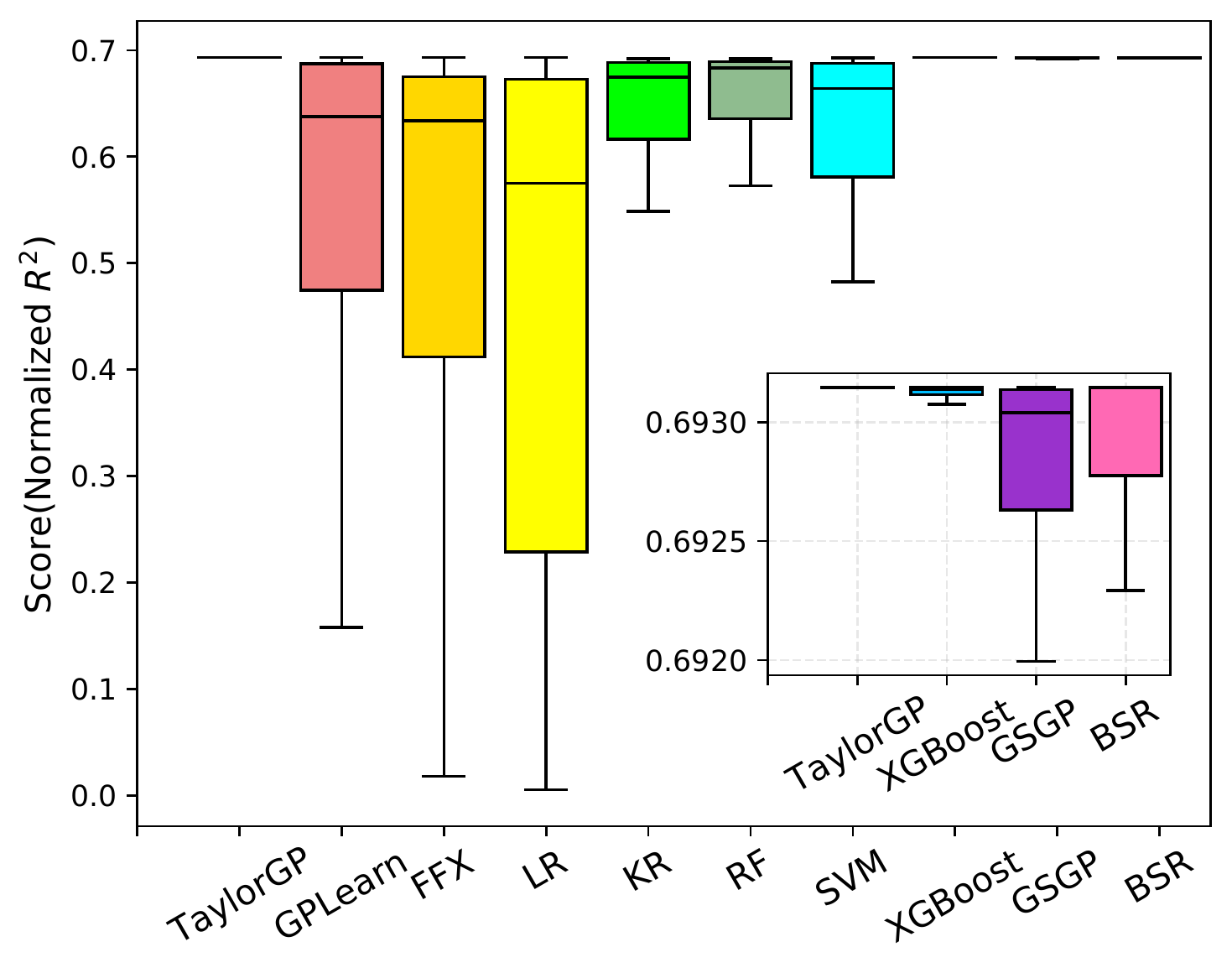}
	}
	\hspace{-0.5cm}  \quad
	\subfigure[FSRB]{
\includegraphics[width=5.5cm]{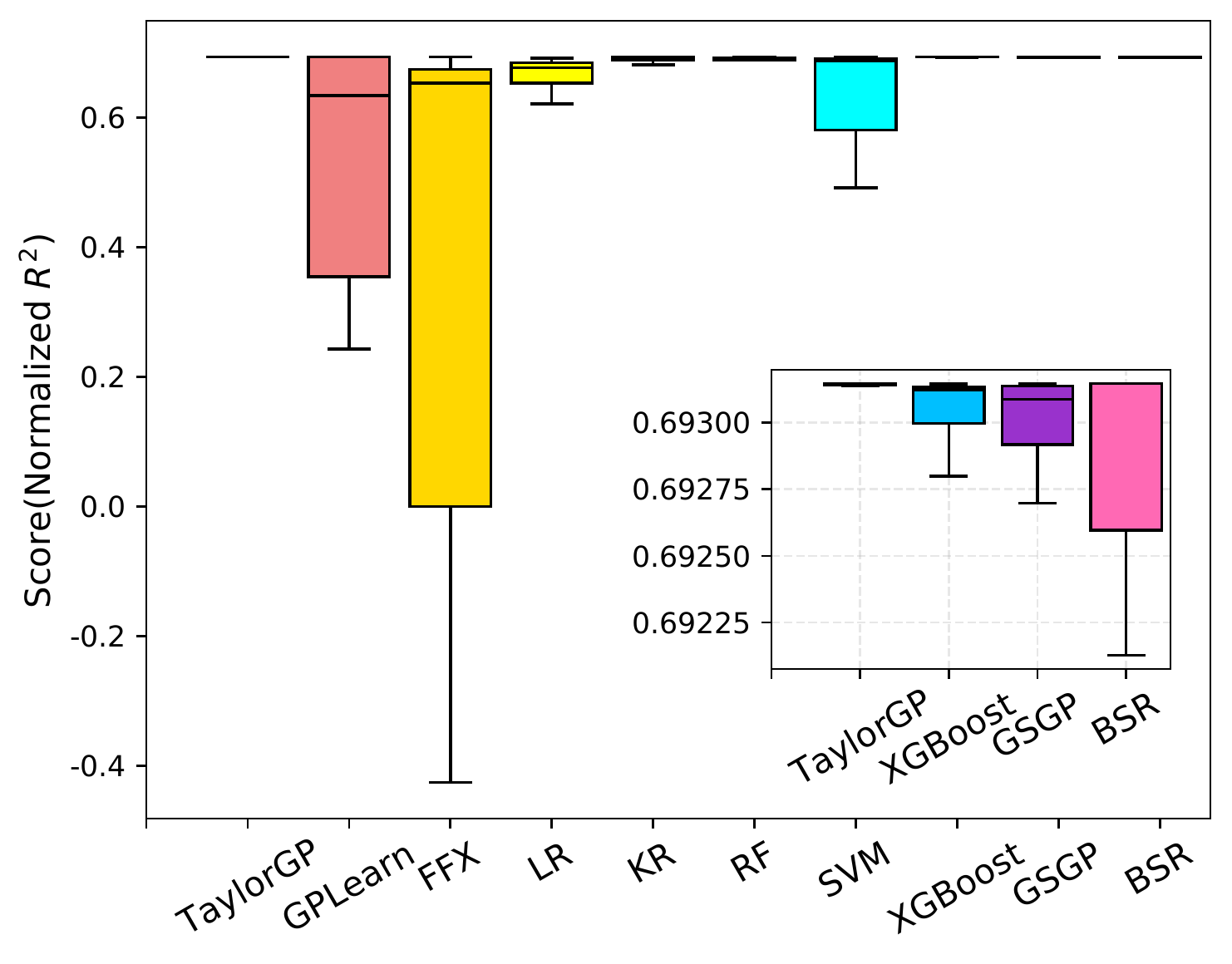}
    }
    \hspace{-0.5cm}  \quad
	\subfigure[PMLB]{
\includegraphics[width=5.5cm]{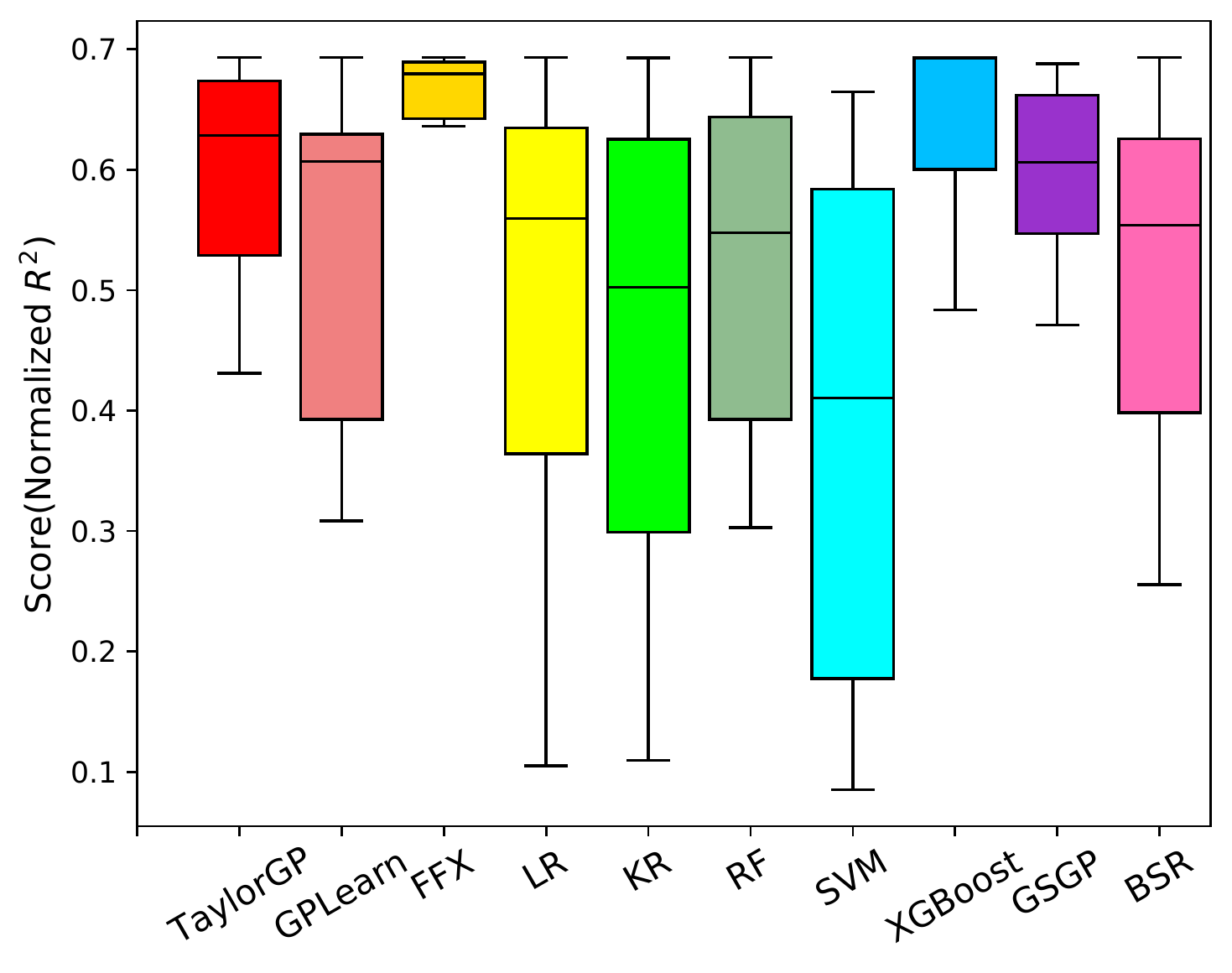}
	}
	\vspace{-0.5cm}
   \caption{Normalized $R^{2}$ comparisons of the ten SR methods on classical Symbolic Regression Benchmarks (\textbf{SRB}), Feynman Symbolic Regression Benchmarks (\textbf{FSRB}), and Penn Machine Learning Benchmarks (\textbf{PMLB}), respectively.}
  \label{fig:GECCO_Feynman_ML_Box}

\end{figure*}

For the Feynman benchmark "$U=\frac{1}{2}k_{spring}x^2$", TaylorGP finds the optimal results at the $0$th generation, because TaylorGP can directly obtain $0.2k_{spring} x^{2}$ owing to the low polynomial discrimination. TaylorGP achieves the Taylor polynomial "$(1.454e-6)x_0^3-(1.181e-6)x_0^2x_1-(7.784e-6)x_0^2+(0.5)x_0x_1^2+(5.809e-6)x_0x_1+(9.249e-6)x_0-(2.53e-6)x_1^3+(2.041e-5)x_1^2-(6.481e-5)x_1+(5.579e-5)$" from the benchmark. After omitting insignificant coefficients that are less than $e-4$, the Taylor polynomial is "$0.5x_0 x_1^{2}$". As the RMSE of "$0.5x_0 x_1^{2}$" is less than the stopping threshold $e-5$, it is the final result that TaylorGP finds.   

Besides the two figures, the figures of the convergence comparison that the four algorithms run on the other benchmarks are listed in the appendix.

\subsubsection{Fitness Analysis}
Figure \ref{fig:GECCO_Feynman_ML_Box} illustrates that TaylorGP, when compared with the nine baseline algorithms, can obtain more accurate and stable results on the two benchmarks, SRB and FSRB. However, on the benchmark PMLB, the two algorithms, FFX and XGBoost, outperform TaylorGP. This is due to PMLB has more features (variables) than the other two benchmark sets, Figure 6 shows that TaylorGP has the best performance (normalized $R^2$ score) on the-low dimensional datasets. 
In contrast, its performance degrades as the dataset's dimension increases. For a high dimension dataset, TaylorGP can only obtain a low order Taylor polynomial according to the analysis in Section \ref{sec:highdimension}. However, the low order Taylor polynomial may not approximate the real function that fits the given high-dimensional dataset. The Taylor features extracted from the Taylor polynomial may be incorrect or incomplete; therefore the features cannot help TaylorGP find a correct result.


\subsubsection{The accuracy of extracting Taylor features}

As the real function that fits the dataset in PMLB is unknown, Table \ref{tab:recognitionrate} lists the accuracy of extracting each Taylor feature on SRB and FSRB (total 71 benchmarks). TaylorGP can correctly identify the two Taylor features, monotone and boundary, on all benchmarks, meaning that the two Taylor features always help Taylor reduce the search space. 

However, TaylorGP recognizes the variable separability and even/odd function with low accuracies (12.5\% and 36.7\%). 
For identifying the variable separability and the odd/even function, TaylorGP requires that the Taylor polynomial can not contain some order terms, i.e., the coefficients in these order terms must be zero. However, as the Taylor polynomial approximates the real function around a point, some inconsistencies exist between the polynomial and the real function. The coefficients on these terms are slight errors. These slight error coefficients affect the recognition of the variable separability and the odd/even function. For example, for $sin(x)$, its Taylor polynomial at the point (0,0) is "$\frac{1}{1!}x - \frac{1}{3!}x^3+ \frac{1}{5!}x^5-\frac{1}{7!}x^7+...$". However, according to Equation \ref{eq:F}, TaylorGP sets a $4$-order Taylor polynomial and obtains the polynomial "$0.015+\frac{1}{1!}x + 0.003x^2- \frac{1}{3!}x^3$" from the given dataset. The polynomial is not an odd function due to the two coefficients, "0.015" and "0.003". To prevent this from happening, we set a threshold for these coefficients and omit the terms whose coefficients are less than the threshold. However, it is not easy to get a suitable threshold for the Taylor polynomial because of the diversity of datasets.

\renewcommand{\arraystretch}{1}
\begin{table}[h]\large
	\centering
	\setlength{\abovecaptionskip}{0.1cm}
	\caption{The accuracy of extracting Taylor features on 71 benchmarks.}
	\label{tab:recognitionrate}
	\scalebox{0.85}{
		\begin{tabular}{cccc}
			\toprule
           	Taylor Features&Accuracy&Correct No &Ground Truth No\\
			\midrule
            LowOrderPoly & 73.9\% &17 &23\\
        	Separability & 12.5\% &3  &24\\
        	Boundary     &100.0\% &71 &71 \\
        	Odd/even function &36.7\% &18 &49\\
        	Monotone &100.0 \%&10 &10\\
			\bottomrule
		\end{tabular}
	}
\end{table}

Although the two Taylor features (variable separability, odd/even function) have a low recognition accuracy, they still can help TaylorGP to find the correct symbolic equation, such as running TaylorGP on the two above benchmarks, "$x_0^{x_1}$" and "$U=\frac{1}{2}k_{spring}x^2$". So, TaylorGP can utilize the Taylor features to reduce its search space and speed up its search.  

\section{Conclusion}
This paper proposes a new method called TaylorGP to search the mathematical expression space using Taylor features. As most of the Taylor features are obtained by the coefficients in a Taylor polynomial, the modeling process can be computationally efficient and straightforward to implement. TaylorGP leverages the two operators based on Taylor features, individual initialization, and individual recombination, to evolve the population. Experiments show that TaylorGP can quickly find the correct result with the help of the two evolution operators. 

However, TaylorGP will degrade when the dataset dimension increases because of the local approximation of the Taylor polynomial. In a high-dimensional dataset, a low order Taylor polynomial obtained from the dataset only represents the dataset's local features, not global features. So, our future work will involve investigating how to utilize many low-order Taylor polynomials to represent global features in high-dimensional datasets.

\begin{acks}
This work is supported by China National Key Research Project (No.2019YFC0312003) 
\end{acks}

\bibliographystyle{ACM-Reference-Format}
\bibliography{references}


\begin{thebibliography}{53}


\ifx \showCODEN    \undefined \def \showCODEN     #1{\unskip}     \fi
\ifx \showDOI      \undefined \def \showDOI       #1{#1}\fi
\ifx \showISBNx    \undefined \def \showISBNx     #1{\unskip}     \fi
\ifx \showISBNxiii \undefined \def \showISBNxiii  #1{\unskip}     \fi
\ifx \showISSN     \undefined \def \showISSN      #1{\unskip}     \fi
\ifx \showLCCN     \undefined \def \showLCCN      #1{\unskip}     \fi
\ifx \shownote     \undefined \def \shownote      #1{#1}          \fi
\ifx \showarticletitle \undefined \def \showarticletitle #1{#1}   \fi
\ifx \showURL      \undefined \def \showURL       {\relax}        \fi
\providecommand\bibfield[2]{#2}
\providecommand\bibinfo[2]{#2}
\providecommand\natexlab[1]{#1}
\providecommand\showeprint[2][]{arXiv:#2}

\bibitem[\protect\citeauthoryear{Arnaldo, Krawiec, and O'Reilly}{Arnaldo
  et~al\mbox{.}}{2014}]%
        {arnaldo2014multiple}
\bibfield{author}{\bibinfo{person}{Ignacio Arnaldo}, \bibinfo{person}{Krzysztof
  Krawiec}, {and} \bibinfo{person}{Una-May O'Reilly}.}
  \bibinfo{year}{2014}\natexlab{}.
\newblock \showarticletitle{Multiple regression genetic programming}. In
  \bibinfo{booktitle}{\emph{Proceedings of the 2014 Annual Conference on
  Genetic and Evolutionary Computation}}. \bibinfo{pages}{879--886}.
\newblock


\bibitem[\protect\citeauthoryear{Biggio, Bendinelli, Neitz, Lucchi, and
  Parascandolo}{Biggio et~al\mbox{.}}{2021}]%
        {biggio2021neural}
\bibfield{author}{\bibinfo{person}{Luca Biggio}, \bibinfo{person}{Tommaso
  Bendinelli}, \bibinfo{person}{Alexander Neitz}, \bibinfo{person}{Aurelien
  Lucchi}, {and} \bibinfo{person}{Giambattista Parascandolo}.}
  \bibinfo{year}{2021}\natexlab{}.
\newblock \showarticletitle{Neural Symbolic Regression that scales}. In
  \bibinfo{booktitle}{\emph{Proceedings of the 38th International Conference on
  Machine Learning}}, Vol.~\bibinfo{volume}{139}. PMLR,
  \bibinfo{pages}{936--945}.
\newblock


\bibitem[\protect\citeauthoryear{Brameier and Banzhaf}{Brameier and
  Banzhaf}{2007}]%
        {brameier2007linear}
\bibfield{author}{\bibinfo{person}{Markus~F Brameier} {and}
  \bibinfo{person}{Wolfgang Banzhaf}.} \bibinfo{year}{2007}\natexlab{}.
\newblock \bibinfo{booktitle}{\emph{Linear genetic programming}}.
\newblock \bibinfo{publisher}{Springer Science \& Business Media}.
\newblock


\bibitem[\protect\citeauthoryear{Brown, Mann, Ryder, Subbiah, Kaplan, Dhariwal,
  Neelakantan, Shyam, Sastry, Askell, et~al\mbox{.}}{Brown
  et~al\mbox{.}}{2020}]%
        {brown2020language}
\bibfield{author}{\bibinfo{person}{Tom~B Brown}, \bibinfo{person}{Benjamin
  Mann}, \bibinfo{person}{Nick Ryder}, \bibinfo{person}{Melanie Subbiah},
  \bibinfo{person}{Jared Kaplan}, \bibinfo{person}{Prafulla Dhariwal},
  \bibinfo{person}{Arvind Neelakantan}, \bibinfo{person}{Pranav Shyam},
  \bibinfo{person}{Girish Sastry}, \bibinfo{person}{Amanda Askell},
  {et~al\mbox{.}}} \bibinfo{year}{2020}\natexlab{}.
\newblock \showarticletitle{Language models are few-shot learners}.
\newblock \bibinfo{journal}{\emph{arXiv preprint arXiv:2005.14165}}
  (\bibinfo{year}{2020}).
\newblock


\bibitem[\protect\citeauthoryear{Cava, Orzechowski, Burlacu, de~Franca,
  Virgolin, Jin, Kommenda, and Moore}{Cava et~al\mbox{.}}{2021}]%
        {cava2021contemporary}
\bibfield{author}{\bibinfo{person}{William~La Cava}, \bibinfo{person}{Patryk
  Orzechowski}, \bibinfo{person}{Bogdan Burlacu},
  \bibinfo{person}{Fabricio~Olivetti de Franca}, \bibinfo{person}{Marco
  Virgolin}, \bibinfo{person}{Ying Jin}, \bibinfo{person}{Michael Kommenda},
  {and} \bibinfo{person}{Jason~H. Moore}.} \bibinfo{year}{2021}\natexlab{}.
\newblock \showarticletitle{Contemporary Symbolic Regression Methods and their
  Relative Performance}. In \bibinfo{booktitle}{\emph{Advances in Neural
  Information Processing Systems Datasets and Benchmarks Track}}.
\newblock


\bibitem[\protect\citeauthoryear{Cava, Singh, Taggart, Suri, and Moore}{Cava
  et~al\mbox{.}}{2019}]%
        {cava2018learning}
\bibfield{author}{\bibinfo{person}{William~La Cava}, \bibinfo{person}{Tilak~Raj
  Singh}, \bibinfo{person}{James Taggart}, \bibinfo{person}{Srinivas Suri},
  {and} \bibinfo{person}{Jason Moore}.} \bibinfo{year}{2019}\natexlab{}.
\newblock \showarticletitle{Learning concise representations for regression by
  evolving networks of trees}. In \bibinfo{booktitle}{\emph{International
  Conference on Learning Representations}}.
\newblock


\bibitem[\protect\citeauthoryear{Chen, Xue, and Zhang}{Chen
  et~al\mbox{.}}{2019}]%
        {chen_improving_2019}
\bibfield{author}{\bibinfo{person}{Q. Chen}, \bibinfo{person}{B. Xue}, {and}
  \bibinfo{person}{M. Zhang}.} \bibinfo{year}{2019}\natexlab{}.
\newblock \showarticletitle{Improving {Generalization} of {Genetic}
  {Programming} for {Symbolic} {Regression} {With} {Angle}-{Driven} {Geometric}
  {Semantic} {Operators}}.
\newblock \bibinfo{journal}{\emph{IEEE Transactions on Evolutionary
  Computation}} \bibinfo{volume}{23}, \bibinfo{number}{3} (\bibinfo{date}{June}
  \bibinfo{year}{2019}), \bibinfo{pages}{488--502}.
\newblock


\bibitem[\protect\citeauthoryear{Chen and Guestrin}{Chen and Guestrin}{2016}]%
        {chen2016xgboost}
\bibfield{author}{\bibinfo{person}{Tianqi Chen} {and} \bibinfo{person}{Carlos
  Guestrin}.} \bibinfo{year}{2016}\natexlab{}.
\newblock \showarticletitle{Xgboost: A scalable tree boosting system}. In
  \bibinfo{booktitle}{\emph{Proceedings of the 22nd acm sigkdd international
  conference on knowledge discovery and data mining}}.
  \bibinfo{pages}{785--794}.
\newblock


\bibitem[\protect\citeauthoryear{Christensen and Oppacher}{Christensen and
  Oppacher}{2002}]%
        {koza2002benchmark}
\bibfield{author}{\bibinfo{person}{Steffen Christensen} {and}
  \bibinfo{person}{Franz Oppacher}.} \bibinfo{year}{2002}\natexlab{}.
\newblock \showarticletitle{An Analysis of Koza's Computational Effort
  Statistic for Genetic Programming}. In \bibinfo{booktitle}{\emph{Genetic
  Programming}}, \bibfield{editor}{\bibinfo{person}{James~A. Foster},
  \bibinfo{person}{Evelyne Lutton}, \bibinfo{person}{Julian Miller},
  \bibinfo{person}{Conor Ryan}, {and} \bibinfo{person}{Andrea Tettamanzi}}
  (Eds.). \bibinfo{publisher}{Springer Berlin Heidelberg},
  \bibinfo{address}{Berlin, Heidelberg}, \bibinfo{pages}{182--191}.
\newblock


\bibitem[\protect\citeauthoryear{Cortes and Vapnik}{Cortes and Vapnik}{1995}]%
        {cortes1995support}
\bibfield{author}{\bibinfo{person}{Corinna Cortes} {and}
  \bibinfo{person}{Vladimir Vapnik}.} \bibinfo{year}{1995}\natexlab{}.
\newblock \showarticletitle{Support-vector networks}.
\newblock \bibinfo{journal}{\emph{Machine learning}} \bibinfo{volume}{20},
  \bibinfo{number}{3} (\bibinfo{year}{1995}), \bibinfo{pages}{273--297}.
\newblock


\bibitem[\protect\citeauthoryear{Cranmer, Sanchez~Gonzalez, Battaglia, Xu,
  Cranmer, Spergel, and Ho}{Cranmer et~al\mbox{.}}{2020}]%
        {cranmer2020discovering}
\bibfield{author}{\bibinfo{person}{Miles Cranmer}, \bibinfo{person}{Alvaro
  Sanchez~Gonzalez}, \bibinfo{person}{Peter Battaglia}, \bibinfo{person}{Rui
  Xu}, \bibinfo{person}{Kyle Cranmer}, \bibinfo{person}{David Spergel}, {and}
  \bibinfo{person}{Shirley Ho}.} \bibinfo{year}{2020}\natexlab{}.
\newblock \showarticletitle{Discovering Symbolic Models from Deep Learning with
  Inductive Biases}. In \bibinfo{booktitle}{\emph{Advances in Neural
  Information Processing Systems}}, Vol.~\bibinfo{volume}{33}.
  \bibinfo{publisher}{Curran Associates, Inc.}, \bibinfo{pages}{17429--17442}.
\newblock


\bibitem[\protect\citeauthoryear{Dawood}{Dawood}{2011}]%
        {dawood2011theories}
\bibfield{author}{\bibinfo{person}{Hend Dawood}.}
  \bibinfo{year}{2011}\natexlab{}.
\newblock \bibinfo{booktitle}{\emph{Theories of interval arithmetic:
  mathematical foundations and applications}}.
\newblock \bibinfo{publisher}{LAP Lambert Academic Publishing}.
\newblock


\bibitem[\protect\citeauthoryear{Devlin, Chang, Lee, and Toutanova}{Devlin
  et~al\mbox{.}}{2019}]%
        {devlin-etal-2019-bert}
\bibfield{author}{\bibinfo{person}{Jacob Devlin}, \bibinfo{person}{Ming-Wei
  Chang}, \bibinfo{person}{Kenton Lee}, {and} \bibinfo{person}{Kristina
  Toutanova}.} \bibinfo{year}{2019}\natexlab{}.
\newblock \showarticletitle{BERT: Pre-training of Deep Bidirectional
  Transformers for Language Understanding}. In
  \bibinfo{booktitle}{\emph{NAACL}}. \bibinfo{publisher}{Association for
  Computational Linguistics}, \bibinfo{pages}{4171--4186}.
\newblock


\bibitem[\protect\citeauthoryear{Ferreira}{Ferreira}{2001}]%
        {ferreira2001gene}
\bibfield{author}{\bibinfo{person}{Candida Ferreira}.}
  \bibinfo{year}{2001}\natexlab{}.
\newblock \showarticletitle{Gene Expression Programming: a New Adaptive
  Algorithm for Solving Problems}.
\newblock \bibinfo{journal}{\emph{Complex Systems}} \bibinfo{volume}{13},
  \bibinfo{number}{2} (\bibinfo{year}{2001}), \bibinfo{pages}{87--129}.
\newblock


\bibitem[\protect\citeauthoryear{Fletcher}{Fletcher}{2013}]%
        {fletcher2013practical}
\bibfield{author}{\bibinfo{person}{Roger Fletcher}.}
  \bibinfo{year}{2013}\natexlab{}.
\newblock \bibinfo{booktitle}{\emph{Practical methods of optimization}}.
\newblock \bibinfo{publisher}{John Wiley \& Sons}.
\newblock


\bibitem[\protect\citeauthoryear{Holmes}{Holmes}{2009}]%
        {holmes2009introduction}
\bibfield{author}{\bibinfo{person}{Mark~H Holmes}.}
  \bibinfo{year}{2009}\natexlab{}.
\newblock \showarticletitle{Introduction to the foundations of applied
  mathematics}.
\newblock  (\bibinfo{year}{2009}).
\newblock


\bibitem[\protect\citeauthoryear{Jeffreys, Jeffreys, and Swirles}{Jeffreys
  et~al\mbox{.}}{1999}]%
        {jeffreys1999methods}
\bibfield{author}{\bibinfo{person}{Harold Jeffreys}, \bibinfo{person}{Bertha
  Jeffreys}, {and} \bibinfo{person}{Bertha Swirles}.}
  \bibinfo{year}{1999}\natexlab{}.
\newblock \bibinfo{booktitle}{\emph{Methods of mathematical physics}}.
\newblock \bibinfo{publisher}{Cambridge university press}.
\newblock


\bibitem[\protect\citeauthoryear{Jin, Fu, Kang, Guo, and Guo}{Jin
  et~al\mbox{.}}{2019}]%
        {jin2019bayesian}
\bibfield{author}{\bibinfo{person}{Ying Jin}, \bibinfo{person}{Weilin Fu},
  \bibinfo{person}{Jian Kang}, \bibinfo{person}{Jiadong Guo}, {and}
  \bibinfo{person}{Jian Guo}.} \bibinfo{year}{2019}\natexlab{}.
\newblock \showarticletitle{Bayesian symbolic regression}.
\newblock \bibinfo{journal}{\emph{arXiv preprint arXiv:1910.08892}}
  (\bibinfo{year}{2019}).
\newblock


\bibitem[\protect\citeauthoryear{Ke, Meng, Finley, Wang, Chen, Ma, Ye, and
  Liu}{Ke et~al\mbox{.}}{2017}]%
        {ke2017lightgbm}
\bibfield{author}{\bibinfo{person}{Guolin Ke}, \bibinfo{person}{Qi Meng},
  \bibinfo{person}{Thomas Finley}, \bibinfo{person}{Taifeng Wang},
  \bibinfo{person}{Wei Chen}, \bibinfo{person}{Weidong Ma},
  \bibinfo{person}{Qiwei Ye}, {and} \bibinfo{person}{Tie-Yan Liu}.}
  \bibinfo{year}{2017}\natexlab{}.
\newblock \showarticletitle{Lightgbm: A highly efficient gradient boosting
  decision tree}.
\newblock \bibinfo{journal}{\emph{Advances in neural information processing
  systems}}  \bibinfo{volume}{30} (\bibinfo{year}{2017}),
  \bibinfo{pages}{3146--3154}.
\newblock


\bibitem[\protect\citeauthoryear{Keijzer}{Keijzer}{2003}]%
        {keijzer2003improving}
\bibfield{author}{\bibinfo{person}{Maarten Keijzer}.}
  \bibinfo{year}{2003}\natexlab{}.
\newblock \showarticletitle{Improving symbolic regression with interval
  arithmetic and linear scaling}. In \bibinfo{booktitle}{\emph{European
  Conference on Genetic Programming}}. Springer, \bibinfo{pages}{70--82}.
\newblock


\bibitem[\protect\citeauthoryear{Kennedy and Eberhart}{Kennedy and
  Eberhart}{1995}]%
        {kennedy1995particle}
\bibfield{author}{\bibinfo{person}{James Kennedy} {and}
  \bibinfo{person}{Russell Eberhart}.} \bibinfo{year}{1995}\natexlab{}.
\newblock \showarticletitle{Particle swarm optimization}. In
  \bibinfo{booktitle}{\emph{Proceedings of ICNN'95-international conference on
  neural networks}}, Vol.~\bibinfo{volume}{4}. IEEE,
  \bibinfo{pages}{1942--1948}.
\newblock


\bibitem[\protect\citeauthoryear{Kim, Lu, Mukherjee, Gilbert, Jing,
  {\v{C}}eperi{\'c}, and Solja{\v{c}}i{\'c}}{Kim et~al\mbox{.}}{2020}]%
        {kim2020integration}
\bibfield{author}{\bibinfo{person}{Samuel Kim}, \bibinfo{person}{Peter~Y Lu},
  \bibinfo{person}{Srijon Mukherjee}, \bibinfo{person}{Michael Gilbert},
  \bibinfo{person}{Li Jing}, \bibinfo{person}{Vladimir {\v{C}}eperi{\'c}},
  {and} \bibinfo{person}{Marin Solja{\v{c}}i{\'c}}.}
  \bibinfo{year}{2020}\natexlab{}.
\newblock \showarticletitle{Integration of neural network-based symbolic
  regression in deep learning for scientific discovery}.
\newblock \bibinfo{journal}{\emph{IEEE Transactions on Neural Networks and
  Learning Systems}} (\bibinfo{year}{2020}).
\newblock


\bibitem[\protect\citeauthoryear{Kingma and Welling}{Kingma and
  Welling}{2013}]%
        {kingma2013auto}
\bibfield{author}{\bibinfo{person}{Diederik~P Kingma} {and}
  \bibinfo{person}{Max Welling}.} \bibinfo{year}{2013}\natexlab{}.
\newblock \showarticletitle{Auto-encoding variational bayes}.
\newblock \bibinfo{journal}{\emph{arXiv preprint arXiv:1312.6114}}
  (\bibinfo{year}{2013}).
\newblock


\bibitem[\protect\citeauthoryear{Korns}{Korns}{2011}]%
        {korns2011accuracy}
\bibfield{author}{\bibinfo{person}{Michael~F Korns}.}
  \bibinfo{year}{2011}\natexlab{}.
\newblock \showarticletitle{Accuracy in symbolic regression}.
\newblock In \bibinfo{booktitle}{\emph{Genetic Programming Theory and Practice
  IX}}. \bibinfo{publisher}{Springer}, \bibinfo{pages}{129--151}.
\newblock


\bibitem[\protect\citeauthoryear{Korns}{Korns}{2013}]%
        {korns2013baseline}
\bibfield{author}{\bibinfo{person}{Michael~F Korns}.}
  \bibinfo{year}{2013}\natexlab{}.
\newblock \showarticletitle{A baseline symbolic regression algorithm}.
\newblock In \bibinfo{booktitle}{\emph{Genetic Programming Theory and Practice
  X}}. \bibinfo{publisher}{Springer}, \bibinfo{pages}{117--137}.
\newblock


\bibitem[\protect\citeauthoryear{{Koza} and {John R.}}{{Koza} and {John
  R.}}{1994}]%
        {koza_genetic_1994}
\bibfield{author}{\bibinfo{person}{{Koza}} {and} \bibinfo{person}{{John R.}}}
  \bibinfo{year}{1994}\natexlab{}.
\newblock \showarticletitle{Genetic Programming as a Means for Programming
  Computers by Natural Selection}.
\newblock \bibinfo{journal}{\emph{Statistics and Computing}}
  \bibinfo{volume}{4}, \bibinfo{number}{2} (\bibinfo{date}{June}
  \bibinfo{year}{1994}), \bibinfo{pages}{87--112}.
\newblock
\showISSN{1573-1375}


\bibitem[\protect\citeauthoryear{Krawiec}{Krawiec}{2002}]%
        {krawiec2002genetic}
\bibfield{author}{\bibinfo{person}{Krzysztof Krawiec}.}
  \bibinfo{year}{2002}\natexlab{}.
\newblock \showarticletitle{Genetic programming-based construction of features
  for machine learning and knowledge discovery tasks}.
\newblock \bibinfo{journal}{\emph{Genetic Programming and Evolvable Machines}}
  \bibinfo{volume}{3}, \bibinfo{number}{4} (\bibinfo{year}{2002}),
  \bibinfo{pages}{329--343}.
\newblock


\bibitem[\protect\citeauthoryear{Kusner, Paige, and
  Hern{\'a}ndez-Lobato}{Kusner et~al\mbox{.}}{2017}]%
        {kusner2017grammar}
\bibfield{author}{\bibinfo{person}{Matt~J Kusner}, \bibinfo{person}{Brooks
  Paige}, {and} \bibinfo{person}{Jos{\'e}~Miguel Hern{\'a}ndez-Lobato}.}
  \bibinfo{year}{2017}\natexlab{}.
\newblock \showarticletitle{Grammar variational autoencoder}. In
  \bibinfo{booktitle}{\emph{International Conference on Machine Learning}}.
  PMLR, \bibinfo{pages}{1945--1954}.
\newblock


\bibitem[\protect\citeauthoryear{La~Cava and Moore}{La~Cava and Moore}{2017}]%
        {la2017general}
\bibfield{author}{\bibinfo{person}{William La~Cava} {and}
  \bibinfo{person}{Jason Moore}.} \bibinfo{year}{2017}\natexlab{}.
\newblock \showarticletitle{A General Feature Engineering Wrapper for Machine
  Learning Using $\epsilon$-Lexicase Survival}. In
  \bibinfo{booktitle}{\emph{European Conference on Genetic Programming}}.
  Springer, \bibinfo{pages}{80--95}.
\newblock


\bibitem[\protect\citeauthoryear{LeCun, Bengio, and Hinton}{LeCun
  et~al\mbox{.}}{2015}]%
        {lecun2015deep}
\bibfield{author}{\bibinfo{person}{Yann LeCun}, \bibinfo{person}{Yoshua
  Bengio}, {and} \bibinfo{person}{Geoffrey Hinton}.}
  \bibinfo{year}{2015}\natexlab{}.
\newblock \showarticletitle{Deep learning}.
\newblock \bibinfo{journal}{\emph{nature}} \bibinfo{volume}{521},
  \bibinfo{number}{7553} (\bibinfo{year}{2015}), \bibinfo{pages}{436--444}.
\newblock


\bibitem[\protect\citeauthoryear{Lu, Zhou, Tao, Luo, and Wang}{Lu
  et~al\mbox{.}}{2021}]%
        {SPJ-GEP2021Lu}
\bibfield{author}{\bibinfo{person}{Qiang Lu}, \bibinfo{person}{Shuo Zhou},
  \bibinfo{person}{Fan Tao}, \bibinfo{person}{Jake Luo}, {and}
  \bibinfo{person}{Zhiguang Wang}.} \bibinfo{year}{2021}\natexlab{}.
\newblock \showarticletitle{Enhancing gene expression programming based on
  space partition and jump for symbolic regression}.
\newblock \bibinfo{journal}{\emph{Information Sciences}}  \bibinfo{volume}{547}
  (\bibinfo{year}{2021}), \bibinfo{pages}{553--567}.
\newblock


\bibitem[\protect\citeauthoryear{Martius and Lampert}{Martius and
  Lampert}{2016}]%
        {martius2016extrapolation}
\bibfield{author}{\bibinfo{person}{Georg Martius} {and}
  \bibinfo{person}{Christoph~H Lampert}.} \bibinfo{year}{2016}\natexlab{}.
\newblock \showarticletitle{Extrapolation and learning equations}.
\newblock \bibinfo{journal}{\emph{arXiv preprint arXiv:1610.02995}}
  (\bibinfo{year}{2016}).
\newblock


\bibitem[\protect\citeauthoryear{McConaghy}{McConaghy}{2011}]%
        {mcconaghy2011ffx}
\bibfield{author}{\bibinfo{person}{Trent McConaghy}.}
  \bibinfo{year}{2011}\natexlab{}.
\newblock \showarticletitle{FFX: Fast, scalable, deterministic symbolic
  regression technology}.
\newblock In \bibinfo{booktitle}{\emph{Genetic Programming Theory and Practice
  IX}}. \bibinfo{publisher}{Springer}, \bibinfo{pages}{235--260}.
\newblock


\bibitem[\protect\citeauthoryear{McDermott, White, Luke, Manzoni, Castelli,
  Vanneschi, Jaskowski, Krawiec, Harper, De~Jong, and O'Reilly}{McDermott
  et~al\mbox{.}}{2012}]%
        {mcdermott12benchmark}
\bibfield{author}{\bibinfo{person}{James McDermott}, \bibinfo{person}{David~R.
  White}, \bibinfo{person}{Sean Luke}, \bibinfo{person}{Luca Manzoni},
  \bibinfo{person}{Mauro Castelli}, \bibinfo{person}{Leonardo Vanneschi},
  \bibinfo{person}{Wojciech Jaskowski}, \bibinfo{person}{Krzysztof Krawiec},
  \bibinfo{person}{Robin Harper}, \bibinfo{person}{Kenneth De~Jong}, {and}
  \bibinfo{person}{Una-May O'Reilly}.} \bibinfo{year}{2012}\natexlab{}.
\newblock \showarticletitle{Genetic Programming Needs Better Benchmarks}
  \emph{(\bibinfo{series}{GECCO '12})}. \bibinfo{publisher}{Association for
  Computing Machinery}, \bibinfo{address}{New York, NY, USA},
  \bibinfo{pages}{791–798}.
\newblock


\bibitem[\protect\citeauthoryear{McKay, Hoai, Whigham, Shan, and O'Neill}{McKay
  et~al\mbox{.}}{2010}]%
        {mckay_grammar-based_2010}
\bibfield{author}{\bibinfo{person}{Robert~I. McKay},
  \bibinfo{person}{Nguyen~Xuan Hoai}, \bibinfo{person}{Peter~Alexander
  Whigham}, \bibinfo{person}{Yin Shan}, {and} \bibinfo{person}{Michael
  O'Neill}.} \bibinfo{year}{2010}\natexlab{}.
\newblock \showarticletitle{Grammar-based {Genetic} {Programming}: a survey}.
\newblock \bibinfo{journal}{\emph{Genetic Programming and Evolvable Machines}}
  \bibinfo{volume}{11}, \bibinfo{number}{3} (\bibinfo{date}{Sept.}
  \bibinfo{year}{2010}), \bibinfo{pages}{365--396}.
\newblock
\showISSN{1573-7632}


\bibitem[\protect\citeauthoryear{Miller}{Miller}{2019}]%
        {miller_cartesian_2019}
\bibfield{author}{\bibinfo{person}{Julian~Francis Miller}.}
  \bibinfo{year}{2019}\natexlab{}.
\newblock \showarticletitle{Cartesian genetic programming: its status and
  future}.
\newblock \bibinfo{journal}{\emph{Genetic Programming and Evolvable Machines}}
  (\bibinfo{date}{Aug.} \bibinfo{year}{2019}), \bibinfo{pages}{1--40}.
\newblock
\showISSN{1573-7632}


\bibitem[\protect\citeauthoryear{Miller and Harding}{Miller and
  Harding}{2008}]%
        {CGP2008Miller}
\bibfield{author}{\bibinfo{person}{Julian~Francis Miller} {and}
  \bibinfo{person}{Simon~L. Harding}.} \bibinfo{year}{2008}\natexlab{}.
\newblock \showarticletitle{Cartesian {Genetic} {Programming}}. In
  \bibinfo{booktitle}{\emph{Proceedings of the 10th {Annual} {Conference}
  {Companion} on {Genetic} and {Evolutionary} {Computation}}}
  \emph{(\bibinfo{series}{{GECCO} '08})}. \bibinfo{publisher}{ACM},
  \bibinfo{address}{New York, NY, USA}, \bibinfo{pages}{2701--2726}.
\newblock


\bibitem[\protect\citeauthoryear{Moraglio, Krawiec, and Johnson}{Moraglio
  et~al\mbox{.}}{2012}]%
        {hutchison_geometric_2012}
\bibfield{author}{\bibinfo{person}{Alberto Moraglio},
  \bibinfo{person}{Krzysztof Krawiec}, {and} \bibinfo{person}{Colin~G.
  Johnson}.} \bibinfo{year}{2012}\natexlab{}.
\newblock \showarticletitle{Geometric {Semantic} {Genetic} {Programming}}.
\newblock In \bibinfo{booktitle}{\emph{Parallel {Problem} {Solving} from
  {Nature} - {PPSN} {XII}}}. Vol.~\bibinfo{volume}{7491}.
  \bibinfo{publisher}{Springer Berlin Heidelberg}, \bibinfo{address}{Berlin,
  Heidelberg}, \bibinfo{pages}{21--31}.
\newblock


\bibitem[\protect\citeauthoryear{Mundhenk, Landajuela, Glatt, Santiago,
  Faissol, and Petersen}{Mundhenk et~al\mbox{.}}{2021}]%
        {mundhenk2021symbolic}
\bibfield{author}{\bibinfo{person}{T~Nathan Mundhenk}, \bibinfo{person}{Mikel
  Landajuela}, \bibinfo{person}{Ruben Glatt}, \bibinfo{person}{Claudio~P
  Santiago}, \bibinfo{person}{Daniel~M Faissol}, {and}
  \bibinfo{person}{Brenden~K Petersen}.} \bibinfo{year}{2021}\natexlab{}.
\newblock \showarticletitle{Symbolic Regression via Neural-Guided Genetic
  Programming Population Seeding}. In \bibinfo{booktitle}{\emph{Advances in
  Neural Information Processing Systems}}.
\newblock


\bibitem[\protect\citeauthoryear{Mu{\~n}oz, Trujillo, Silva, Castelli, and
  Vanneschi}{Mu{\~n}oz et~al\mbox{.}}{2019}]%
        {munoz2019evolving}
\bibfield{author}{\bibinfo{person}{Luis Mu{\~n}oz}, \bibinfo{person}{Leonardo
  Trujillo}, \bibinfo{person}{Sara Silva}, \bibinfo{person}{Mauro Castelli},
  {and} \bibinfo{person}{Leonardo Vanneschi}.} \bibinfo{year}{2019}\natexlab{}.
\newblock \showarticletitle{Evolving multidimensional transformations for
  symbolic regression with M3GP}.
\newblock \bibinfo{journal}{\emph{Memetic Computing}} \bibinfo{volume}{11},
  \bibinfo{number}{2} (\bibinfo{year}{2019}), \bibinfo{pages}{111--126}.
\newblock


\bibitem[\protect\citeauthoryear{Olson, La~Cava, Orzechowski, Urbanowicz, and
  Moore}{Olson et~al\mbox{.}}{2017}]%
        {Olson2017PMLB}
\bibfield{author}{\bibinfo{person}{Randal~S. Olson}, \bibinfo{person}{William
  La~Cava}, \bibinfo{person}{Patryk Orzechowski}, \bibinfo{person}{Ryan~J.
  Urbanowicz}, {and} \bibinfo{person}{Jason~H. Moore}.}
  \bibinfo{year}{2017}\natexlab{}.
\newblock \showarticletitle{PMLB: a large benchmark suite for machine learning
  evaluation and comparison}.
\newblock \bibinfo{journal}{\emph{BioData Mining}} \bibinfo{volume}{10},
  \bibinfo{number}{36} (\bibinfo{date}{11 Dec} \bibinfo{year}{2017}),
  \bibinfo{pages}{1--13}.
\newblock
\showISSN{1756-0381}


\bibitem[\protect\citeauthoryear{Petersen, Larma, Mundhenk, Santiago, Kim, and
  Kim}{Petersen et~al\mbox{.}}{2021}]%
        {petersen2021deep}
\bibfield{author}{\bibinfo{person}{Brenden~K Petersen},
  \bibinfo{person}{Mikel~Landajuela Larma}, \bibinfo{person}{Terrell~N
  Mundhenk}, \bibinfo{person}{Claudio~Prata Santiago},
  \bibinfo{person}{Soo~Kyung Kim}, {and} \bibinfo{person}{Joanne~Taery Kim}.}
  \bibinfo{year}{2021}\natexlab{}.
\newblock \showarticletitle{Deep symbolic regression: Recovering mathematical
  expressions from data via risk-seeking policy gradients}. In
  \bibinfo{booktitle}{\emph{International Conference on Learning
  Representations}}.
\newblock


\bibitem[\protect\citeauthoryear{Sahoo, Lampert, and Martius}{Sahoo
  et~al\mbox{.}}{2018}]%
        {sahoo2018learning}
\bibfield{author}{\bibinfo{person}{Subham Sahoo}, \bibinfo{person}{Christoph
  Lampert}, {and} \bibinfo{person}{Georg Martius}.}
  \bibinfo{year}{2018}\natexlab{}.
\newblock \showarticletitle{Learning equations for extrapolation and control}.
  In \bibinfo{booktitle}{\emph{International Conference on Machine Learning}}.
  PMLR, \bibinfo{pages}{4442--4450}.
\newblock


\bibitem[\protect\citeauthoryear{Schmidt and Lipson}{Schmidt and
  Lipson}{2009}]%
        {schmidt_distilling_2009}
\bibfield{author}{\bibinfo{person}{Michael Schmidt} {and} \bibinfo{person}{Hod
  Lipson}.} \bibinfo{year}{2009}\natexlab{}.
\newblock \showarticletitle{Distilling {Free}-{Form} {Natural} {Laws} from
  {Experimental} {Data}}.
\newblock \bibinfo{journal}{\emph{Science}} \bibinfo{volume}{324},
  \bibinfo{number}{5923} (\bibinfo{year}{2009}), \bibinfo{pages}{81--85}.
\newblock


\bibitem[\protect\citeauthoryear{Trujillo, Mu{\~n}oz, Galv{\'a}n-L{\'o}pez, and
  Silva}{Trujillo et~al\mbox{.}}{2016}]%
        {trujillo2016neat}
\bibfield{author}{\bibinfo{person}{Leonardo Trujillo}, \bibinfo{person}{Luis
  Mu{\~n}oz}, \bibinfo{person}{Edgar Galv{\'a}n-L{\'o}pez}, {and}
  \bibinfo{person}{Sara Silva}.} \bibinfo{year}{2016}\natexlab{}.
\newblock \showarticletitle{neat genetic programming: Controlling bloat
  naturally}.
\newblock \bibinfo{journal}{\emph{Information Sciences}}  \bibinfo{volume}{333}
  (\bibinfo{year}{2016}), \bibinfo{pages}{21--43}.
\newblock


\bibitem[\protect\citeauthoryear{Udrescu, Tan, Feng, Neto, Wu, and
  Tegmark}{Udrescu et~al\mbox{.}}{2020}]%
        {udrescu2020ai_b}
\bibfield{author}{\bibinfo{person}{Silviu-Marian Udrescu},
  \bibinfo{person}{Andrew Tan}, \bibinfo{person}{Jiahai Feng},
  \bibinfo{person}{Orisvaldo Neto}, \bibinfo{person}{Tailin Wu}, {and}
  \bibinfo{person}{Max Tegmark}.} \bibinfo{year}{2020}\natexlab{}.
\newblock \showarticletitle{AI Feynman 2.0: Pareto-optimal symbolic regression
  exploiting graph modularity}.
\newblock \bibinfo{journal}{\emph{arXiv preprint arXiv:2006.10782}}
  (\bibinfo{year}{2020}).
\newblock


\bibitem[\protect\citeauthoryear{Udrescu and Tegmark}{Udrescu and
  Tegmark}{2020}]%
        {udrescu2020ai_a}
\bibfield{author}{\bibinfo{person}{Silviu-Marian Udrescu} {and}
  \bibinfo{person}{Max Tegmark}.} \bibinfo{year}{2020}\natexlab{}.
\newblock \showarticletitle{AI Feynman: A physics-inspired method for symbolic
  regression}.
\newblock \bibinfo{journal}{\emph{Science Advances}} \bibinfo{volume}{6},
  \bibinfo{number}{16} (\bibinfo{year}{2020}), \bibinfo{pages}{eaay2631}.
\newblock


\bibitem[\protect\citeauthoryear{Uy, Hoai, O’Neill, McKay, and
  Galv{\'a}n-L{\'o}pez}{Uy et~al\mbox{.}}{2011}]%
        {uy2011semantically}
\bibfield{author}{\bibinfo{person}{Nguyen~Quang Uy},
  \bibinfo{person}{Nguyen~Xuan Hoai}, \bibinfo{person}{Michael O’Neill},
  \bibinfo{person}{Robert~I McKay}, {and} \bibinfo{person}{Edgar
  Galv{\'a}n-L{\'o}pez}.} \bibinfo{year}{2011}\natexlab{}.
\newblock \showarticletitle{Semantically-based crossover in genetic
  programming: application to real-valued symbolic regression}.
\newblock \bibinfo{journal}{\emph{Genetic Programming and Evolvable Machines}}
  \bibinfo{volume}{12}, \bibinfo{number}{2} (\bibinfo{year}{2011}),
  \bibinfo{pages}{91--119}.
\newblock


\bibitem[\protect\citeauthoryear{Vladislavleva, Smits, and
  Den~Hertog}{Vladislavleva et~al\mbox{.}}{2008}]%
        {vladislavleva2008order}
\bibfield{author}{\bibinfo{person}{Ekaterina~J Vladislavleva},
  \bibinfo{person}{Guido~F Smits}, {and} \bibinfo{person}{Dick Den~Hertog}.}
  \bibinfo{year}{2008}\natexlab{}.
\newblock \showarticletitle{Order of nonlinearity as a complexity measure for
  models generated by symbolic regression via pareto genetic programming}.
\newblock \bibinfo{journal}{\emph{IEEE Transactions on Evolutionary
  Computation}} \bibinfo{volume}{13}, \bibinfo{number}{2}
  (\bibinfo{year}{2008}), \bibinfo{pages}{333--349}.
\newblock


\bibitem[\protect\citeauthoryear{Weisberg}{Weisberg}{2005}]%
        {weisberg2005applied}
\bibfield{author}{\bibinfo{person}{Sanford Weisberg}.}
  \bibinfo{year}{2005}\natexlab{}.
\newblock \bibinfo{booktitle}{\emph{Applied linear regression}}.
  Vol.~\bibinfo{volume}{528}.
\newblock \bibinfo{publisher}{John Wiley \& Sons}.
\newblock


\bibitem[\protect\citeauthoryear{Worm and Chiu}{Worm and Chiu}{2013}]%
        {worm2013prioritized}
\bibfield{author}{\bibinfo{person}{Tony Worm} {and} \bibinfo{person}{Kenneth
  Chiu}.} \bibinfo{year}{2013}\natexlab{}.
\newblock \showarticletitle{Prioritized grammar enumeration: symbolic
  regression by dynamic programming}. In \bibinfo{booktitle}{\emph{Proceedings
  of the 15th annual conference on Genetic and evolutionary computation}}.
  \bibinfo{pages}{1021--1028}.
\newblock


\bibitem[\protect\citeauthoryear{Xing, Salleb-Aouissi, and Verma}{Xing
  et~al\mbox{.}}{2021}]%
        {xing2021automated}
\bibfield{author}{\bibinfo{person}{Hengrui Xing}, \bibinfo{person}{Ansaf
  Salleb-Aouissi}, {and} \bibinfo{person}{Nakul Verma}.}
  \bibinfo{year}{2021}\natexlab{}.
\newblock \showarticletitle{Automated Symbolic Law Discovery: A Computer Vision
  Approach}. In \bibinfo{booktitle}{\emph{Proceedings of the AAAI Conference on
  Artificial Intelligence}}, Vol.~\bibinfo{volume}{35}.
  \bibinfo{pages}{660--668}.
\newblock


\bibitem[\protect\citeauthoryear{Zhong, Lin, Lu, and Huang}{Zhong
  et~al\mbox{.}}{2018}]%
        {zhong2018deep}
\bibfield{author}{\bibinfo{person}{Jinghui Zhong}, \bibinfo{person}{Yusen Lin},
  \bibinfo{person}{Chengyu Lu}, {and} \bibinfo{person}{Zhixing Huang}.}
  \bibinfo{year}{2018}\natexlab{}.
\newblock \showarticletitle{A deep learning assisted gene expression
  programming framework for symbolic regression problems}. In
  \bibinfo{booktitle}{\emph{International Conference on Neural Information
  Processing}}. Springer, \bibinfo{pages}{530--541}.
\newblock


\end{thebibliography}

\end{document}


\section{Appendix}

\begin{table*}[htbp]
\caption{Classical Symbolic Regression Benchmarks(SRB).}
\label{SRBBenchmarkdetail}
\begin{tabular}{llll}
\toprule
FileNumber & FileName    & Objective Function & Data Set \\
\midrule
F1 & Keijzer-5 & $log(x)$ & U{[}0,2,20{]}\\
F2 & Nguyen-8& $sqrt(x)$& U{[}0,2,20{]}\\
F3 & Korns-1 & $1.57 + 24.3x$& U{[}-1,1,20{]} \\
F4 & Korns-6 & $6.87 + 11  cos(7.23x^{3})$ & U{[}-0.5,0.5,20{]} \\
F5 & Nguyen-4& $x^{6} + x^{5} + x^{4} + x^{3} + x^{2} + x$ & U{[}-1,1,20{]} \\
F6 & Nguyen-3& $x^{5} + x^{4} + x^{3} + x^{2} + x$& U{[}-1,1,20{]} \\
F7 & Koza-1,Nguyen-2 & $x^{4} + x^{3} + x^{2} + x$ & U{[}-1,1,20{]} \\
F8 & Nguyen-1& $x^{3} + x^{2} + x$& U{[}-1,1,20{]} \\
F9 & Koza-3& $x^{6} - 2x^{4} + x^{2}$ & U{[}-1,1,20{]} \\
F10& Koza-2& $x^{5} - 2x^{3} + x$& U{[}-1,1,20{]} \\
F11& Nguyen-5& $cos(x)sin(x^{2}) - 1$ & U{[}-1.6,1.6,20{]} \\
F12& Nguyen-6& $sin(x)+sin(x + x^{2})$ & U{[}-1,1,20{]} \\
F13& Nguyen-11 & $x^{y}$ & U{[}2,4,400{]} \\
F14& Keijzer-11& $xy + sin((x-1)(y-1))$ & U{[}-1,1,400{]}\\
F15& Nguyen-12 & $x^{4} - x^{3} +y^{2}/2- y$& U{[}-1,1,400{]}    \\
F16   & Keijzer-13 & $6sin(x)cos(y)$  & U{[}-1,1,400{]}    \\
F17   & Keijzer-15 & $x^{3}/5 + y^{3}/2 -y -x$ & U{[}-1,1,400{]}    \\
F18   & Nguyen-9   & $sin(x) + sin(y^{2})$    & U{[}-1,1,400{]}    \\
F19   & Nguyen-10  & $2sin(x)cos(y)$    & U{[}-1,1,400{]}\\
F20  & Vladislavleva-1 & $exp(-(x-1))^{2} /(1.2 + (y-2.5)^{2})$   & U{[}-1,1,400{]}\\
F21  & Keijzer-3 & $30xz/((x-10)y^{y})$& x,z:{[}-1,1,1000{]} y:U{[}1,3,1000{]}  \\
F22  & Korns-2   & $0.23 + 14.2 (x + y)/(3z)$     & x,y:U{[}-1,1,1000{]} z:U{[}1,3,1000{]} \\
F23  & Vladislavleva-5 & $30((x-1)(z-1)) / (y^{2}(x-10))$& U{[}0,2,1000{]}     \\
\bottomrule
\end{tabular}
\end{table*}

\begin{table*}[htbp]
\caption{Feynman Symbolic Regression Benchmarks(FSRB).}
\label{FSRBBenchmarkdetail}
\begin{tabular}{lllll}
			\toprule
FileNumber & FileName  & Output & Objective Function     & Data Set  \\
			\midrule
F24  & I.6.2a    & $f$& $exp(-\theta^{2}/2)/sqrt(2\pi )$  & U{[}1,3,20{]}   \\
F25  & I.6.2     & $f$& $exp(-(\theta/\sigma )^{2}/2)/(sqrt(2\pi )\sigma )$    & U{[}1,3,400{]}  \\
F26  & I.12.1    & $F$&$muN_{n}$& U{[}2,4,400{]}  \\
F27  & I.12.5    & $F$& $q_{2}Ef$& U{[}2,4,400{]}  \\
F28  & I.14.4    & $U$& $1/2k_{spring}x^{2}$     & U{[}2,4,400{]}  \\
F29  & I.25.13   & $Volt$   & $q/C$  & U{[}2,4,400{]}  \\
F30  & I.26.2    & $\theta_{1}$ & $arcsin(nsin(\theta_{2}))$  & $n$:U{[}0,1,400{]} $\theta_{2}$:U{[}2,4,400{]}  \\
F31  & I.29.4    & $k$& $\omega/c$    & U{[}2,4,400{]}  \\
F32  & I.34.27   & $E_{n}$   & $(h/(2\pi ))\omega$ & U{[}2,4,400{]}  \\
F33  & I.39.1    & $E_{n}$   & $3/2 p r V$   & U{[}2,4,400{]}  \\
F34  & II.3.24   & $flux$   & $Pw r/(4\pi r^{2})$  & U{[}2,4,400{]}  \\
F35  & II.8.31   & $E_{den}$ & $\epsilon$ ${Ef}^{2}/2$  & U{[}2,4,400{]}  \\
F36  & II.11.28  & $\theta$  & $1+n \alpha /(1-( n \alpha /3))$    & U{[}0,1,400{]}  \\
F37  & II.27.18  & $E_{den}$ & $\epsilon Ef^{2}$    & U{[}2,4,400{]}  \\
F38  & II.38.14  & $mu_{S}$  & $Y/(2(1+\sigma ))$  & U{[}2,4,400{]}  \\
F39  & III.12.43 & $L$ & $n(h/(2 \pi ))$     & U{[}2,4,400{]}  \\
F40  & II.37.1   & $E_{n}$   & $mom(1+chi)B$    & U{[}2,4,1000{]} \\
F41  & I.18.12   &$ tau $   & $rFsin(\theta)$   & U{[}2,4,1000{]} \\
F42  & I.6.2b    & $f$& $exp(-((\theta-\theta1)/\sigma )^{2}/2)/(sqrt(2\pi )\sigma )$ & U{[}1,3,1000{]} \\
F43  & I.10.7    & $m$& $m_{0}/sqrt(1-v^{2}/c^{2})$ & $m_{0}, c$:U{[}3,5,1000{]} $v$:U{[}1,2,1000{]}  \\
F44  & I.12.4    & $Ef$     & $q_{1}r/(4\pi \epsilon r^{3})$     & U{[}2,4,1000{]} \\
F45  & I.14.3    & $U$& $mgz$& U{[}2,4,1000{]} \\
F46  & I.15.1    & $p$& $m_{0} v/sqrt(1-v^{2}/c^{2})$     & $m0, c$:U{[}3,5,1000{]} $v$:U{[}1,2,1000{]}  \\
F47  & I.16.6    & $v_{1}$     & $(u+v)/(1+uv/c^{2})$     & U{[}2,4,1000{]} \\
F48  & I.27.6    & $foc$    & $1/(1/d_{1}+n/d_{2})$    & U{[}2,4,1000{]} \\
F49  & I.30.3    & $Int$    & $Int_{0} sin( n \theta /2)^{2} / sin(\theta /2)^{2}$ & U{[}2,4,1000{]} \\
F50  & I.30.5    & $\theta$  & $arcsin(\lambda /(n d))$    & $\lambda$ :U{[}1,2,1000{]} $d,n$:U{[}2,4,1000{]}     \\
F51  & I.34.1    & $\omega$  & $\omega_{0}/(1-v/c)$ & c, $\omega_{0}$ :U{[}3,5,1000{]} v:U{[}1,2,1000{]}  \\
F52  & I.34.14   & $\omega$  & $(1+v/c)/sqrt(1-v^{2}/c^{2})\omega_{0}$  & $c, \omega_{0}$ :U{[}3,5,1000{]} $v$:U{[}1,2,1000{]}  \\
F53  & I.37.4    & $Int$    & $I_{1}+I_{2}+2sqrt(I_{1}I_{2})cos(\delta )$     & U{[}2,4,1000{]} \\
F54  & I.39.11   & $E_{n}$   & $1/(\gamma -1)prV$ & U{[}2,4,1000{]} \\
F55  & I.43.31   & $D$& $mo b_{k} b_{T}$   & U{[}2,4,1000{]} \\
F56  & I.47.23   & $c$& $sqrt(\gamma pr/rho)$     & U{[}2,4,1000{]} \\
F57  & II.4.23   & $Volt$   & $q/(4\pi \epsilon r)$     & U{[}2,4,1000{]} \\
F58  & II.8.7    & $E_{n}$   & $3/5q^{2}/(4\pi \epsilon d)$    & U{[}2,4,1000{]} \\
F59  & II.10.9   & $Ef$     & $\sigma_{den}/\epsilon 1/(1+chi)$ & U{[}2,4,1000{]} \\
F60  & II.13.23  & $rho_{c}$ & $rho_{c0} /sqrt(1-v^{2}/c^{2})$  & $rho_{c0}, c$:U{[}3,5,1000{]} $v$:U{[}1,2,1000{]} \\
F61  & II.13.34  & $j$& $rho_{c0} v/sqrt(1-v^{2}/c^{2})$& $rho_{c0}, c$:U{[}3,5,1000{]} $v$:U{[}1,2,1000{]} \\
F62  & II.27.16  & $flux$   & $\epsilon cEf^{2}$  & U{[}2,4,1000{]} \\
F63  & II.34.2a  & $I$&$ qv/(2\pi r)$     & U{[}2,4,1000{]} \\
F64  & II.34.2   & $mom$    & $qvr/2$    & U{[}2,4,1000{]} \\
F65  & II.34.29a & $mom$    & $qh/(4\pi m)$     & U{[}2,4,1000{]} \\
F66  & III.7.38  & $\omega$  & $2 mom B/(h/(2\pi ))$     & U{[}2,4,1000{]} \\
F67  & III.8.54  & $prob$   & $sin(E_{n} t/(h/(2\pi )))^{2}$    & U{[}1,2,1000{]} \\
F68  & III.15.12 & $E_{n}$   & $2U(1-cos(kd))$ & U{[}2,4,1000{]} \\
F69  & II.15.4   & $E_{n}$   & $-mom B cos(\theta)$   & U{[}2,4,1000{]} \\
F70  & II.15.5   & $E_{n}$  & $-p_{d} E_{f}cos(\theta)$    & U{[}2,4,1000{]} \\
F71  & I.18.14   & $L$& $m r v sin(\theta)$ & U{[}2,4,4000{]}\\
\bottomrule
\end{tabular}
\end{table*}

\begin{table*}[htbp]
\caption{Penn Machine Learning Benchmarks(PMLB).}
\label{PMLBBenchmarkdetail}
\begin{tabular}{lllll}
\toprule
FileNumber & FileName & Samples & Variables & Task     \\
\midrule
F72  & 210\_cloud     & 108     & 5& regression     \\
F73  & 519\_vinnie    & 380     & 2& regression     \\
F74  & 573\_cpu\_act  & 1000    & 21     & regression     \\
F75  & 1027\_ESL& 488     & 4& regression     \\
F76  & 1028\_SWD& 1000    & 10     & regression     \\
F77  & 1029\_LEV& 1000    & 4& regression     \\
F78  & analcatdata\_boxing1 & 120     & 3& classification \\
F79  & car-evaluation & 1000    & 21     & classification \\
F80  & wine\_quality\_white & 1000    & 11     & classification \\
F81  & towerData& 1000    & 25     & regression \\
\bottomrule
\end{tabular}
\end{table*}

\begin{table*}[htbp]  
	\hspace*{-1.7cm}
	\setlength\tabcolsep{3pt}
	\centering  
	\scalebox{0.8}{
	\fontsize{7}{7}\selectfont  
	\begin{threeparttable}  
		\caption{Fitness Metrics}  
		\label{tab:performance_comparison} 
		\begin{tabular}{c|ccccccccccccccccccccc}  
			\toprule
			\multirow{2}{*}{Kinds}&  
			\multirow{2}{*}{Function}&  
			\multicolumn{2}{c}{TaylorGP}&\multicolumn{2}{c}{ GPLearn}&\multicolumn{2}{c}{FFX}&\multicolumn{2}{c}{GSGP}&\multicolumn{2}{c}{BSR}&\multicolumn{2}{c}{LR}&\multicolumn{2}{c}{KR}&\multicolumn{2}{c}{RF}&\multicolumn{2}{c}{SVM}&\multicolumn{2}{c}{XGBoost}\cr  
			\cmidrule(lr){3-4} \cmidrule(lr){5-6} \cmidrule(lr){7-8} \cmidrule(lr){9-10}\cmidrule(lr){11-12}\cmidrule(lr){13-14}\cmidrule(lr){15-16}\cmidrule(lr){17-18}\cmidrule(lr){19-20}\cmidrule(lr){21-22}
			&&CR&RMSE&CR&RMSE&CR&RMSE&CR&RMSE&CR&RMSE&CR&RMSE&CR&RMSE&CR&RMSE&CR&RMSE&CR&RMSE\cr  
			\midrule  
	SRB&F1&\textbf{100.0\%}&\textbf{.0000}&\textbf{100.0\%}&\textbf{.0000}&0.0\%&.2650&0.0\%&.0023&0.0\%&.0011&0.0\%&.1839&0.0\%&.1963&0.0\%&.1226&0.0\%&.0962&0.0\%&.0008\cr
	&F2&\textbf{100.0\%}&\textbf{.0000}&\textbf{100.0\%}&\textbf{.0000}&0.0\%&.2730&0.0\%&.0004&0.0\%&.0004&0.0\%&.0371&0.0\%&.0407&0.0\%&.0515&0.0\%&.0751&0.0\%&.0014\cr
	&F3&\textbf{100.0\%}&\textbf{.0000}&0.0\%&.0285&0.0\%&.2973&0.0\%&.1966&\textbf{100.0\%}&\textbf{.0000}&\textbf{100.0\%}&\textbf{.0000}&0.0\%&1.0412&0.0\%&2.0636&0.0\%&8.7351&0.0\%&.0015\cr
	&F4&\textbf{0.0\%}&\textbf{.0000}&\textbf{0.0\%}&.2189&\textbf{0.0\%}&.2620&\textbf{0.0\%}&.1641&\textbf{0.0\%}&.0006&\textbf{0.0\%}&.3308&\textbf{0.0\%}&1.1220&\textbf{0.0\%}&.1975&\textbf{0.0\%}&.1207&\textbf{0.0\%}&.0009\cr
	&F5&\textbf{100.0\%}&\textbf{.0000}&0.0\%&.1267&0.0\%&.2233&0.0\%&.0066&0.0\%&.0106&0.0\%&.8329&0.0\%&.4767&0.0\%&.6471&0.0\%&.5646&0.0\%&.0009\cr
	&F6&\textbf{100.0\%}&\textbf{.0000}&0.0\%&.5840&0.0\%&.1887&0.0\%&.0051&3.3\%&.0089&0.0\%&.6777&0.0\%&.4043&0.0\%&.3615&0.0\%&.5594&0.0\%&.0012\cr
	&F7&\textbf{100.0\%}&\textbf{.0000}&0.0\%&.2741&0.0\%&.2518&26.7\%&.0023&93.3\%&.0002&0.0\%&.6353&0.0\%&.2515&0.0\%&.3047&0.0\%&.2207&0.0\%&.0007\cr
	&F8&\textbf{100.0\%}&\textbf{.0000}&0.0\%&.3495&0.0\%&.2223&60.0\%&.0008&\textbf{100.0\%}&\textbf{.0000}&0.0\%&.3693&0.0\%&.1798&0.0\%&.2191&0.0\%&.2142&0.0\%&.0011\cr
	&F9&\textbf{100.0\%}&\textbf{.0000}&0.0\%&.0487&0.0\%&.3376&0.0\%&.0033&0.0\%&.0004&0.0\%&.0529&0.0\%&.0549&0.0\%&.0346&0.0\%&.0537&0.0\%&.0009\cr
	&F10&\textbf{100.0\%}&\textbf{.0000}&0.0\%&.1503&0.0\%&.3168&30.0\%&.0037&33.3\%&.0015&0.0\%&.1440&0.0\%&.1190&0.0\%&.0575&0.0\%&.0676&0.0\%&.0012\cr
	&F11&\textbf{100.0\%}&\textbf{.0000}&0.0\%&.1525&0.0\%&.3478&0.0\%&.0078&\textbf{100.0\%}&\textbf{.0000}&0.0\%&.1656&0.0\%&.1718&0.0\%&.0918&0.0\%&.0752&0.0\%&.0010\cr
	&F12&3.3\%&.0022&0.0\%&.3177&0.0\%&.3543&0.0\%&.0044&\textbf{100.0\%}&\textbf{.0000}&0.0\%&.2112&0.0\%&.1833&0.0\%&.1118&0.0\%&.0815&0.0\%&.0011\cr
	&F13&\textbf{100.0\%}&\textbf{.0000}&73.3\%&.4512&0.0\%&.1846&0.0\%&3.6371&33.3\%&1.1589&0.0\%&21.0820&0.0\%&6.0679&0.0\%&4.0784&0.0\%&22.7293&0.0\%&.0875\cr
	&F14&\textbf{0.0\%}&.0206&\textbf{0.0\%}&.3290&\textbf{0.0\%}&.2093&\textbf{0.0\%}&.0136&\textbf{0.0\%}&.1126&\textbf{0.0\%}&.4062&\textbf{0.0\%}&.0672&\textbf{0.0\%}&.0402&\textbf{0.0\%}&.0695&\textbf{0.0\%}&\textbf{.0053}\cr
	&F15&\textbf{100.0\%}&\textbf{.0000}&0.0\%&.3309&0.0\%&.2028&0.0\%&.0248&0.0\%&.0493&0.0\%&.3579&0.0\%&.1188&0.0\%&.0493&0.0\%&.0778&0.0\%&.0040\cr
	&F16&0.0\%&.0219&0.0\%&.3614&0.0\%&.2713&0.0\%&.0303&\textbf{100.0\%}&\textbf{.0000}&0.0\%&.4800&0.0\%&.1365&0.0\%&.1291&0.0\%&.0613&0.0\%&.0081\cr
	&F17&\textbf{100.0\%}&\textbf{.0000}&0.0\%&.1279&0.0\%&.2473&0.0\%&.0096&0.0\%&.0085&0.0\%&.0879&0.0\%&.0414&0.0\%&.0320&0.0\%&.0532&0.0\%&.0058\cr
	&F18&0.0\%&.0028&0.0\%&.3101&0.0\%&.2454&0.0\%&.0071&\textbf{100.0\%}&\textbf{.0000}&0.0\%&.2770&0.0\%&.0289&0.0\%&.0496&0.0\%&.0635&0.0\%&.0054\cr
	&F19&0.0\%&.0024&13.3\%&.2261&0.0\%&.2721&3.3\%&.0086&\textbf{100.0\%}&\textbf{.0000}&0.0\%&.1611&0.0\%&.0454&0.0\%&.0421&0.0\%&.0571&0.0\%&.0038\cr
	&F20&0.0\%&.0396&0.0\%&.0642&0.0\%&.1687&\textbf{3.3\%}&.0715&0.0\%&.1058&0.0\%&1.3504&0.0\%&.2706&0.0\%&.2488&0.0\%&.3687&0.0\%&\textbf{.0077}\cr
	&F21&0.0\%&\textbf{.0000}&0.0\%&.0394&0.0\%&.1256&\textbf{3.3\%}&.0012&0.0\%&.0022&0.0\%&.0533&0.0\%&.0190&0.0\%&.0048&0.0\%&.0287&0.0\%&.0018\cr
	&F22&0.0\%&.0001&0.0\%&.4750&0.0\%&.1861&3.3\%&.0078&\textbf{100.0\%}&\textbf{.0000}&0.0\%&.1052&0.0\%&.0666&0.0\%&.0401&0.0\%&.0502&0.0\%&.0062\cr
	&F23&\textbf{0.0\%}&\textbf{.0050}&\textbf{0.0\%}&.1827&\textbf{0.0\%}&.0958&\textbf{0.0\%}&.0189&\textbf{0.0\%}&.1298&\textbf{0.0\%}&.2734&\textbf{0.0\%}&.1461&\textbf{0.0\%}&.0388&\textbf{0.0\%}&.0406&\textbf{0.0\%}&.0057\cr
    \bottomrule
		best&&\textbf{16}&\textbf{16}&5&2&3&0&5&0&11&8&4&1&3&0&3&0&3&0&3&2\cr   
	\bottomrule 
	FSRB&F24&0.0\%&.0008&0.0\%&.0560&0.0\%&.3044&0.0\%&.0001&\textbf{13.3\%}&\textbf{.0001}&0.0\%&.0212&0.0\%&.0283&0.0\%&.0113&0.0\%&.0771&0.0\%&.0011\cr
	&F25&\textbf{0.0\%}&\textbf{.0001}&\textbf{0.0\%}&.0246&\textbf{0.0\%}&.1937&\textbf{0.0\%}&.0006&\textbf{0.0\%}&.0054&\textbf{0.0\%}&.0167&\textbf{0.0\%}&.0038&\textbf{0.0\%}&.0026&\textbf{0.0\%}&.0343&\textbf{0.0\%}&.0010\cr
	&F26&\textbf{100.0\%}&\textbf{.0000}&\textbf{100.0\%}&\textbf{.0000}&0.0\%&.2363&\textbf{100.0\%}&\textbf{.0000}&\textbf{100.0\%}&\textbf{.0000}&0.0\%&1.4143&0.0\%&.0086&0.0\%&.3157&0.0\%&.3371&0.0\%&.0157\cr
	&F27&\textbf{100.0\%}&\textbf{.0000}&\textbf{100.0\%}&\textbf{.0000}&0.0\%&.2265&\textbf{100.0\%}&\textbf{.0000}&\textbf{100.0\%}&\textbf{.0000}&0.0\%&1.4301&0.0\%&.0090&0.0\%&.3248&0.0\%&.3430&0.0\%&.0158\cr
	&F28&\textbf{100.0\%}&\textbf{.0000}&53.3\%&.1140&0.0\%&.2257&6.7\%&.2939&\textbf{100.0\%}&\textbf{.0000}&0.0\%&4.8804&0.0\%&.0025&0.0\%&.8526&0.0\%&2.1924&0.0\%&.0339\cr
	&F29&\textbf{100.0\%}&\textbf{.0000}&\textbf{100.0\%}&\textbf{.0000}&0.0\%&.1817&\textbf{100.0\%}&\textbf{.0000}&\textbf{100.0\%}&\textbf{.0000}&0.0\%&.3584&0.0\%&.0906&0.0\%&.0690&0.0\%&.0799&0.0\%&.0052\cr
	&F30&0.0\%&.0070&0.0\%&.2547&0.0\%&.1709&0.0\%&.0127&\textbf{3.3\%}&.0125&0.0\%&.2095&0.0\%&.0374&0.0\%&.0218&0.0\%&.0660&0.0\%&\textbf{.0036}\cr
	&F31&\textbf{100.0\%}&\textbf{.0000}&\textbf{100.0\%}&\textbf{.0000}&0.0\%&.1810&\textbf{100.0\%}&\textbf{.0000}&\textbf{100.0\%}&\textbf{.0000}&0.0\%&.3685&0.0\%&.0858&0.0\%&.0685&0.0\%&.0856&0.0\%&.0053\cr
	&F32&\textbf{100.0\%}&\textbf{.0000}&0.0\%&.0080&0.0\%&.2283&0.0\%&.0182&\textbf{100.0\%}&\textbf{.0000}&0.0\%&.2293&0.0\%&.0014&0.0\%&.0514&0.0\%&.0660&0.0\%&.0047\cr
	&F33&\textbf{100.0\%}&\textbf{.0000}&16.7\%&.0017&0.0\%&.2307&13.3\%&.0163&\textbf{100.0\%}&\textbf{.0000}&0.0\%&2.1527&0.0\%&.0136&0.0\%&.4842&0.0\%&.6682&0.0\%&.0202\cr
	&F34&0.0\%&.0004&0.0\%&.0348&0.0\%&.1672&0.0\%&.0006&\textbf{100.0\%}&\textbf{.0000}&0.0\%&.0363&0.0\%&.0128&0.0\%&.0055&0.0\%&.0888&0.0\%&.0011\cr
	&F35&\textbf{100.0\%}&\textbf{.0000}&46.7\%&.1284&0.0\%&.2278&10.0\%&.1321&\textbf{100.0\%}&\textbf{.0000}&0.0\%&4.8270&0.0\%&.0025&0.0\%&.9328&0.0\%&2.4628&0.0\%&.0333\cr
	&F36&0.0\%&\textbf{.0000}&0.0\%&.2302&0.0\%&.2178&0.0\%&.0030&\textbf{3.3\%}&.0002&0.0\%&.1287&0.0\%&.0302&0.0\%&.0233&0.0\%&.0521&0.0\%&.0035\cr
	&F37&\textbf{100.0\%}&\textbf{.0000}&\textbf{100.0\%}&\textbf{.0000}&0.0\%&.2238&\textbf{100.0\%}&\textbf{.0000}&\textbf{100.0\%}&\textbf{.0000}&0.0\%&9.5658&0.0\%&.0051&0.0\%&1.7310&0.0\%&8.4967&0.0\%&.0491\cr
	&F38&0.0\%&\textbf{.0001}&0.0\%&.1666&0.0\%&.1995&6.7\%&.0015&\textbf{36.7\%}&.0006&0.0\%&.0687&0.0\%&.0108&0.0\%&.0161&0.0\%&.0602&0.0\%&.0030\cr
	&F39&\textbf{100.0\%}&\textbf{.0000}&0.0\%&.0081&0.0\%&.2307&0.0\%&.0218&\textbf{100.0\%}&\textbf{.0000}&0.0\%&.2295&0.0\%&.0014&0.0\%&.0519&0.0\%&.0662&0.0\%&.0049\cr
	&F40&90.0\%&.0002&90.0\%&.0276&0.0\%&.1917&6.7\%&.0086&\textbf{100.0\%}&\textbf{.0000}&0.0\%&.2616&0.0\%&.0201&0.0\%&.2134&0.0\%&.0815&0.0\%&.0125\cr
	&F41&\textbf{100.0\%}&\textbf{.0000}&\textbf{100.0\%}&\textbf{.0000}&0.0\%&.1923&0.0\%&.0195&\textbf{100.0\%}&\textbf{.0000}&0.0\%&.1400&0.0\%&.0564&0.0\%&.0597&0.0\%&.0511&0.0\%&.0057\cr
	&F42&\textbf{0.0\%}&.0020&\textbf{0.0\%}&.0215&\textbf{0.0\%}&.2736&\textbf{0.0\%}&\textbf{.0004}&\textbf{0.0\%}&.0018&\textbf{0.0\%}&.0051&\textbf{0.0\%}&.0050&\textbf{0.0\%}&.0014&\textbf{0.0\%}&.0256&\textbf{0.0\%}&.0005\cr
	&F43&\textbf{0.0\%}&.0083&\textbf{0.0\%}&.2578&\textbf{0.0\%}&.2149&\textbf{0.0\%}&\textbf{.0033}&\textbf{0.0\%}&.0167&\textbf{0.0\%}&.0895&\textbf{0.0\%}&.0138&\textbf{0.0\%}&.0190&\textbf{0.0\%}&.0638&\textbf{0.0\%}&.0039\cr
	&F44&0.0\%&\textbf{.0000}&0.0\%&.0070&0.0\%&.1796&0.0\%&.0002&\textbf{6.7\%}&.0009&0.0\%&.0024&0.0\%&.0014&0.0\%&.0008&0.0\%&.0143&0.0\%&.0004\cr
	&F45&\textbf{100.0\%}&\textbf{.0000}&\textbf{100.0\%}&\textbf{.0000}&0.0\%&.1932&80.0\%&.0028&\textbf{100.0\%}&\textbf{.0000}&0.0\%&.2215&0.0\%&.0172&0.0\%&.1632&0.0\%&.0665&0.0\%&.0106\cr
	&F46&\textbf{0.0\%}&.0091&\textbf{0.0\%}&.2578&\textbf{0.0\%}&.2183&\textbf{0.0\%}&\textbf{.0033}&\textbf{0.0\%}&.0168&\textbf{0.0\%}&.0887&\textbf{0.0\%}&.0143&\textbf{0.0\%}&.0192&\textbf{0.0\%}&.0641&\textbf{0.0\%}&.0039\cr
	&F47&\textbf{0.0\%}&\textbf{.0010}&\textbf{0.0\%}&.0220&\textbf{0.0\%}&.2776&\textbf{0.0\%}&.0017&\textbf{0.0\%}&.0147&\textbf{0.0\%}&.0310&\textbf{0.0\%}&.0106&\textbf{0.0\%}&.0052&\textbf{0.0\%}&.0504&\textbf{0.0\%}&.0017\cr
	&F48&\textbf{3.3\%}&\textbf{.0002}&\textbf{3.3\%}&.0834&0.0\%&.1882&\textbf{3.3\%}&.0012&\textbf{3.3\%}&.0078&0.0\%&.0142&0.0\%&.0066&0.0\%&.0086&0.0\%&.0458&0.0\%&.0022\cr
	&F49&\textbf{0.0\%}&.0518&\textbf{0.0\%}&.8525&\textbf{0.0\%}&.2279&\textbf{0.0\%}&.0404&\textbf{0.0\%}&.1000&\textbf{0.0\%}&.2555&\textbf{0.0\%}&.1114&\textbf{0.0\%}&.0605&\textbf{0.0\%}&.0559&\textbf{0.0\%}&\textbf{.0079}\cr
	&F50&\textbf{0.0\%}&.0008&\textbf{0.0\%}&.0396&\textbf{0.0\%}&.1870&\textbf{0.0\%}&\textbf{.0003}&\textbf{0.0\%}&.0004&\textbf{0.0\%}&.0075&\textbf{0.0\%}&.0044&\textbf{0.0\%}&.0039&\textbf{0.0\%}&.0518&\textbf{0.0\%}&.0014\cr
	&F51&\textbf{100.0\%}&\textbf{.0000}&73.3\%&.0748&0.0\%&.1902&10.0\%&.0049&20.0\%&.0210&0.0\%&.0758&0.0\%&.0231&0.0\%&.0403&0.0\%&.0596&0.0\%&.0039\cr
	&F52&\textbf{0.0\%}&\textbf{.0003}&\textbf{0.0\%}&.2010&\textbf{0.0\%}&.2007&\textbf{0.0\%}&.0033&\textbf{0.0\%}&.0141&\textbf{0.0\%}&.0440&\textbf{0.0\%}&.0159&\textbf{0.0\%}&.0274&\textbf{0.0\%}&.0665&\textbf{0.0\%}&.0039\cr
	&F53&\textbf{0.0\%}&\textbf{.0005}&\textbf{0.0\%}&.3903&\textbf{0.0\%}&.2456&\textbf{0.0\%}&.0186&\textbf{0.0\%}&.0105&\textbf{0.0\%}&.0946&\textbf{0.0\%}&.0607&\textbf{0.0\%}&.0359&\textbf{0.0\%}&.0606&\textbf{0.0\%}&.0051\cr
	&F54&\textbf{13.3\%}&\textbf{.0001}&0.0\%&.2937&0.0\%&.1806&0.0\%&.0048&0.0\%&.0037&0.0\%&.0647&0.0\%&.0272&0.0\%&.0417&0.0\%&.0448&0.0\%&.0046\cr
	&F55&\textbf{100.0\%}&\textbf{.0000}&\textbf{100.0\%}&\textbf{.0000}&0.0\%&.1909&86.7\%&.0016&\textbf{100.0\%}&\textbf{.0000}&0.0\%&.2201&0.0\%&.0178&0.0\%&.1605&0.0\%&.0635&0.0\%&.0108\cr
	&F56&\textbf{100.0\%}&\textbf{.0000}&93.3\%&.0067&0.0\%&.1829&0.0\%&.0010&0.0\%&.0070&0.0\%&.0133&0.0\%&.0108&0.0\%&.0142&0.0\%&.0394&0.0\%&.0027\cr
	&F57&0.0\%&\textbf{.0000}&0.0\%&.0116&0.0\%&.1643&0.0\%&.0003&\textbf{80.0\%}&.0002&0.0\%&.0036&0.0\%&.0016&0.0\%&.0015&0.0\%&.0232&0.0\%&.0008\cr
	&F58&\textbf{20.0\%}&\textbf{.0000}&0.0\%&.0162&0.0\%&.1764&0.0\%&.0004&10.0\%&.0015&0.0\%&.0044&0.0\%&.0015&0.0\%&.0018&0.0\%&.0306&0.0\%&.0006\cr
	&F59&\textbf{56.7\%}&\textbf{.0000}&0.0\%&.0689&0.0\%&.1810&3.3\%&.0008&6.7\%&.0056&0.0\%&.0154&0.0\%&.0091&0.0\%&.0079&0.0\%&.0533&0.0\%&.0023\cr
	&F60&\textbf{0.0\%}&\textbf{.0004}&\textbf{0.0\%}&.1609&\textbf{0.0\%}&.2430&\textbf{0.0\%}&.0011&\textbf{0.0\%}&.0082&\textbf{0.0\%}&.0155&\textbf{0.0\%}&.0096&\textbf{0.0\%}&.0093&\textbf{0.0\%}&.0508&\textbf{0.0\%}&.0023\cr
	&F61&\textbf{0.0\%}&.0090&\textbf{0.0\%}&.2635&\textbf{0.0\%}&.2197&\textbf{0.0\%}&\textbf{.0032}&\textbf{0.0\%}&.0166&\textbf{0.0\%}&.0886&\textbf{0.0\%}&.0139&\textbf{0.0\%}&.0196&\textbf{0.0\%}&.0646&\textbf{0.0\%}&.0039\cr
	&F62&\textbf{100.0\%}&\textbf{.0000}&\textbf{100.0\%}&\textbf{.0000}&0.0\%&.1988&53.3\%&.0186&\textbf{100.0\%}&\textbf{.0000}&0.0\%&.7969&0.0\%&.1435&0.0\%&.4133&0.0\%&.2370&0.0\%&.0293\cr
	&F63&33.3\%&.0000&0.0\%&.0717&0.0\%&.1785&0.0\%&.0017&\textbf{100.0\%}&\textbf{.0000}&0.0\%&.0167&0.0\%&.0037&0.0\%&.0085&0.0\%&.0508&0.0\%&.0028\cr
	&F64&\textbf{100.0\%}&\textbf{.0000}&13.3\%&.3832&0.0\%&.1920&0.0\%&.0137&\textbf{100.0\%}&\textbf{.0000}&0.0\%&.1113&0.0\%&.0086&0.0\%&.0794&0.0\%&.0520&0.0\%&.0062\cr
	&F65&43.3\%&.0000&0.0\%&.0557&0.0\%&.1712&0.0\%&.0022&\textbf{100.0\%}&\textbf{.0000}&0.0\%&.0204&0.0\%&.0034&0.0\%&.0069&0.0\%&.0616&0.0\%&.0022\cr
	&F66&0.0\%&.0213&0.0\%&.4185&0.0\%&.1805&0.0\%&.1470&\textbf{100.0\%}&\textbf{.0000}&0.0\%&.8053&0.0\%&.2694&0.0\%&.5194&0.0\%&.3044&0.0\%&.0258\cr
	&F67&0.0\%&\textbf{.0002}&0.0\%&.0034&0.0\%&.1589&\textbf{3.3\%}&.0002&0.0\%&.0007&0.0\%&.0017&0.0\%&.0006&0.0\%&.0005&0.0\%&.0098&0.0\%&.0003\cr
	&F68&0.0\%&.0010&0.0\%&.4232&0.0\%&.1785&0.0\%&.0145&\textbf{100.0\%}&\textbf{.0000}&0.0\%&.1147&0.0\%&.0976&0.0\%&.0674&0.0\%&.0507&0.0\%&.0064\cr
	&F69&\textbf{100.0\%}&\textbf{.0000}&\textbf{100.0\%}&\textbf{.0000}&0.0\%&.2414&0.0\%&.0185&\textbf{100.0\%}&\textbf{.0000}&0.0\%&.1715&0.0\%&.0352&0.0\%&.0511&0.0\%&.0577&0.0\%&.0052\cr
	&F70&\textbf{100.0\%}&\textbf{.0000}&\textbf{100.0\%}&\textbf{.0000}&0.0\%&.2405&0.0\%&.0179&\textbf{100.0\%}&\textbf{.0000}&0.0\%&.1696&0.0\%&.0358&0.0\%&.0504&0.0\%&.0570&0.0\%&.0052\cr
	&F71&96.7\%&.0002&90.0\%&.0237&0.0\%&.2350&0.0\%&.3014&\textbf{100.0\%}&\textbf{.0000}&0.0\%&.9990&0.0\%&.1960&0.0\%&.1456&0.0\%&.1701&0.0\%&.0238\cr
    \bottomrule
		best&&34&\textbf{33}&23&11&11&0&18&10&\textbf{42}&25&11&0&11&0&11&0&11&0&11&2\cr   
	\bottomrule 
	PMLB&F72&\textbf{0.0\%}&.2481&\textbf{0.0\%}&.3688&\textbf{0.0\%}&.1678&\textbf{0.0\%}&.1228&\textbf{0.0\%}&.2856&\textbf{0.0\%}&.3301&\textbf{0.0\%}&.3623&\textbf{0.0\%}&.4857&\textbf{0.0\%}&.4146&\textbf{0.0\%}&\textbf{.0014}\cr
	&F73&\textbf{0.0\%}&1.2169&\textbf{0.0\%}&1.3312&\textbf{0.0\%}&\textbf{.2063}&\textbf{0.0\%}&1.1944&\textbf{0.0\%}&1.5121&\textbf{0.0\%}&1.5429&\textbf{0.0\%}&1.5339&\textbf{0.0\%}&1.7134&\textbf{0.0\%}&1.6835&\textbf{0.0\%}&1.4172\cr
	&F74&\textbf{0.0\%}&3.2050&\textbf{0.0\%}&4.4539&\textbf{0.0\%}&.1844&\textbf{0.0\%}&11.4619&\textbf{0.0\%}&6.4230&\textbf{0.0\%}&8.9662&\textbf{0.0\%}&14.7675&\textbf{0.0\%}&2.8844&\textbf{0.0\%}&14.8341&\textbf{0.0\%}&\textbf{.0869}\cr
	&F75&\textbf{0.0\%}&.4316&\textbf{0.0\%}&.5287&\textbf{0.0\%}&\textbf{.1770}&\textbf{0.0\%}&.3300&\textbf{0.0\%}&.5126&\textbf{0.0\%}&.5217&\textbf{0.0\%}&.5495&\textbf{0.0\%}&.5751&\textbf{0.0\%}&.4748&\textbf{0.0\%}&.3541\cr
	&F76&\textbf{0.0\%}&.5305&\textbf{0.0\%}&.6163&\textbf{0.0\%}&\textbf{.2702}&\textbf{0.0\%}&.4354&\textbf{0.0\%}&.6406&\textbf{0.0\%}&.6249&\textbf{0.0\%}&.6319&\textbf{0.0\%}&.6469&\textbf{0.0\%}&.6038&\textbf{0.0\%}&.5625\cr
	&F77&0.0\%&.5359&0.0\%&.7612&0.0\%&\textbf{.2380}&\textbf{3.3\%}&.4000&0.0\%&.6279&0.0\%&.6419&0.0\%&.6306&0.0\%&.6671&0.0\%&.6231&0.0\%&.5859\cr
	&F78&\textbf{0.0\%}&.2457&\textbf{0.0\%}&.3440&\textbf{0.0\%}&.4811&\textbf{0.0\%}&.2252&\textbf{0.0\%}&.3491&\textbf{0.0\%}&.4483&\textbf{0.0\%}&.4479&\textbf{0.0\%}&.3813&\textbf{0.0\%}&.4368&\textbf{0.0\%}&\textbf{.0058}\cr
	&F79&\textbf{100.0\%}&\textbf{.0000}&\textbf{100.0\%}&\textbf{.0000}&0.0\%&.4716&0.0\%&.0335&\textbf{100.0\%}&\textbf{.0000}&\textbf{100.0\%}&\textbf{.0000}&0.0\%&.0082&\textbf{100.0\%}&\textbf{.0000}&0.0\%&.0985&0.0\%&.0000\cr
	&F80&\textbf{0.0\%}&.6331&\textbf{0.0\%}&.6690&\textbf{0.0\%}&.1561&\textbf{0.0\%}&.5122&\textbf{0.0\%}&.7848&\textbf{0.0\%}&.7822&\textbf{0.0\%}&.7923&\textbf{0.0\%}&.6766&\textbf{0.0\%}&.8713&\textbf{0.0\%}&\textbf{.0360}\cr
	&F81&\textbf{0.0\%}&15.1959&\textbf{0.0\%}&32.7227&\textbf{0.0\%}&\textbf{.1916}&\textbf{0.0\%}&25.3951&\textbf{0.0\%}&45.7915&\textbf{0.0\%}&28.3523&\textbf{0.0\%}&30.6675&\textbf{0.0\%}&23.4327&\textbf{0.0\%}&83.1058&\textbf{0.0\%}&.8736\cr
    \bottomrule
		best&&\textbf{9}&1&\textbf{9}&1&8&\textbf{5}&\textbf{9}&0&\textbf{9}&1&\textbf{9}&1&8&0&\textbf{9}&1&8&0&8&4\cr   
	\bottomrule \end{tabular} 
	
            \begin{tablenotes}
                \item CR represents the proportion of finding the correct result whose fitness computed by RMSE is less than 1e-5.
            \end{tablenotes}
        \end{threeparttable}  
    }
    \end{table*}

\begin{figure*}[htb]
	\centering
    \hspace{-0.5cm} \quad
		\subfigure{ \includegraphics[width=3cm]{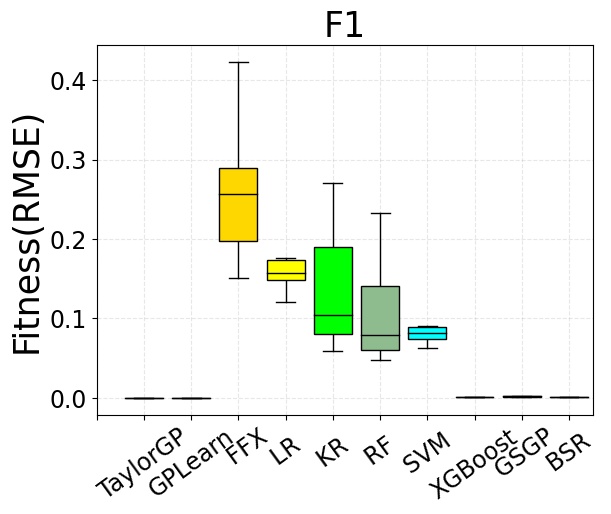}
	}
	\hspace{-0.5cm} \quad
		\subfigure{ \includegraphics[width=3cm]{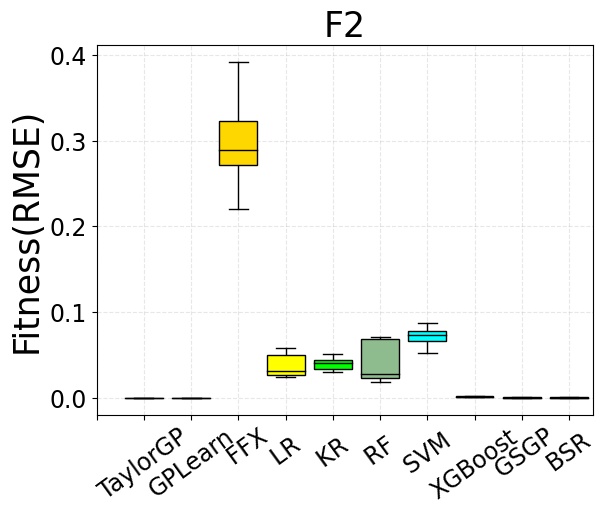}
    }
    \hspace{-0.5cm} \quad
		\subfigure{ \includegraphics[width=3cm]{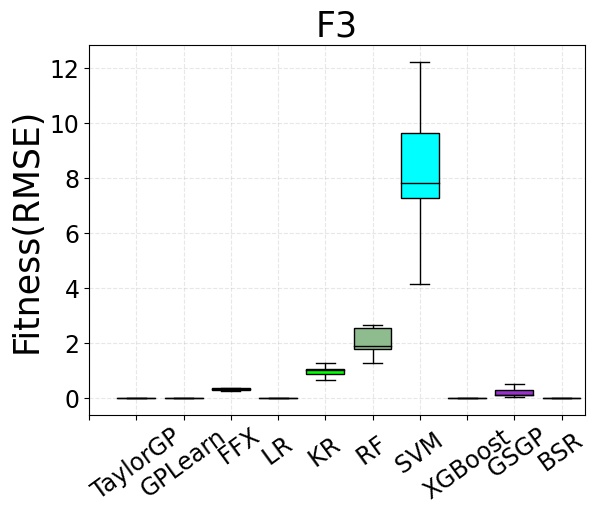}
	}
	\hspace{-0.5cm} \quad
		\subfigure{ \includegraphics[width=3cm]{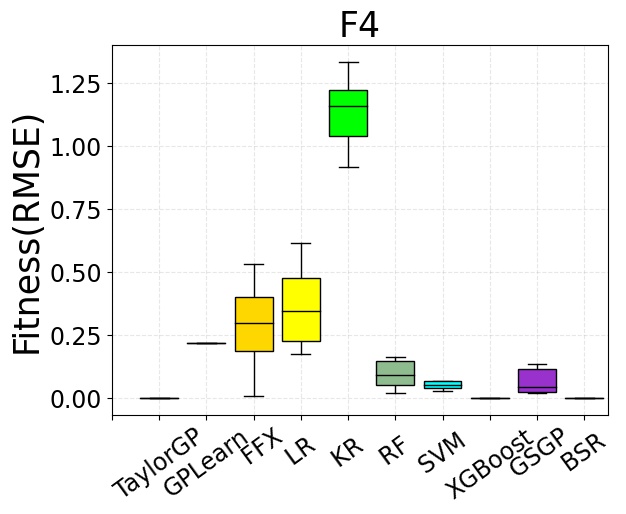}
    }
    \hspace{-0.5cm} \quad
		\subfigure{ \includegraphics[width=3cm]{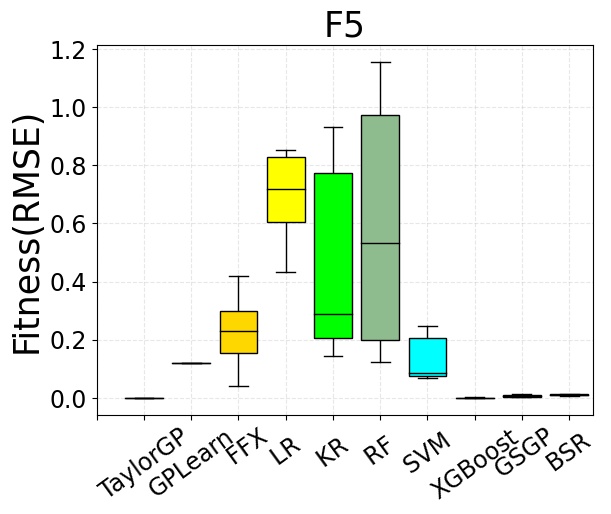}
	}
	\hspace{-0.5cm} \quad
		\subfigure{ \includegraphics[width=3cm]{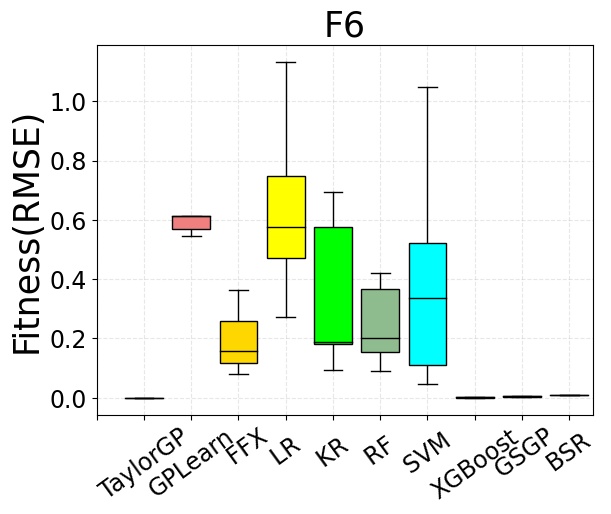}
    }
    \hspace{-0.5cm} \quad
		\subfigure{ \includegraphics[width=3cm]{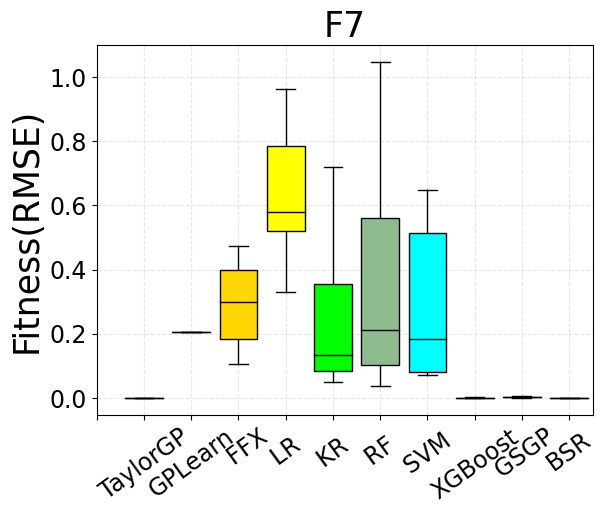}
	}
	\hspace{-0.5cm} \quad
		\subfigure{ \includegraphics[width=3cm]{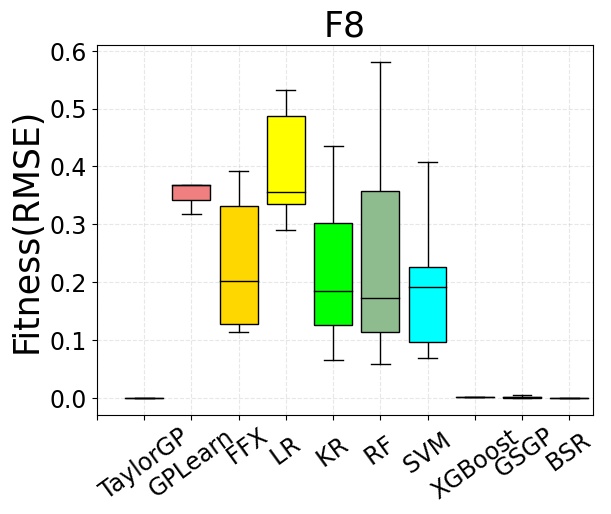}
    }
    \hspace{-0.5cm} \quad
		\subfigure{ \includegraphics[width=3cm]{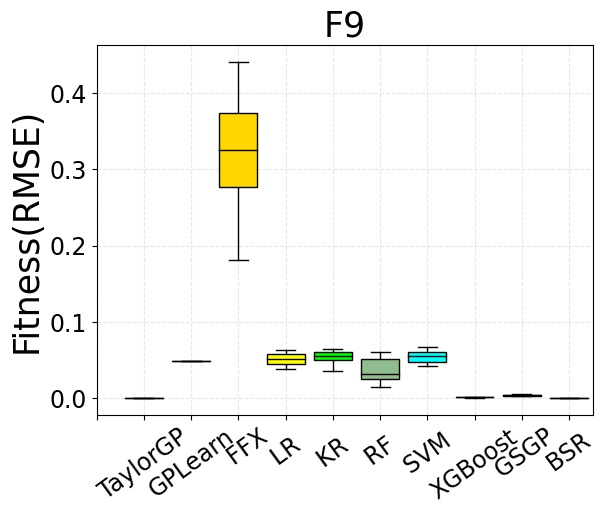}
	}
	\hspace{-0.5cm} \quad
		\subfigure{ \includegraphics[width=3cm]{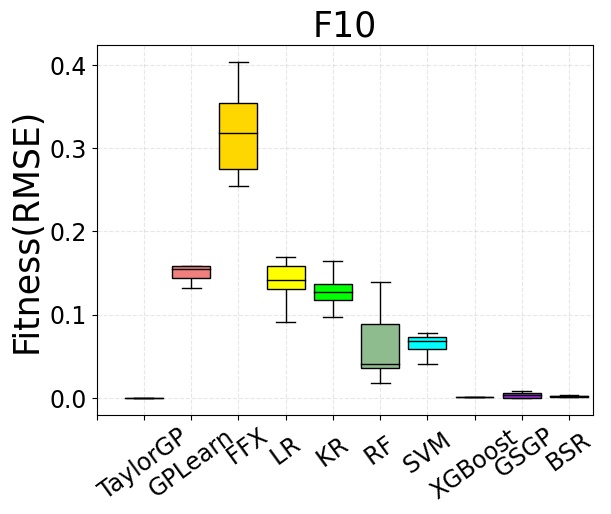}
    }
    \hspace{-0.5cm} \quad
		\subfigure{ \includegraphics[width=3cm]{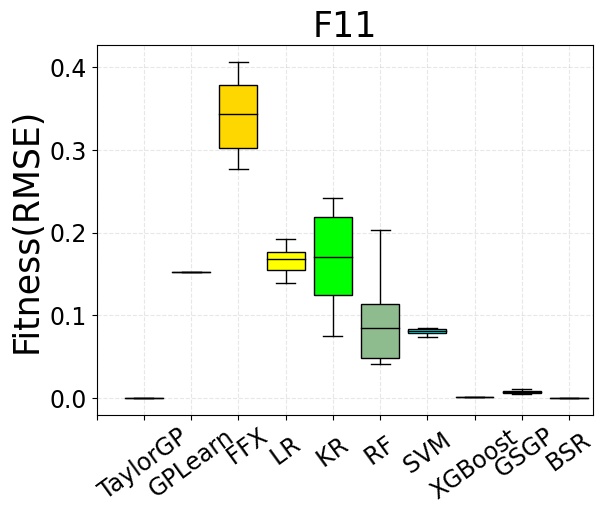}
	}
	\hspace{-0.5cm} \quad
		\subfigure{ \includegraphics[width=3cm]{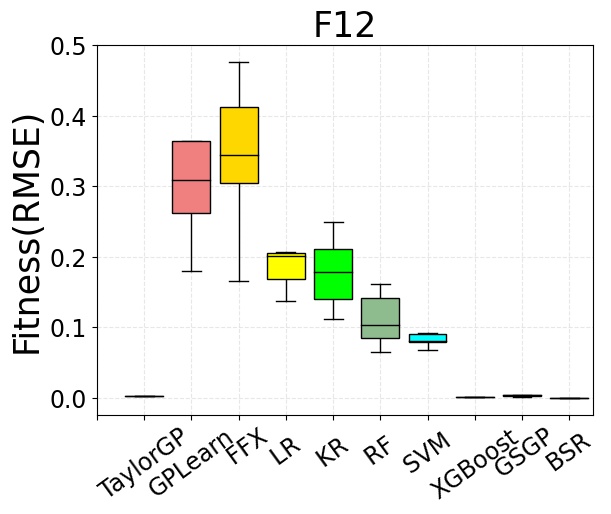}
    }
    \hspace{-0.5cm} \quad
		\subfigure{ \includegraphics[width=3cm]{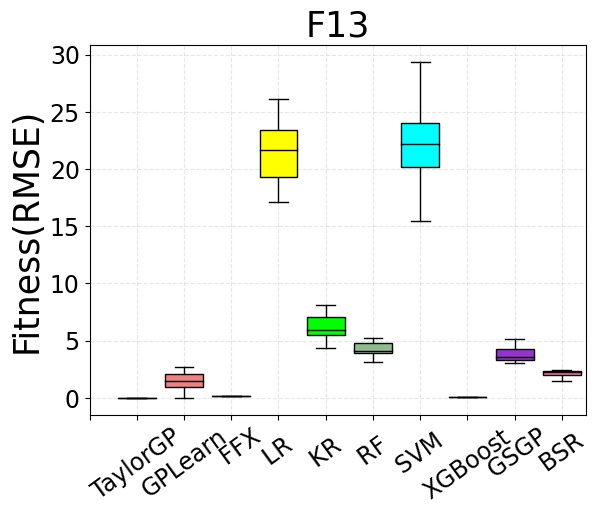}
	}
	\hspace{-0.5cm} \quad
		\subfigure{ \includegraphics[width=3cm]{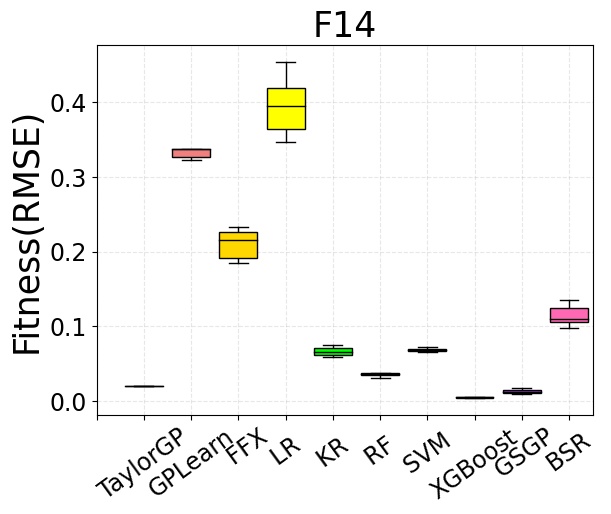}
    }
    \hspace{-0.5cm} \quad
		\subfigure{ \includegraphics[width=3cm]{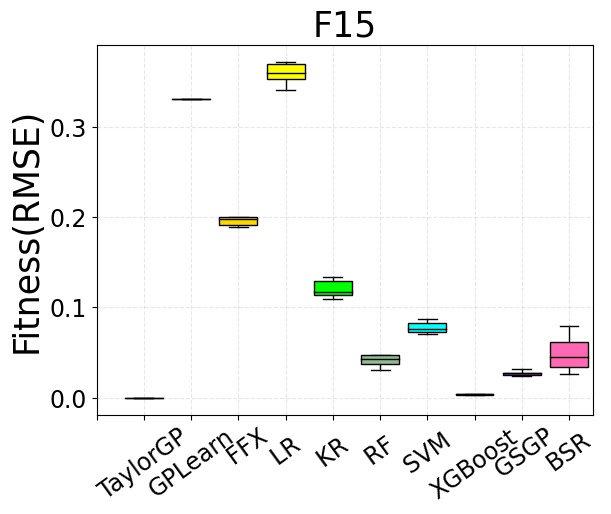}
	}
	\hspace{-0.5cm} \quad
		\subfigure{ \includegraphics[width=3cm]{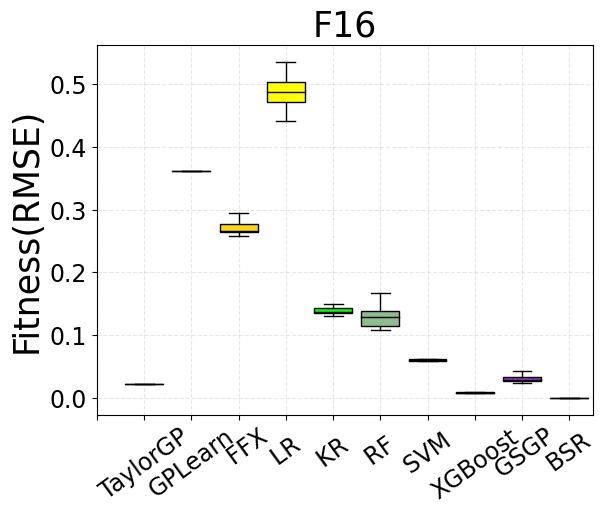}
    }
    \hspace{-0.5cm} \quad
		\subfigure{ \includegraphics[width=3cm]{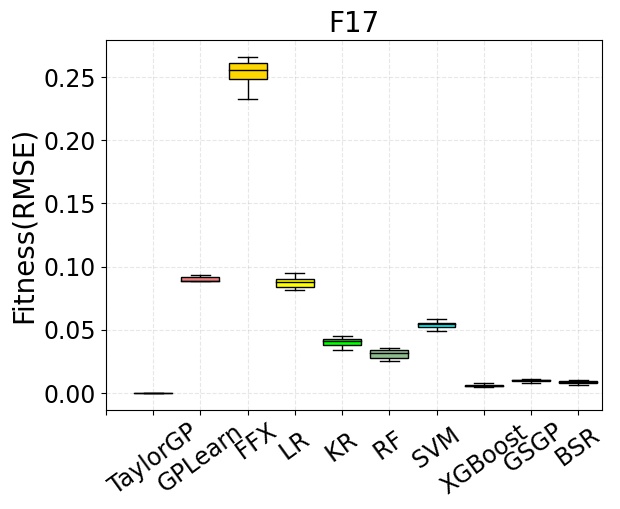}
	}
	\hspace{-0.5cm} \quad
		\subfigure{ \includegraphics[width=3cm]{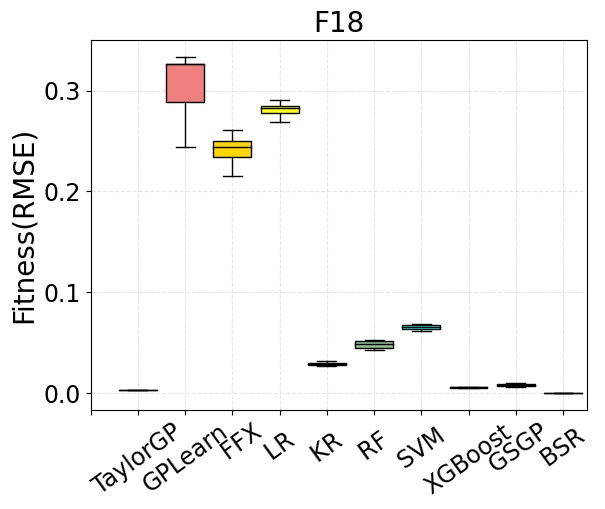}
    }
    \hspace{-0.5cm} \quad
		\subfigure{ \includegraphics[width=3cm]{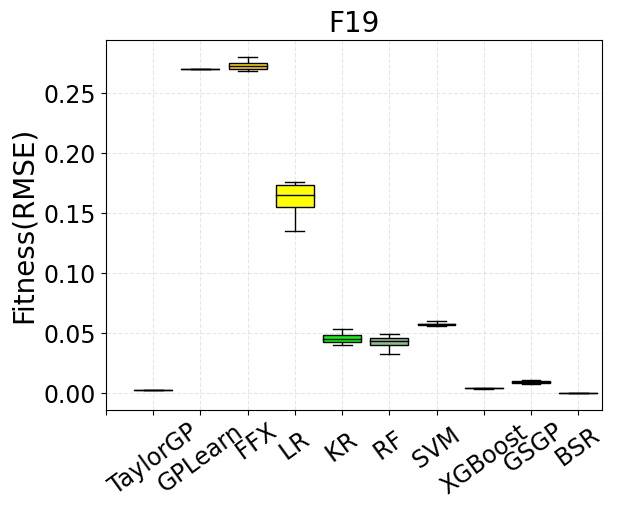}
	}
	\hspace{-0.5cm} \quad
		\subfigure{ \includegraphics[width=3cm]{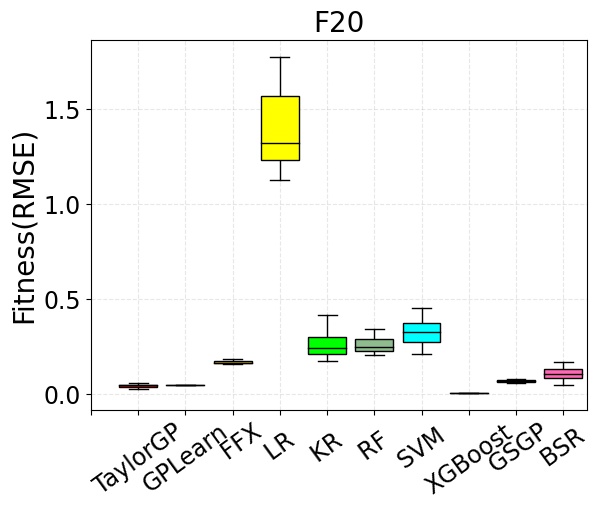}
    }
    \hspace{-0.5cm} \quad
		\subfigure{ \includegraphics[width=3cm]{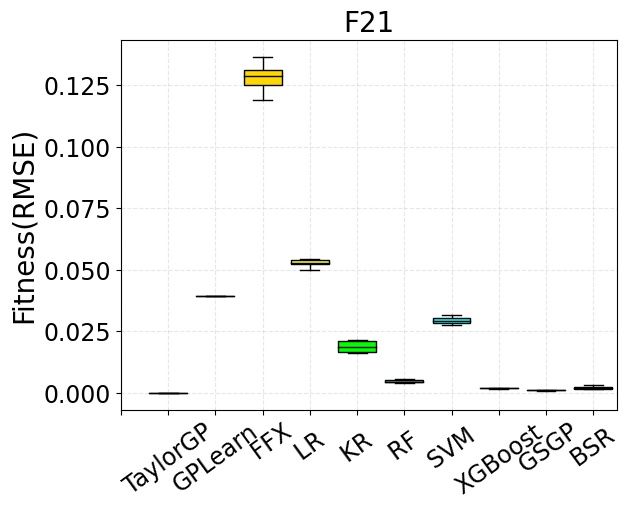}
	}
	\hspace{-0.5cm} \quad
		\subfigure{ \includegraphics[width=3cm]{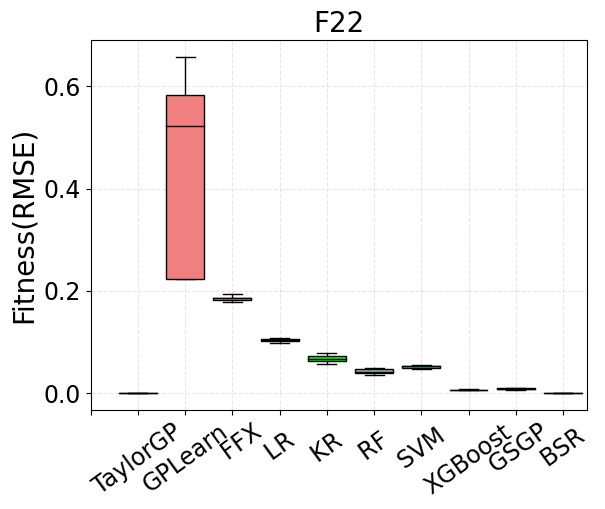}
    }
    \hspace{-0.5cm} \quad
		\subfigure{ \includegraphics[width=3cm]{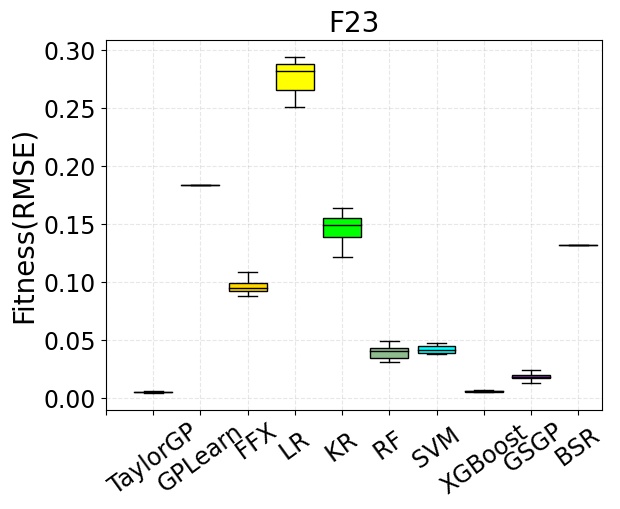}
	}
    \hspace{-0.5cm} \quad
		\subfigure{ \includegraphics[width=3cm]{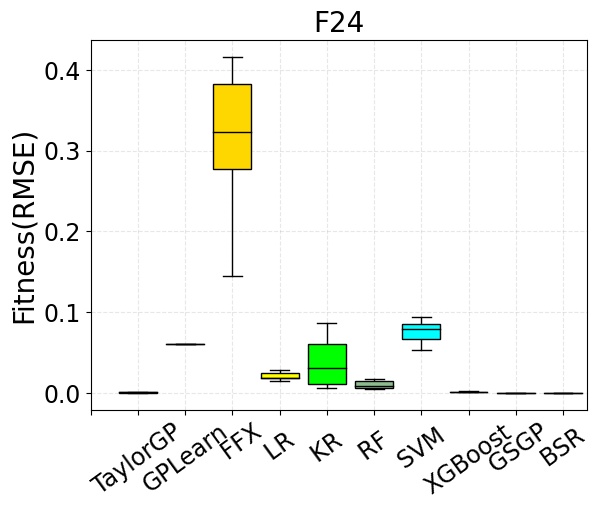}
	}
    \hspace{-0.5cm} \quad
		\subfigure{ \includegraphics[width=3cm]{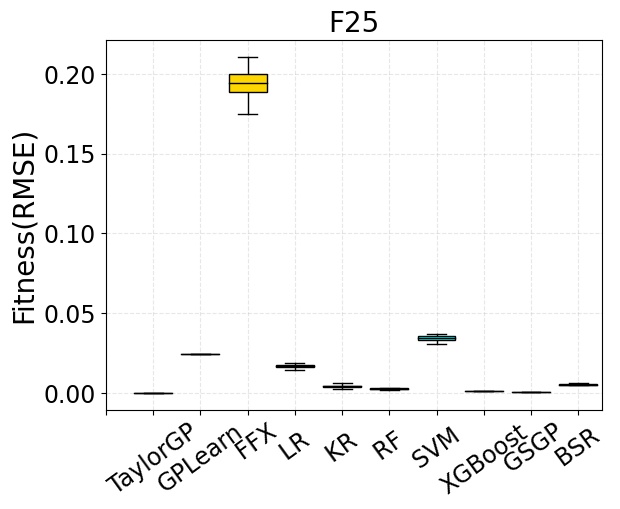}
	}
	    \hspace{-0.5cm} \quad
		\subfigure{ \includegraphics[width=3cm]{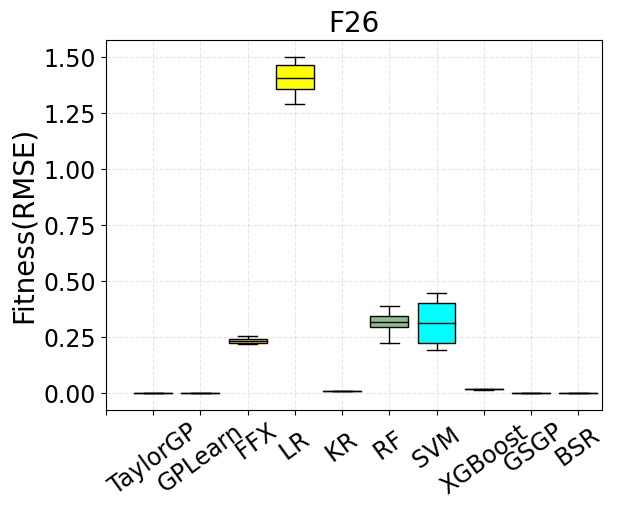}
	}
    \hspace{-0.5cm} \quad
		\subfigure{ \includegraphics[width=3cm]{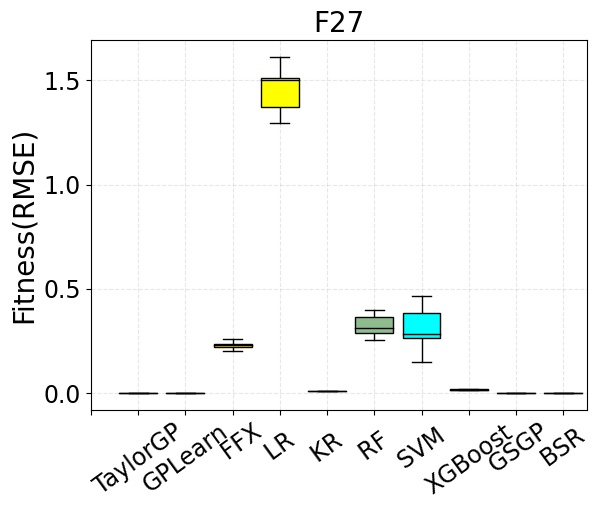}
	}
	    \hspace{-0.5cm} \quad
		\subfigure{ \includegraphics[width=3cm]{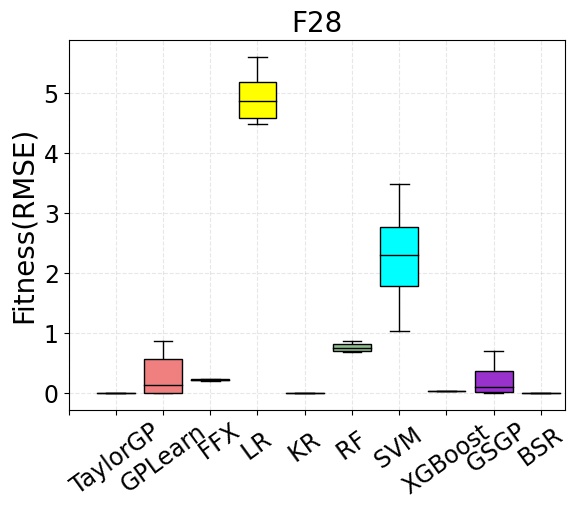}
	}
    \hspace{-0.5cm} \quad
		\subfigure{ \includegraphics[width=3cm]{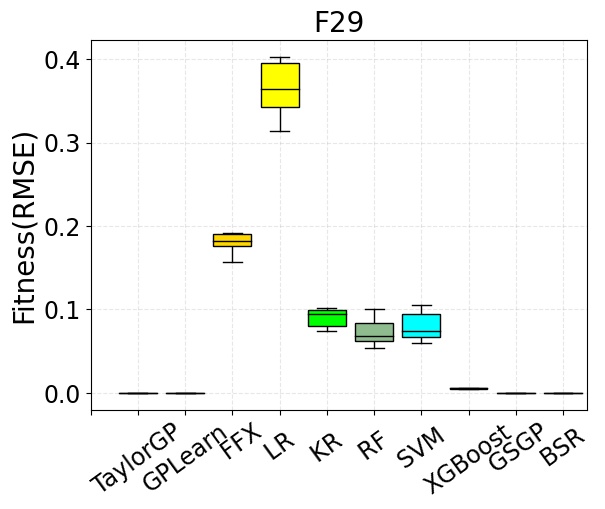}
	}
    \hspace{-0.5cm} \quad
		\subfigure{ \includegraphics[width=3cm]{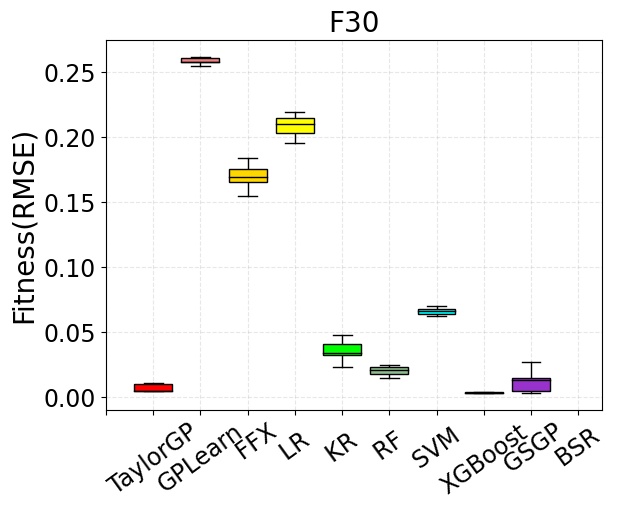}
	}

	\caption{Fitness comparison. The RMSE results on the benchmarks F1-F30.}
	\label{fig:overallBox}
\end{figure*}

\begin{figure*}[htb]
	\centering

    \hspace{-0.5cm} \quad 
		\subfigure{ \includegraphics[width=3cm]{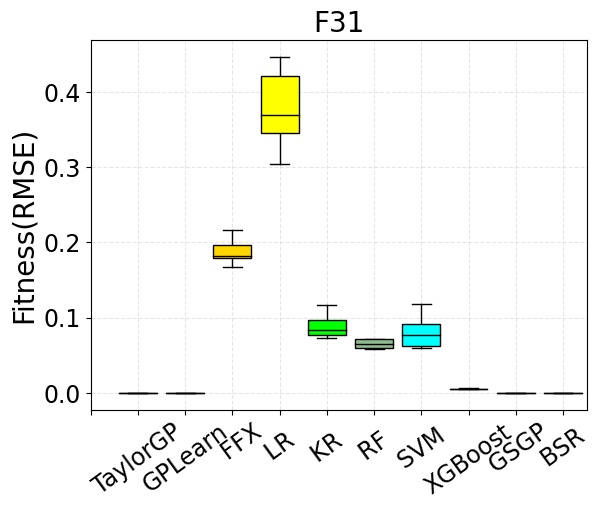}
	}
    \hspace{-0.5cm} \quad
		\subfigure{ \includegraphics[width=3cm]{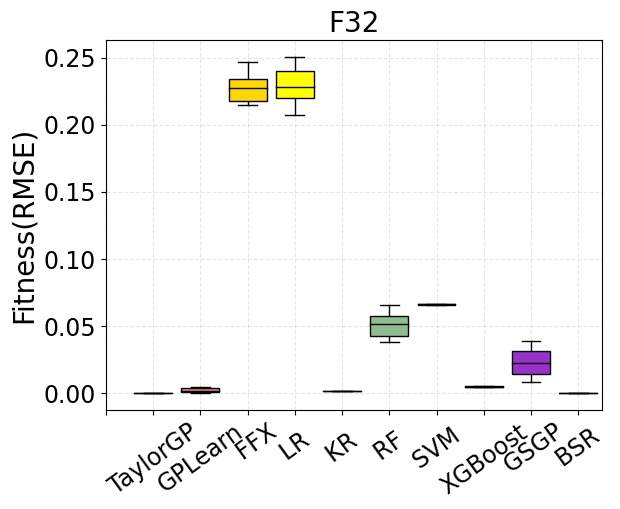}
	}
    \hspace{-0.5cm} \quad
		\subfigure{ \includegraphics[width=3cm]{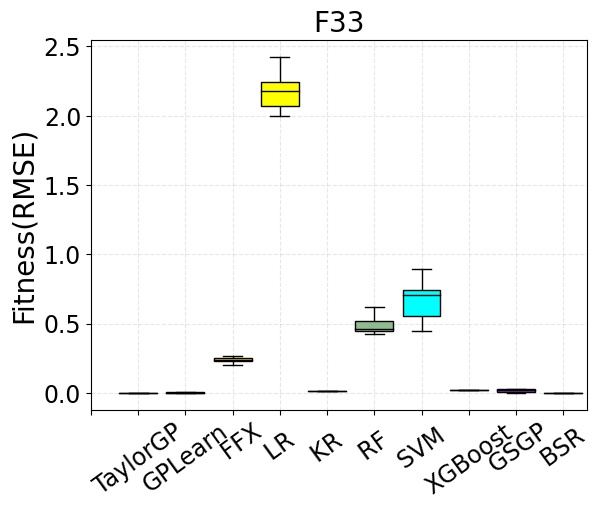}
	}
    \hspace{-0.5cm} \quad
		\subfigure{ \includegraphics[width=3cm]{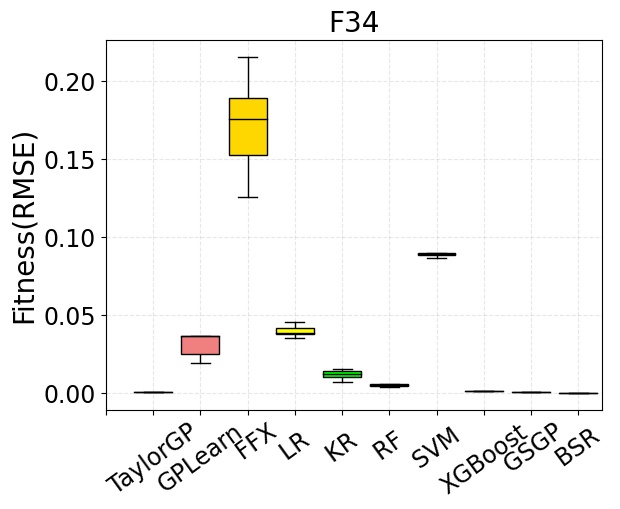}
	}
    \hspace{-0.5cm} \quad
		\subfigure{ \includegraphics[width=3cm]{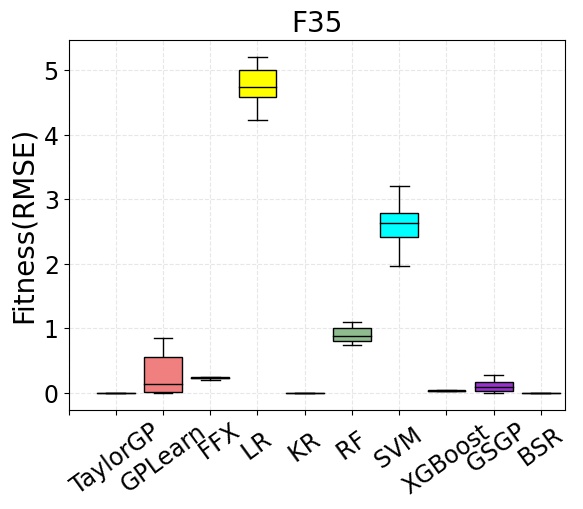}
	}
    \hspace{-0.5cm} \quad
		\subfigure{ \includegraphics[width=3cm]{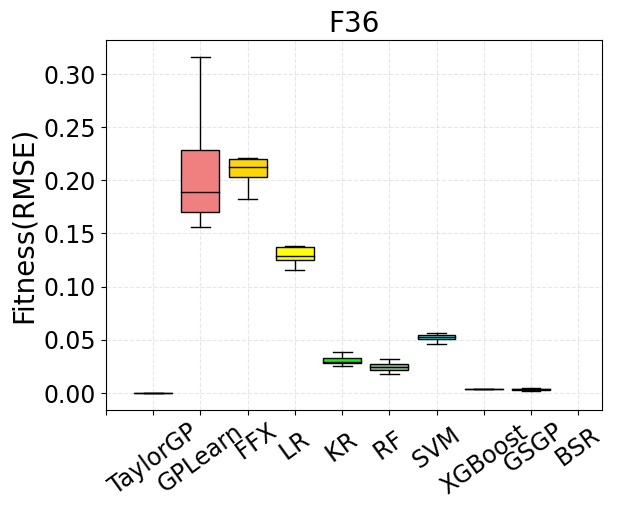}
	}
    \hspace{-0.5cm} \quad
		\subfigure{ \includegraphics[width=3cm]{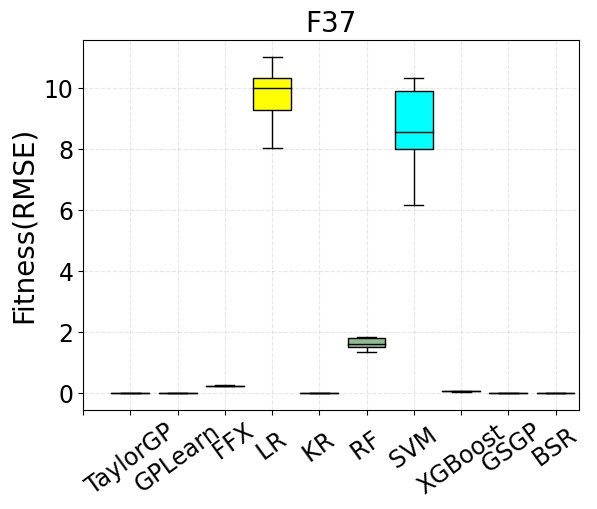}
	}
    \hspace{-0.5cm} \quad
		\subfigure{ \includegraphics[width=3cm]{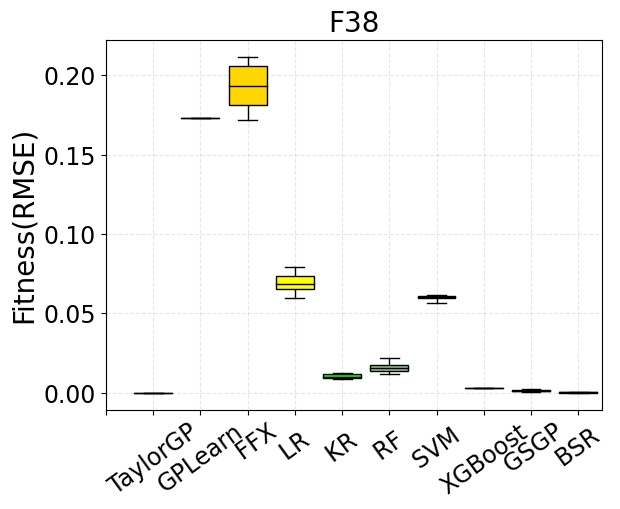}
	}
    \hspace{-0.5cm} \quad
		\subfigure{ \includegraphics[width=3cm]{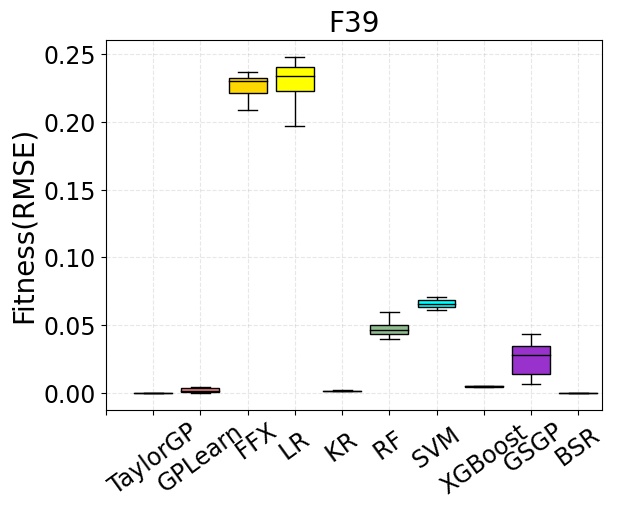}
	}
    \hspace{-0.5cm} \quad
		\subfigure{ \includegraphics[width=3cm]{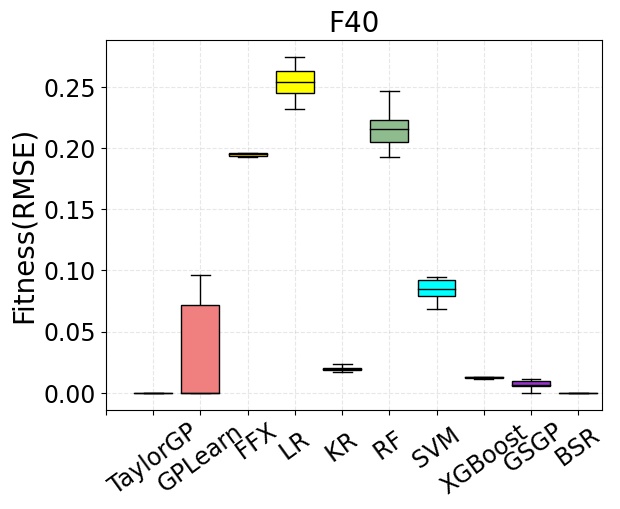}
	}
    \hspace{-0.5cm} \quad
		\subfigure{ \includegraphics[width=3cm]{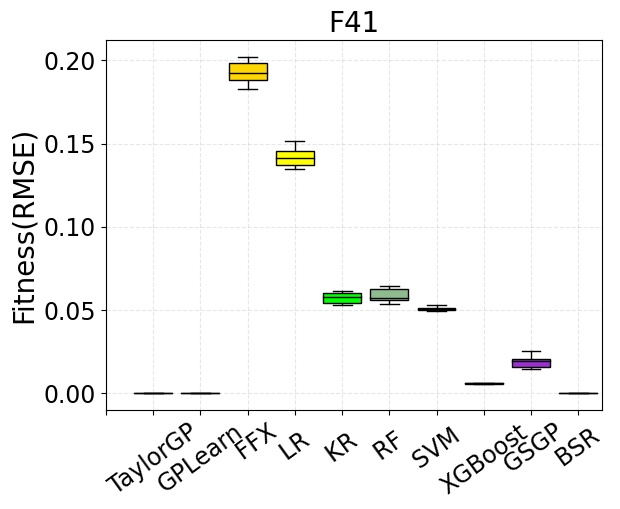}
	}
    \hspace{-0.5cm} \quad
		\subfigure{ \includegraphics[width=3cm]{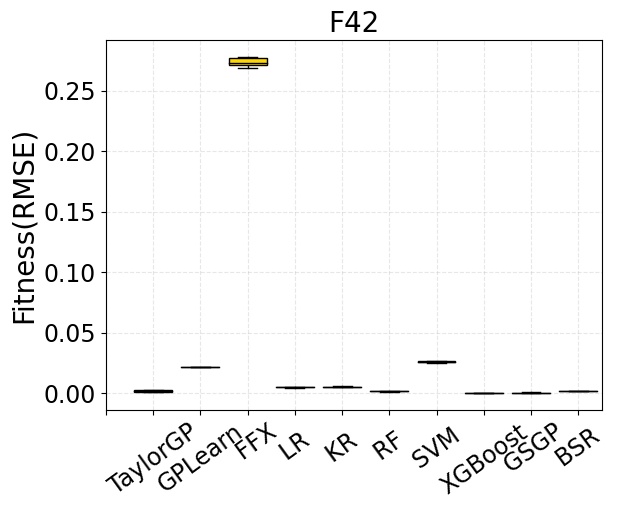}
	}
    \hspace{-0.5cm} \quad
		\subfigure{ \includegraphics[width=3cm]{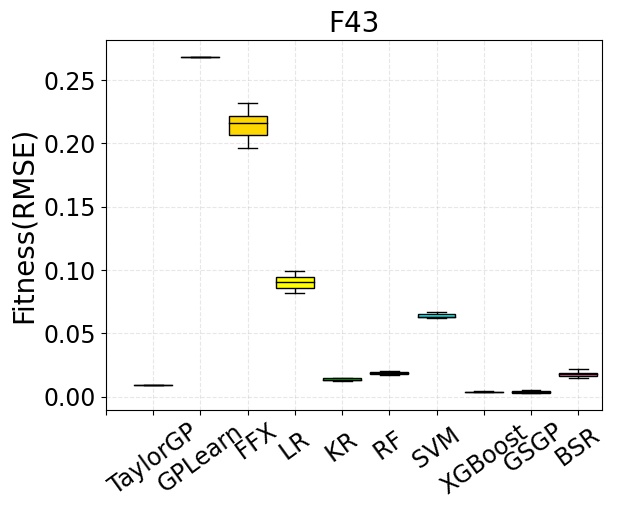}
	}
    \hspace{-0.5cm} \quad
		\subfigure{ \includegraphics[width=3cm]{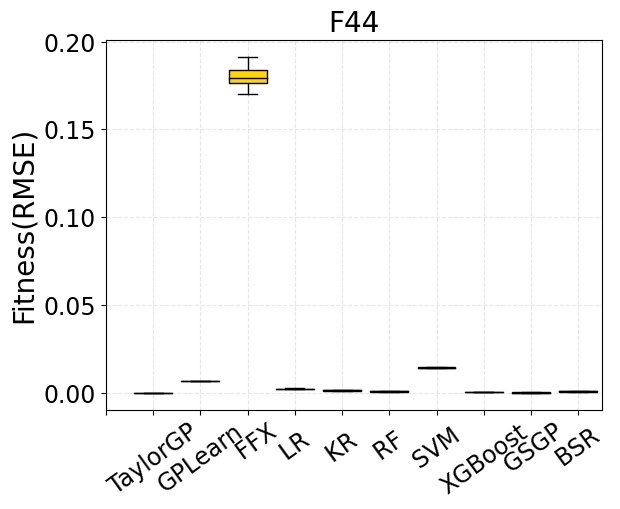}
	}
    \hspace{-0.5cm} \quad
		\subfigure{ \includegraphics[width=3cm]{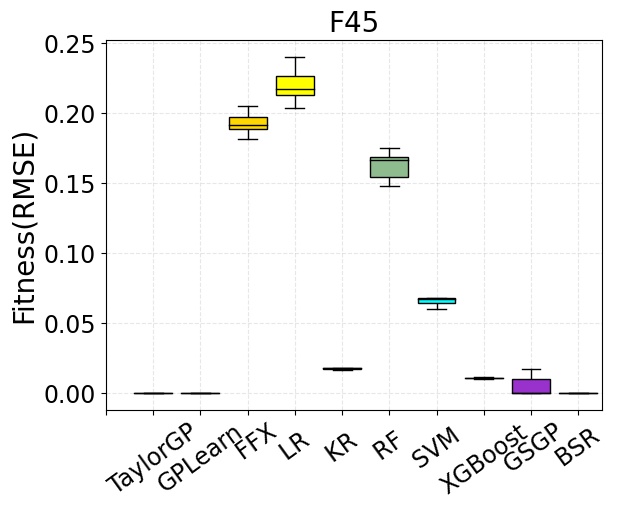}
	}
    \hspace{-0.5cm} \quad
		\subfigure{ \includegraphics[width=3cm]{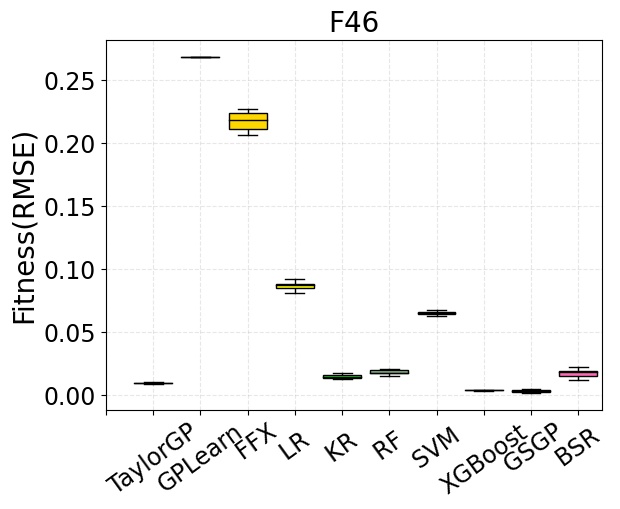}
	}
    \hspace{-0.5cm} \quad
		\subfigure{ \includegraphics[width=3cm]{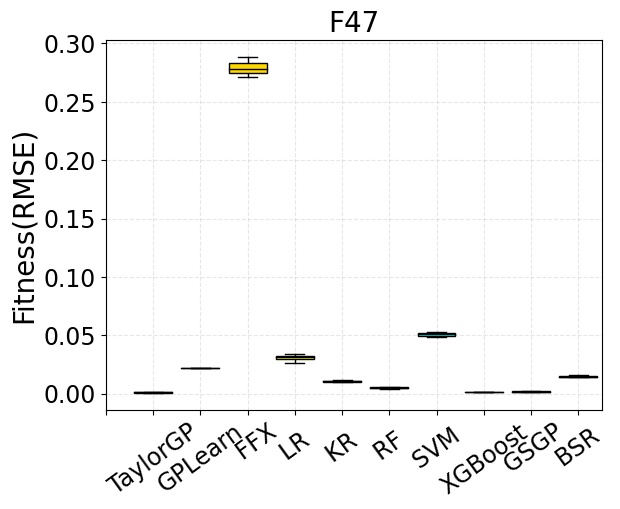}
	}
    \hspace{-0.5cm} \quad
		\subfigure{ \includegraphics[width=3cm]{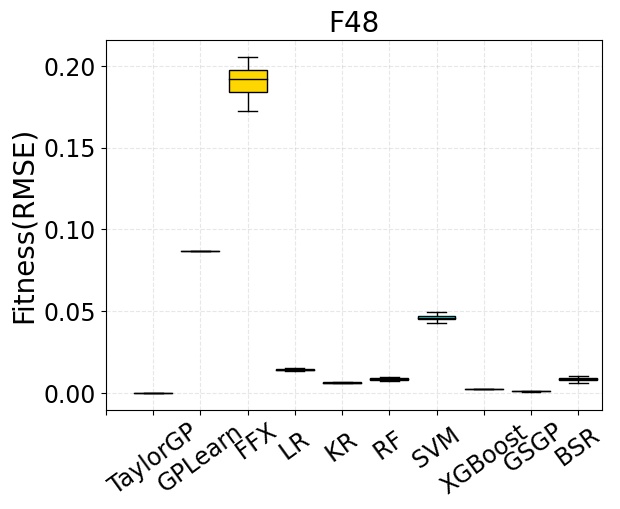}
	}
    \hspace{-0.5cm} \quad
		\subfigure{ \includegraphics[width=3cm]{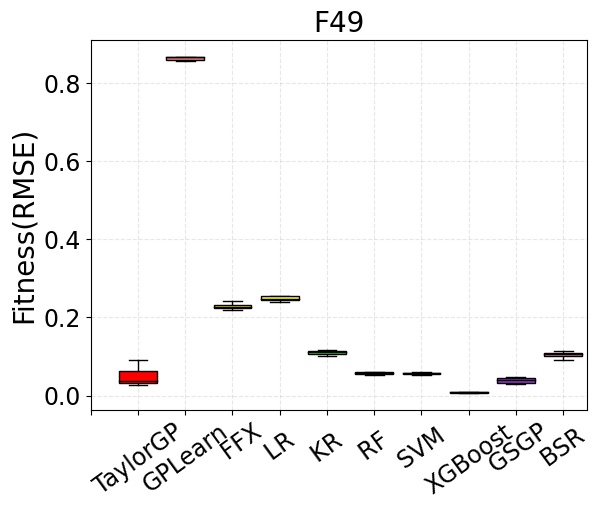}
	}
    \hspace{-0.5cm} \quad
		\subfigure{ \includegraphics[width=3cm]{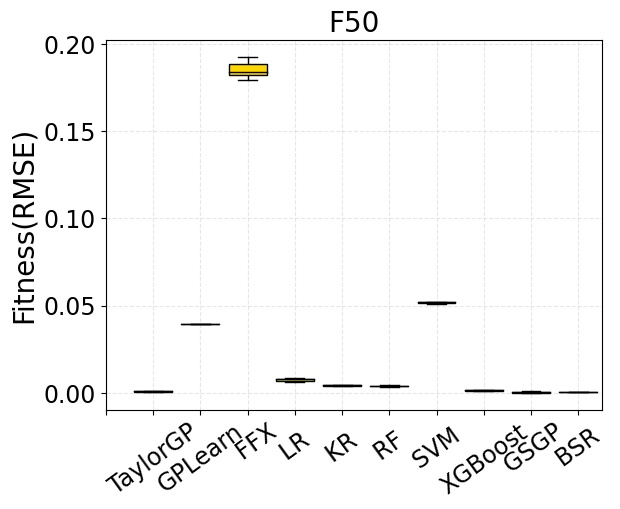}
	}

	    \hspace{-0.5cm} \quad
		\subfigure{ \includegraphics[width=3cm]{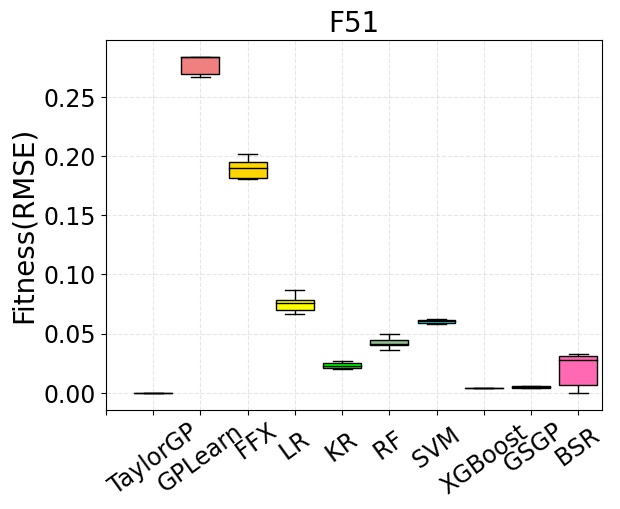}
	}
    \hspace{-0.5cm} \quad
		\subfigure{ \includegraphics[width=3cm]{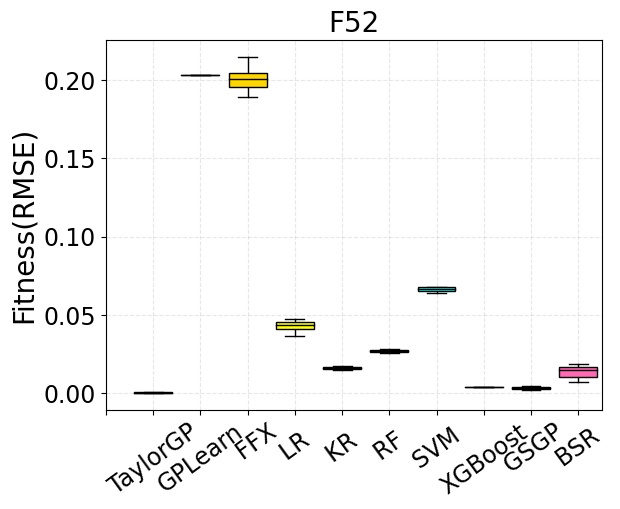}
	}
    \hspace{-0.5cm} \quad
		\subfigure{ \includegraphics[width=3cm]{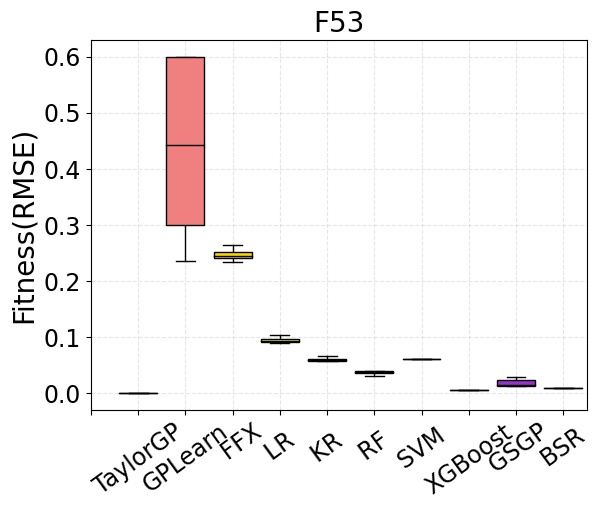}
	}
    \hspace{-0.5cm} \quad
		\subfigure{ \includegraphics[width=3cm]{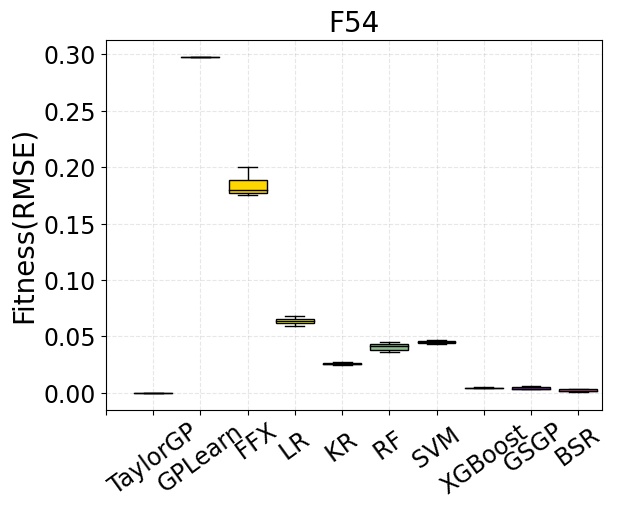}
	}
	    \hspace{-0.5cm} \quad
		\subfigure{ \includegraphics[width=3cm]{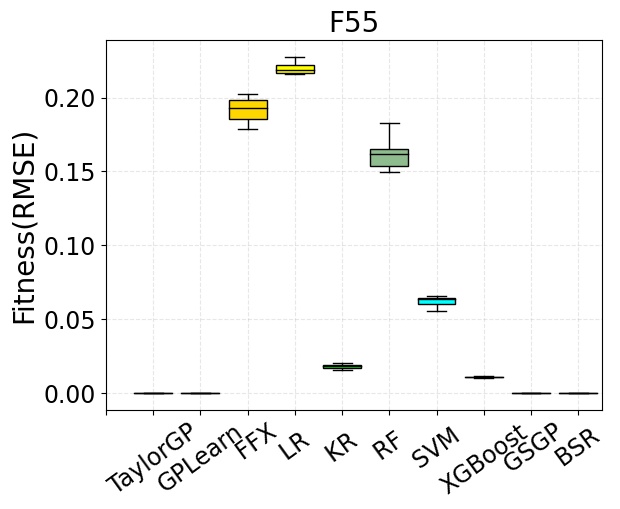}
	}
    \hspace{-0.5cm} \quad
		\subfigure{ \includegraphics[width=3cm]{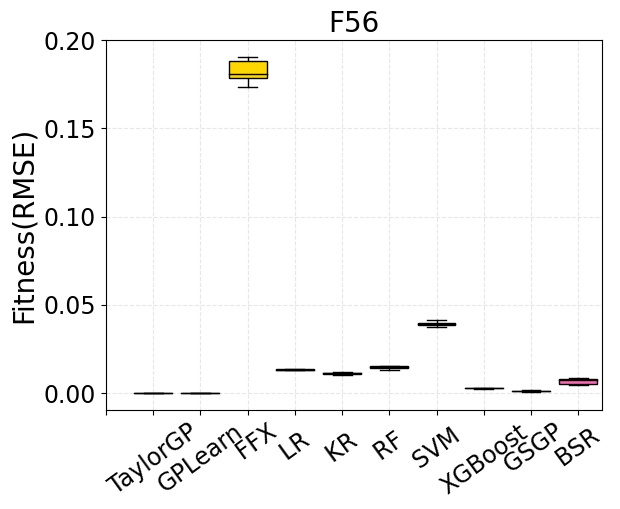}
	}
    \hspace{-0.5cm} \quad
		\subfigure{ \includegraphics[width=3cm]{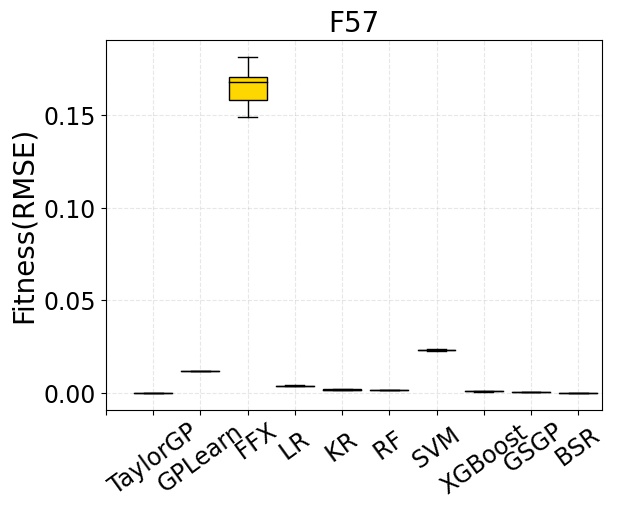}
	}
    \hspace{-0.5cm} \quad
		\subfigure{ \includegraphics[width=3cm]{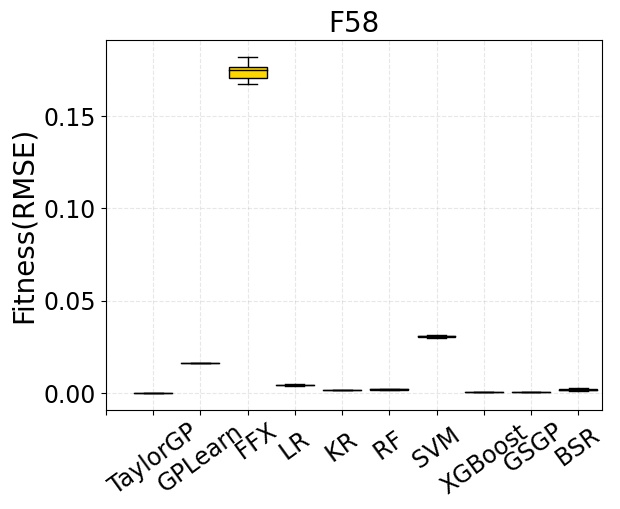}
	}
    \hspace{-0.5cm} \quad
		\subfigure{ \includegraphics[width=3cm]{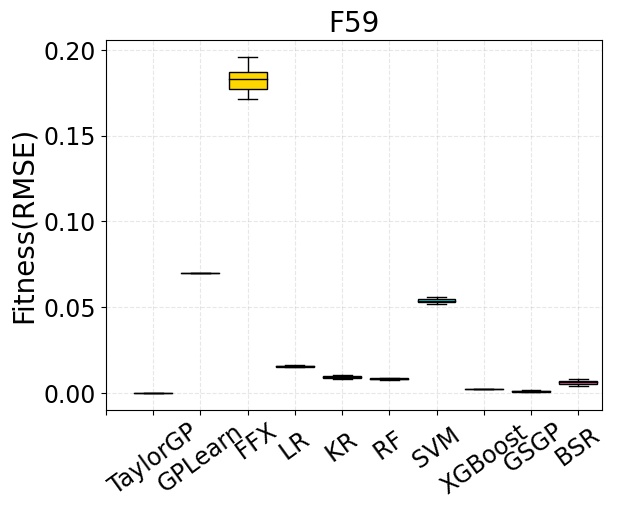}
	}
    \hspace{-0.5cm} \quad
		\subfigure{ \includegraphics[width=3cm]{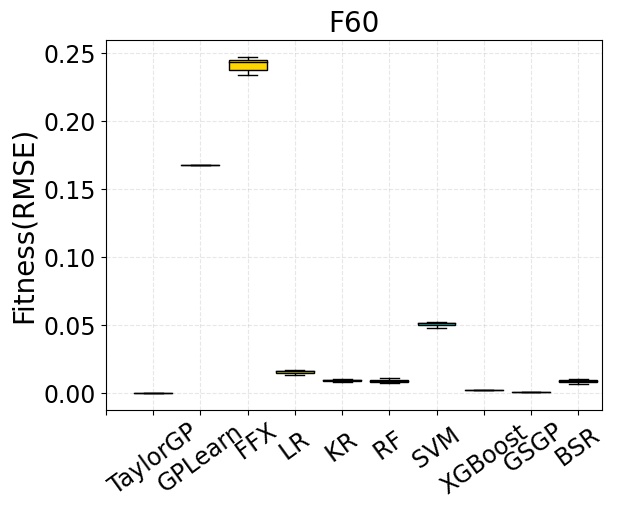}
	}

	\caption{Fitness comparison. The RMSE results on the benchmarks F31-F60.}
	\label{fig:overallBox2}
\end{figure*}

\begin{figure*}[htb]
	\centering

    \hspace{-0.5cm} \quad
		\subfigure{ \includegraphics[width=3cm]{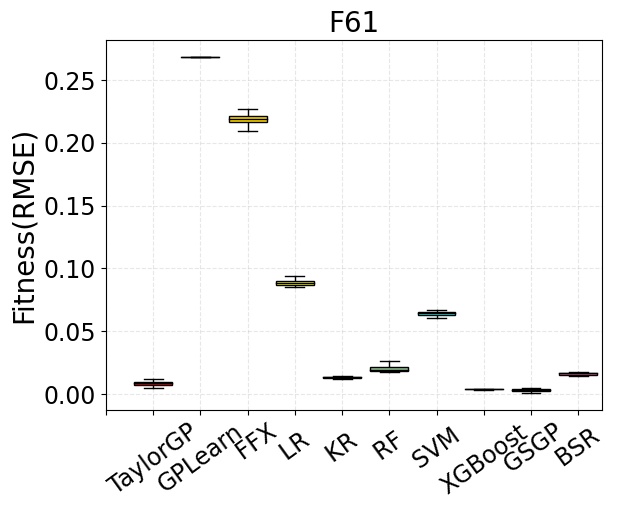}
	}
    \hspace{-0.5cm} \quad
		\subfigure{ \includegraphics[width=3cm]{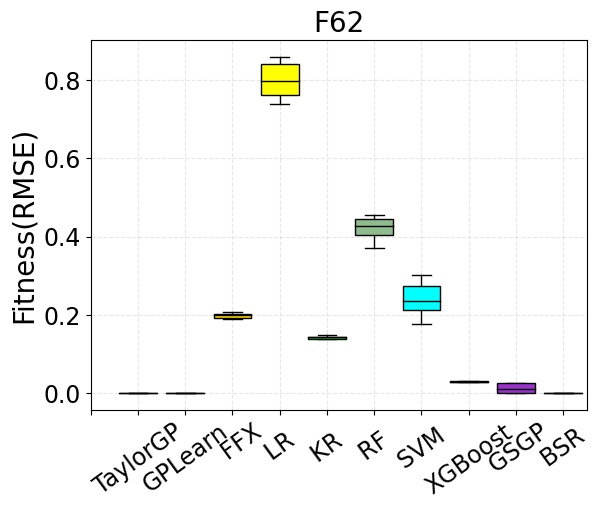}
	}
    \hspace{-0.5cm} \quad
		\subfigure{ \includegraphics[width=3cm]{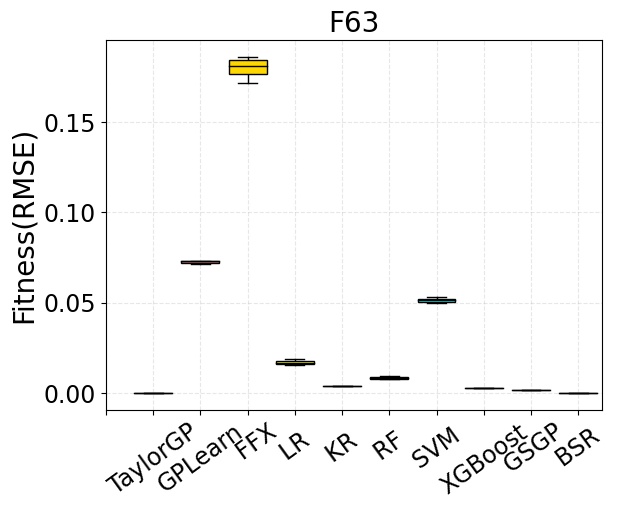}
	}
    \hspace{-0.5cm} \quad
		\subfigure{ \includegraphics[width=3cm]{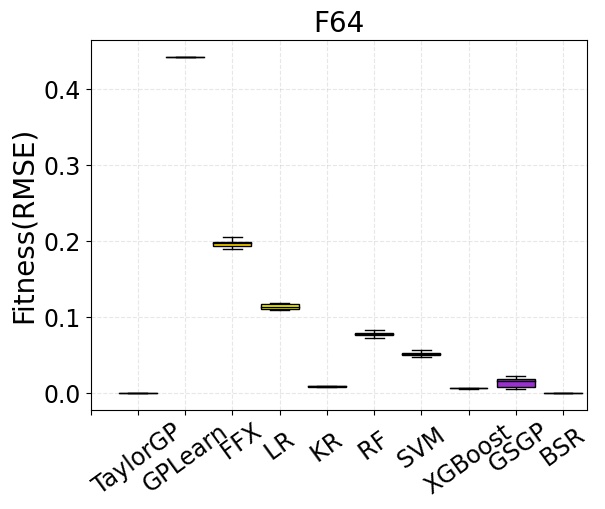}
	}
    \hspace{-0.5cm} \quad
		\subfigure{ \includegraphics[width=3cm]{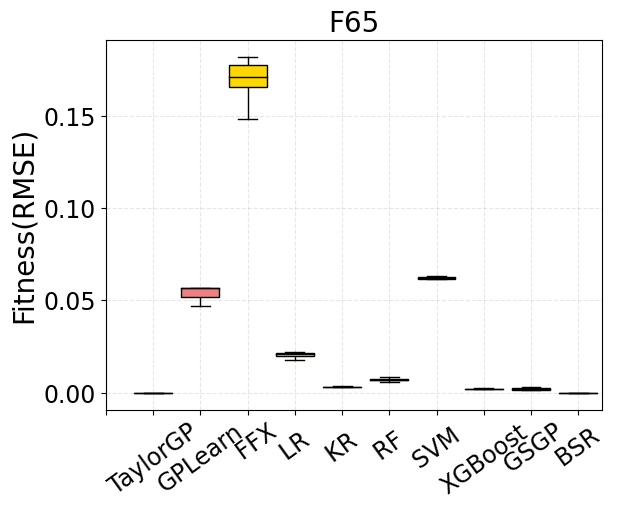}
	}
    \hspace{-0.5cm} \quad
		\subfigure{ \includegraphics[width=3cm]{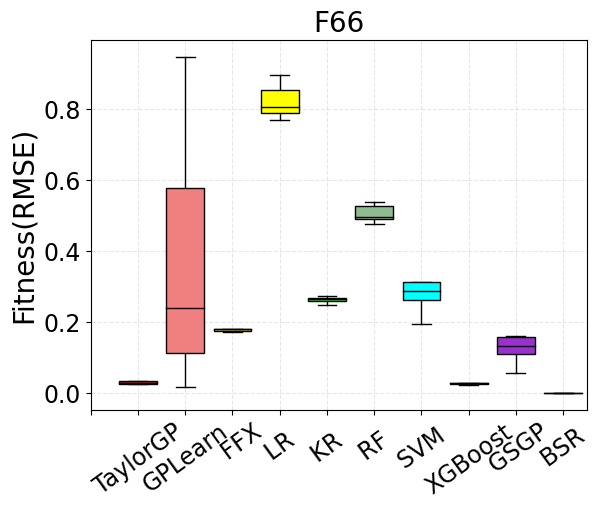}
	}
    \hspace{-0.5cm} \quad
		\subfigure{ \includegraphics[width=3cm]{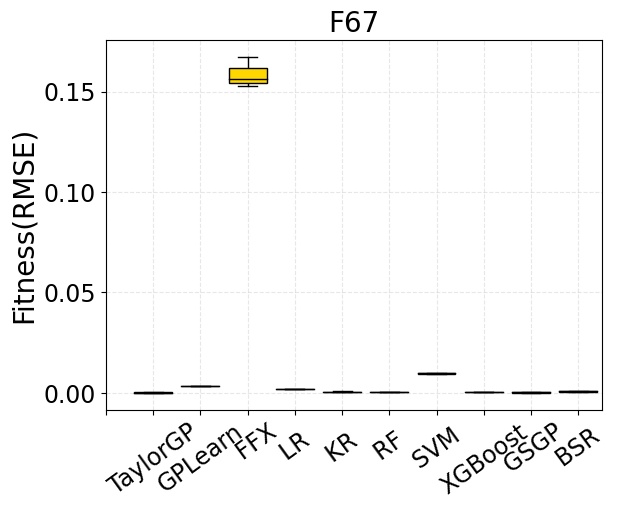}
	}
    \hspace{-0.5cm} \quad
		\subfigure{ \includegraphics[width=3cm]{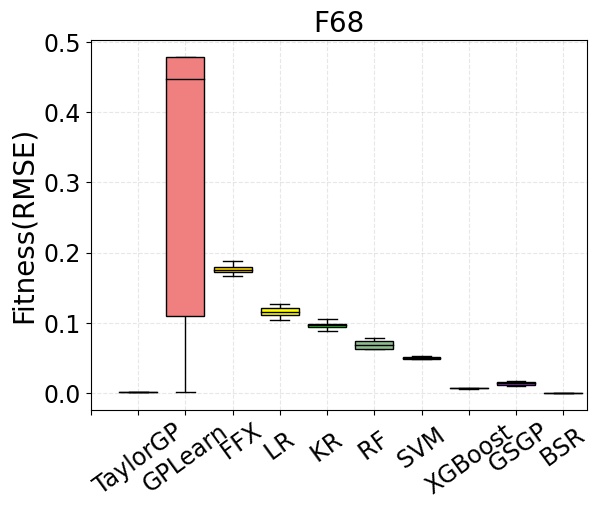}
	}
    \hspace{-0.5cm} \quad
		\subfigure{ \includegraphics[width=3cm]{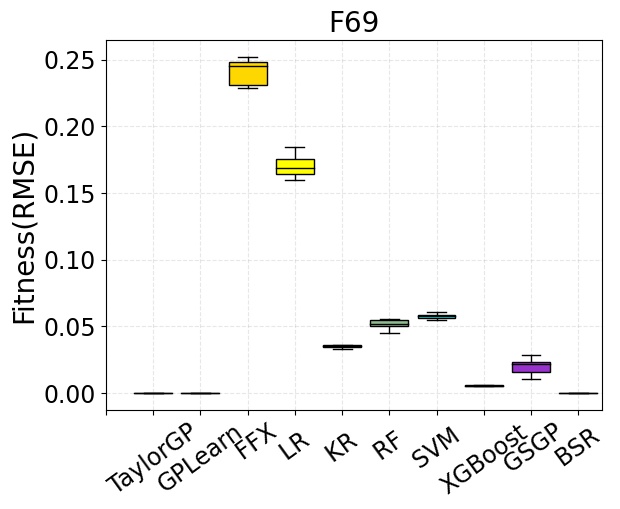}
	}
    \hspace{-0.5cm} \quad
		\subfigure{ \includegraphics[width=3cm]{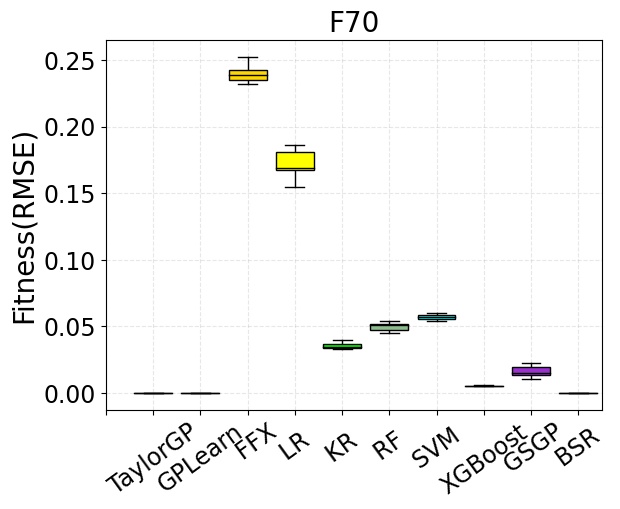}
	}
    \hspace{-0.5cm} \quad
		\subfigure{ \includegraphics[width=3cm]{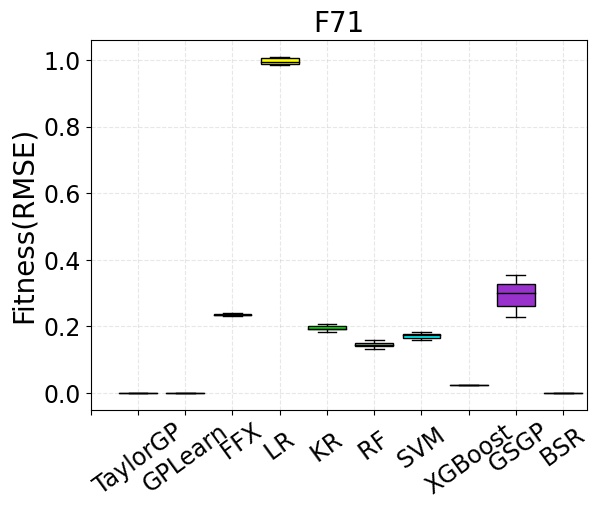}
	}
    \hspace{-0.5cm} \quad
		\subfigure{ \includegraphics[width=3cm]{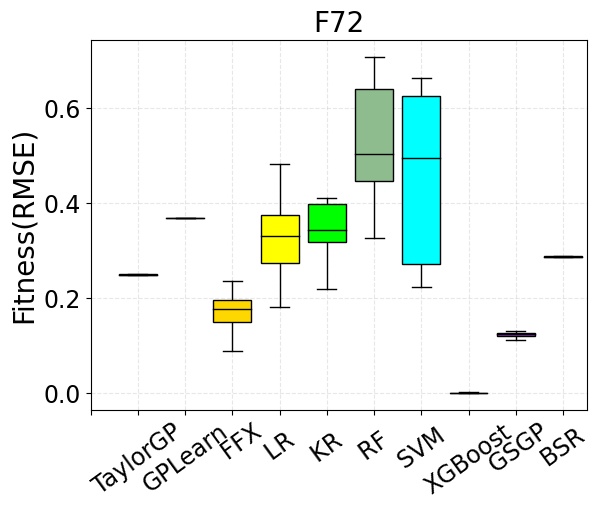}
	}
    \hspace{-0.5cm} \quad
		\subfigure{ \includegraphics[width=3cm]{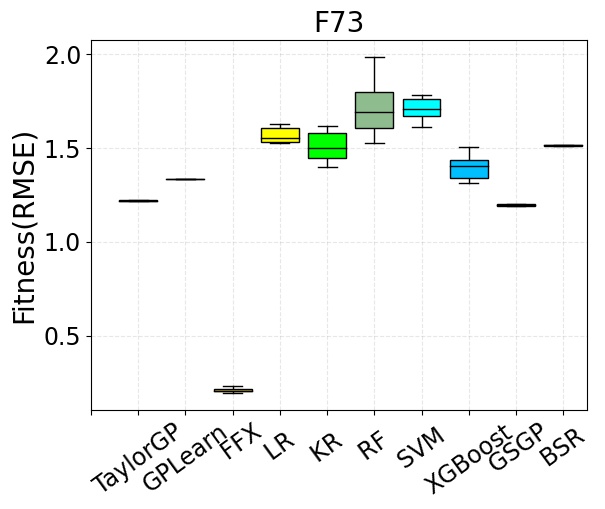}
	}
    \hspace{-0.5cm} \quad
		\subfigure{ \includegraphics[width=3cm]{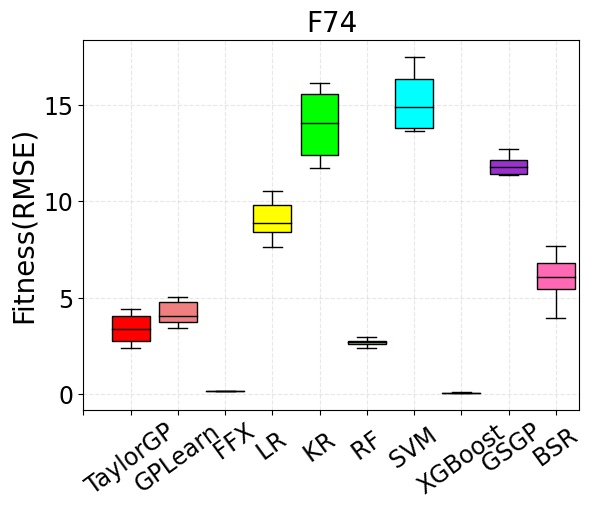}
	}
    \hspace{-0.5cm} \quad
		\subfigure{ \includegraphics[width=3cm]{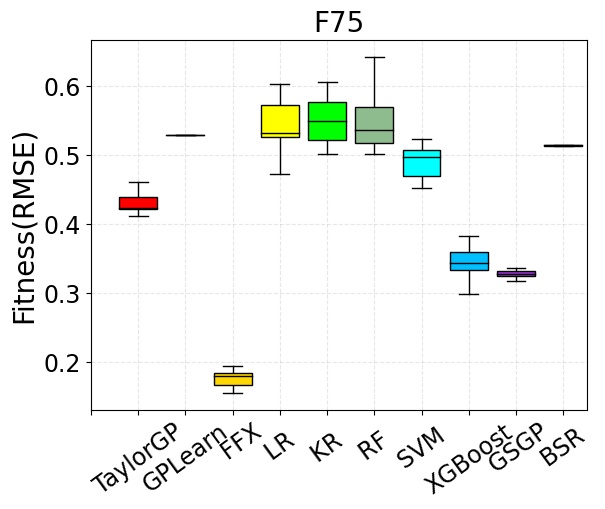}
	}
    \hspace{-0.5cm} \quad
		\subfigure{ \includegraphics[width=3cm]{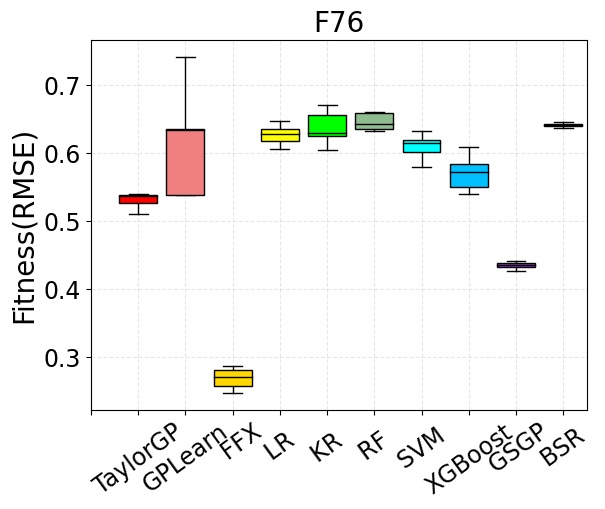}
	}
    \hspace{-0.5cm} \quad
		\subfigure{ \includegraphics[width=3cm]{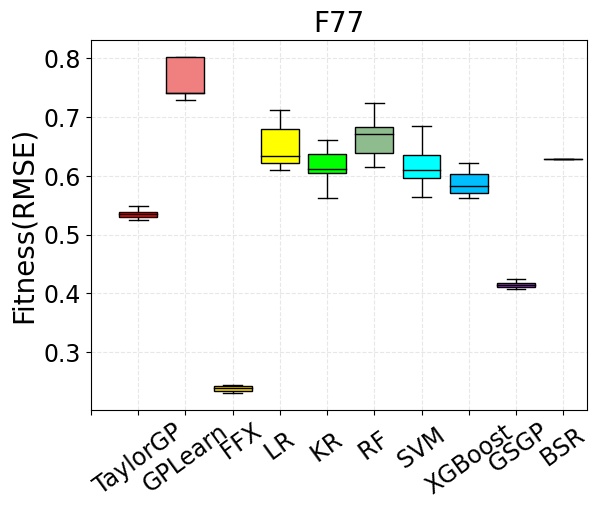}
	}
    \hspace{-0.5cm} \quad
		\subfigure{ \includegraphics[width=3cm]{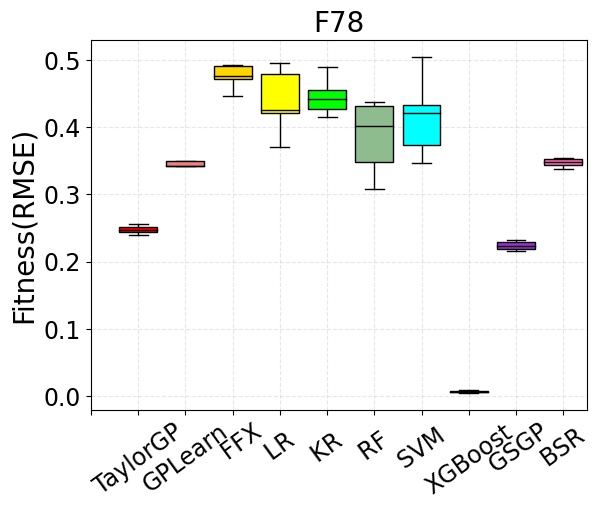}
	}
    \hspace{-0.5cm} \quad
		\subfigure{ \includegraphics[width=3cm]{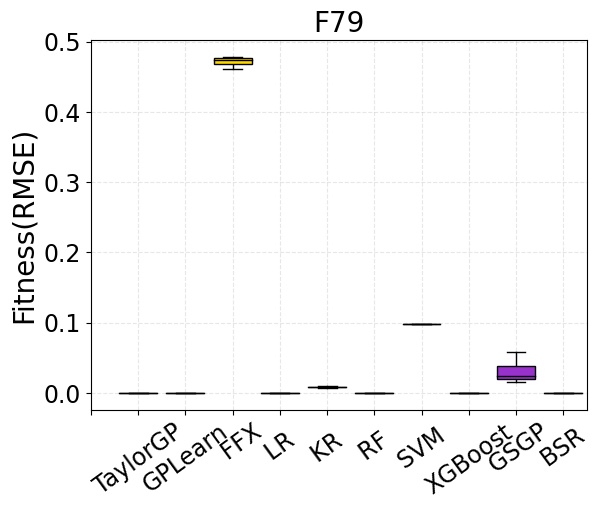}
	}
    \hspace{-0.5cm} \quad
		\subfigure{ \includegraphics[width=3cm]{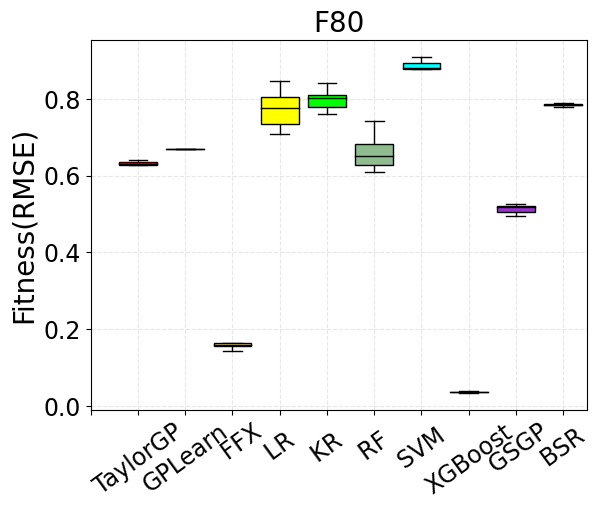}
	}
    \hspace{-0.5cm} \quad
		\subfigure{\includegraphics[width=3cm]{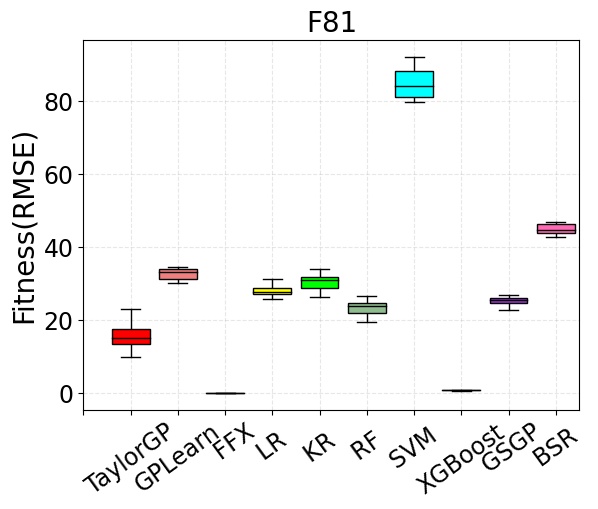}
	}

	\caption{Fitness comparison. The RMSE results on the benchmarks  F61-F81.}
	\label{fig:overallBox3}
\end{figure*}

\begin{figure*}[htb]
	\centering
    \hspace{-0.5cm} \quad
		\subfigure{ \includegraphics[width=3cm]{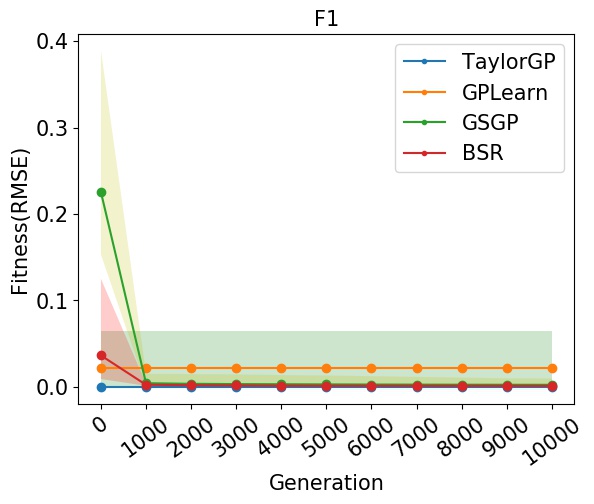}
	}
	\hspace{-0.5cm} \quad
		\subfigure{ \includegraphics[width=3cm]{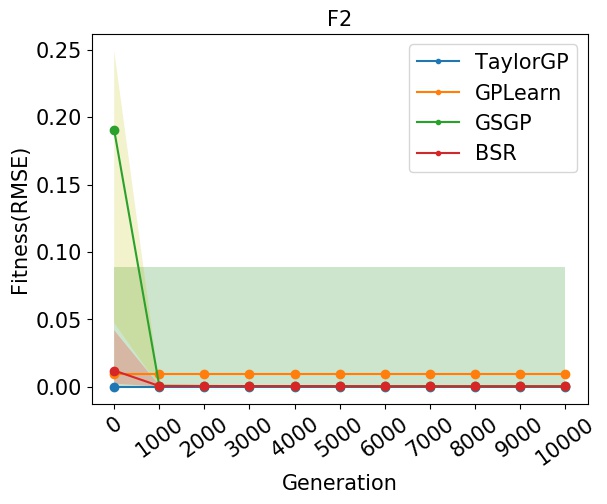}
    }
    \hspace{-0.5cm} \quad
		\subfigure{ \includegraphics[width=3cm]{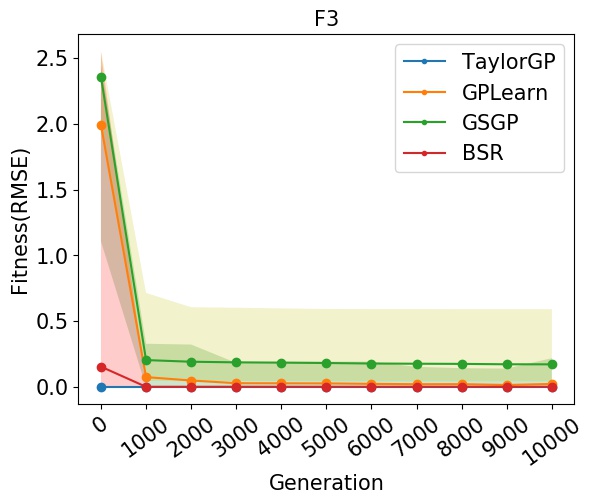}
	}
	\hspace{-0.5cm} \quad
		\subfigure{ \includegraphics[width=3cm]{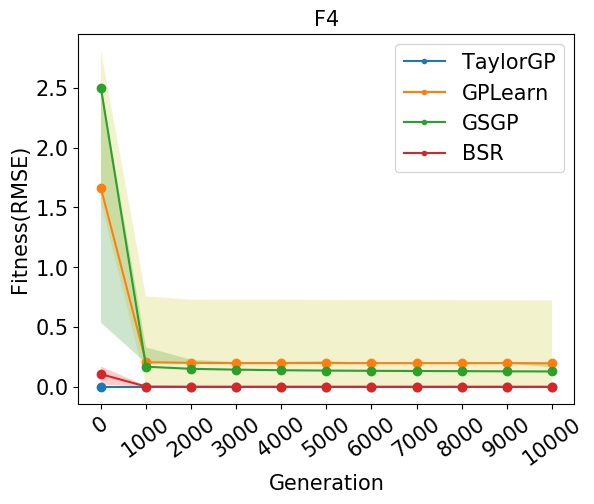}
    }
    \hspace{-0.5cm} \quad
		\subfigure{ \includegraphics[width=3cm]{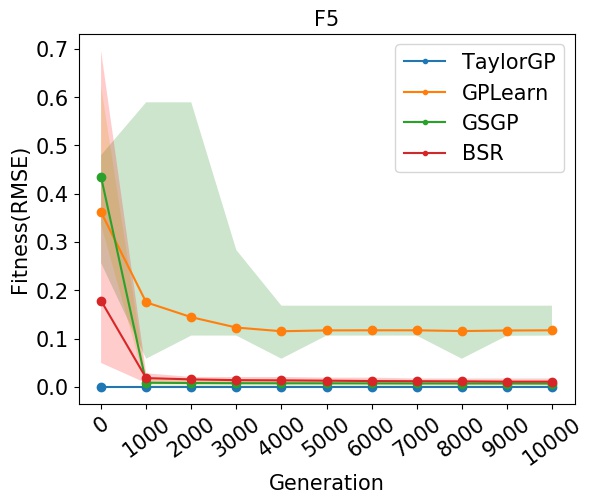}
	}
	\hspace{-0.5cm} \quad
		\subfigure{ \includegraphics[width=3cm]{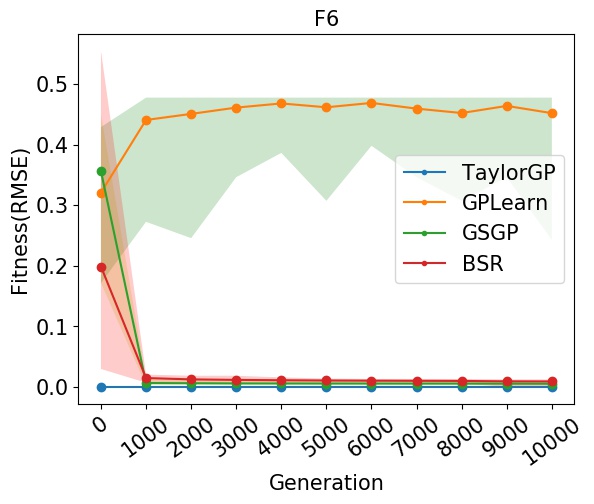}
    }
    \hspace{-0.5cm} \quad
		\subfigure{ \includegraphics[width=3cm]{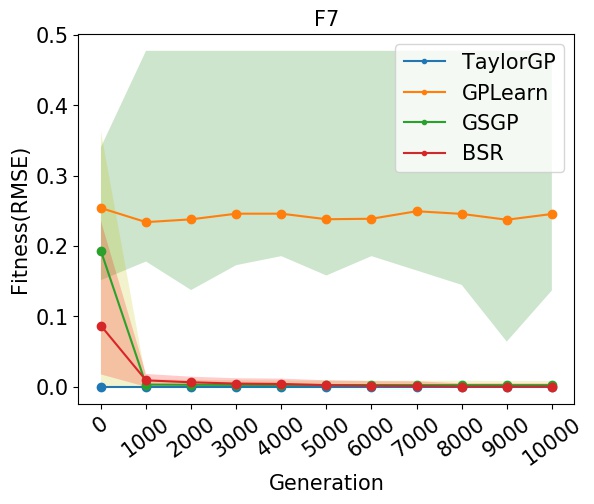}
	}
	\hspace{-0.5cm} \quad
		\subfigure{ \includegraphics[width=3cm]{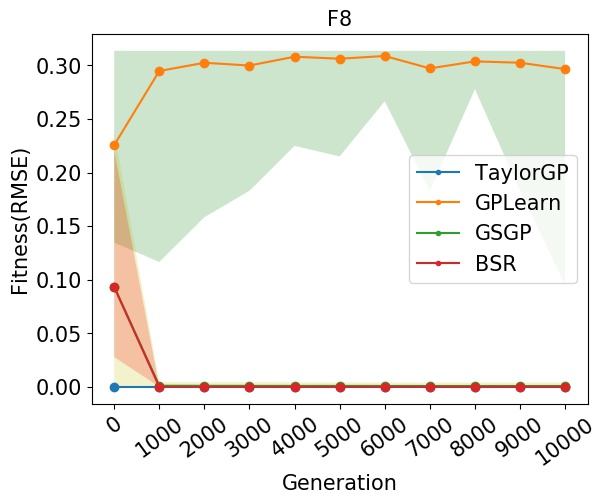}
    }
    \hspace{-0.5cm} \quad
		\subfigure{ \includegraphics[width=3cm]{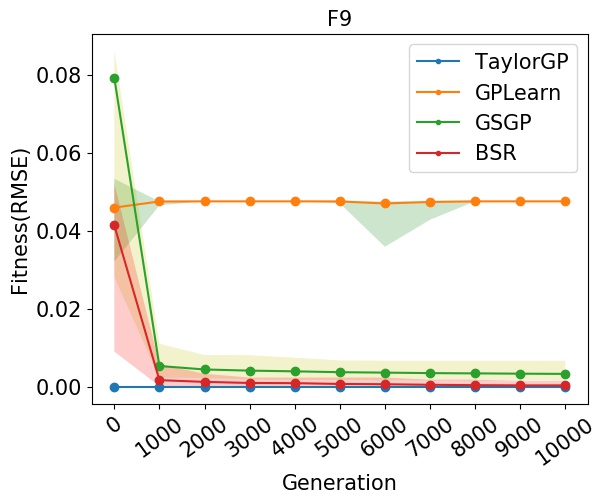}
	}
	\hspace{-0.5cm} \quad
		\subfigure{ \includegraphics[width=3cm]{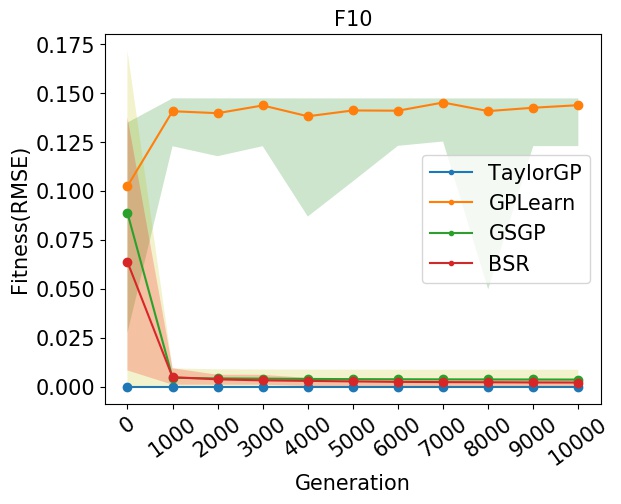}
    }
    \hspace{-0.5cm} \quad
		\subfigure{ \includegraphics[width=3cm]{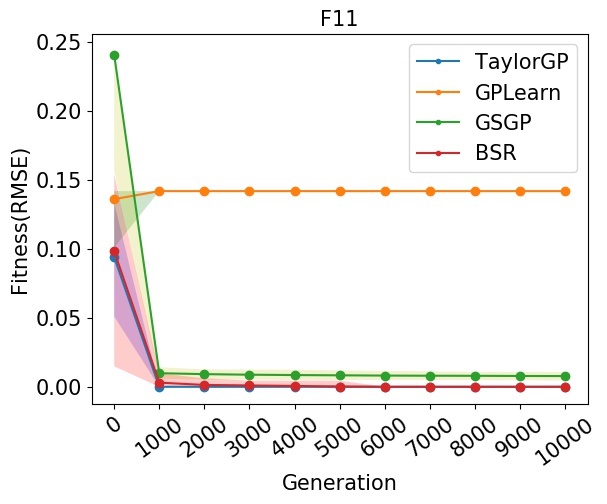}
	}
	\hspace{-0.5cm} \quad
		\subfigure{ \includegraphics[width=3cm]{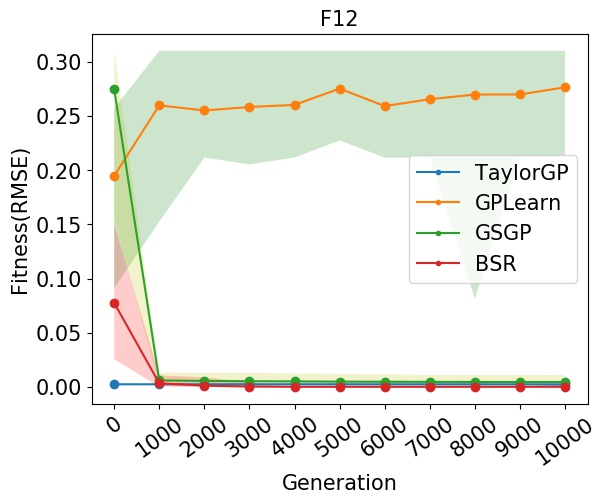}
    }
    \hspace{-0.5cm} \quad
		\subfigure{ \includegraphics[width=3cm]{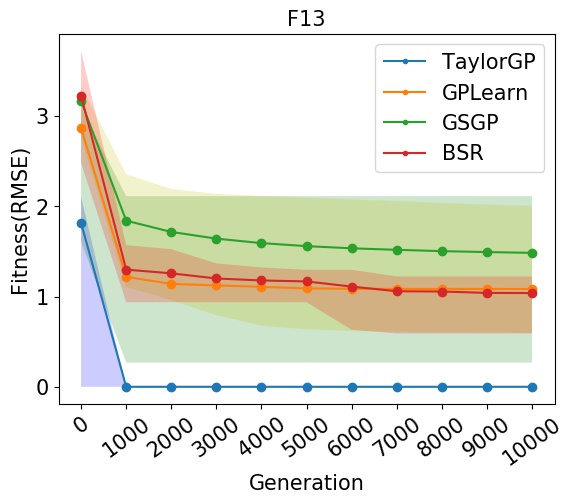}
	}
	\hspace{-0.5cm} \quad
		\subfigure{ \includegraphics[width=3cm]{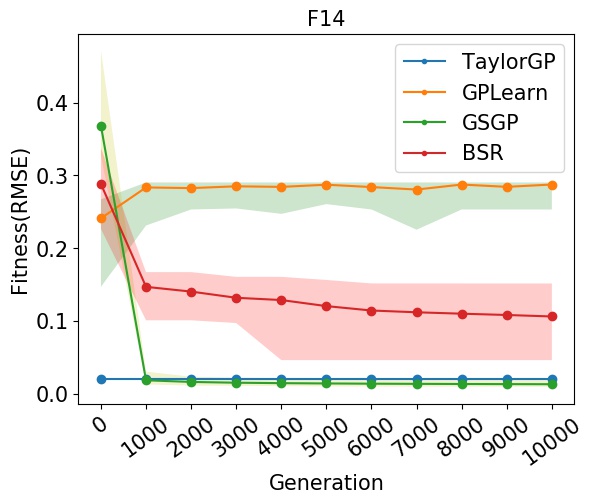}
    }
    \hspace{-0.5cm} \quad
		\subfigure{ \includegraphics[width=3cm]{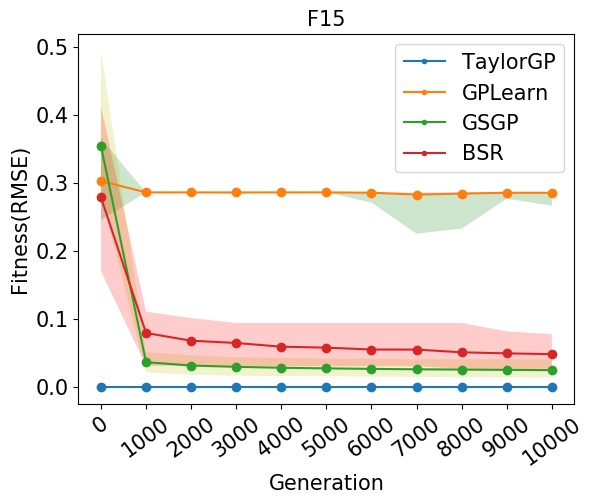}
	}
	\hspace{-0.5cm} \quad
		\subfigure{ \includegraphics[width=3cm]{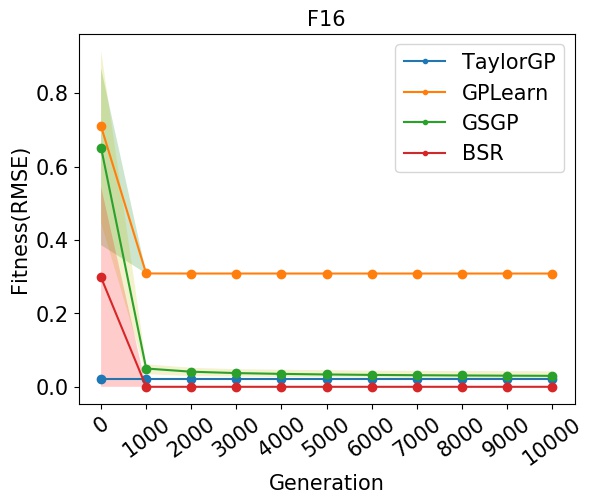}
    }
    \hspace{-0.5cm} \quad
		\subfigure{ \includegraphics[width=3cm]{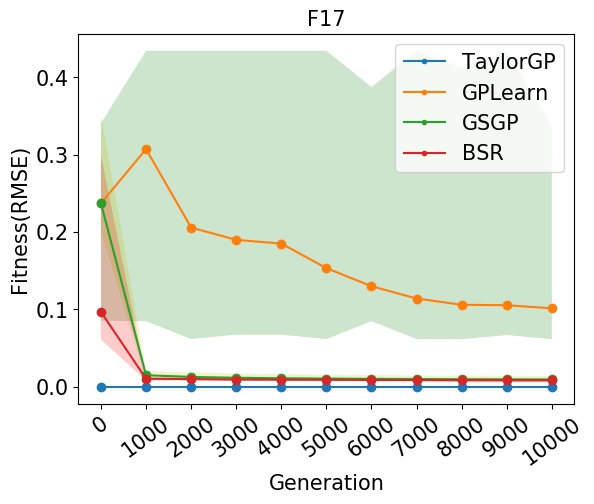}
	}
	\hspace{-0.5cm} \quad
		\subfigure{ \includegraphics[width=3cm]{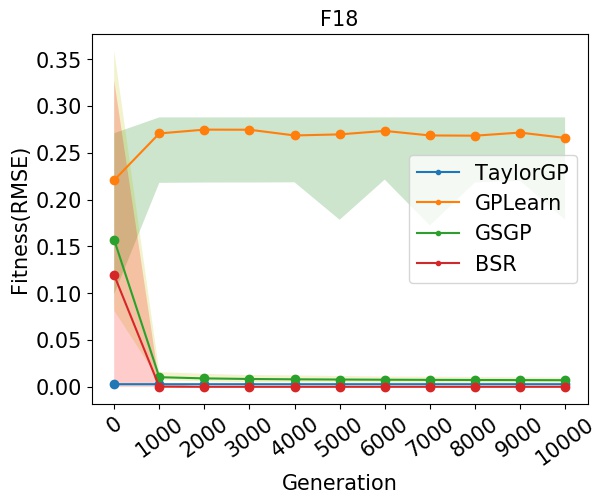}
    }
    \hspace{-0.5cm} \quad
		\subfigure{ \includegraphics[width=3cm]{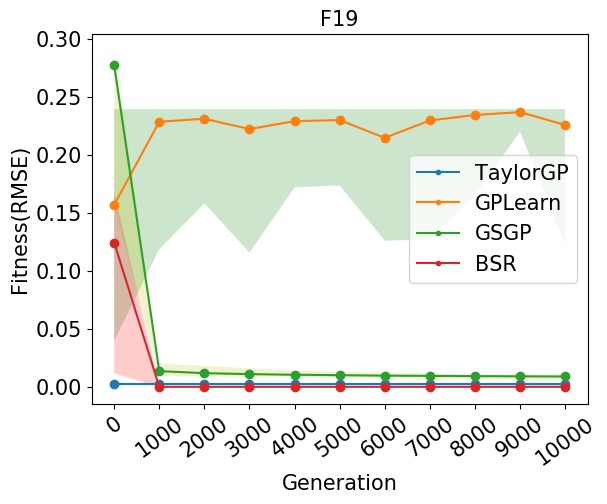}
	}
	\hspace{-0.5cm} \quad
		\subfigure{ \includegraphics[width=3cm]{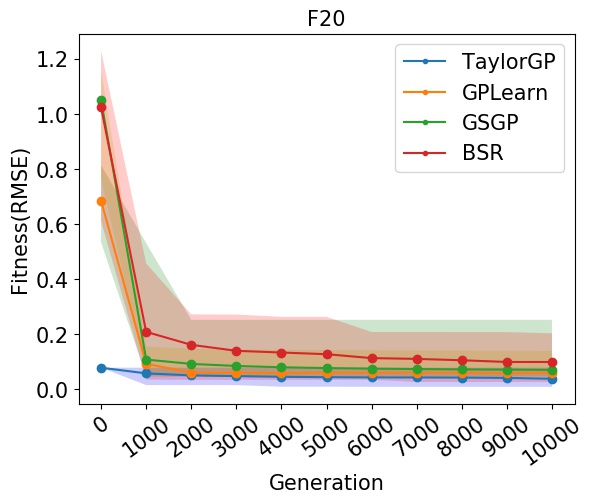}
    }
    \hspace{-0.5cm} \quad
		\subfigure{ \includegraphics[width=3cm]{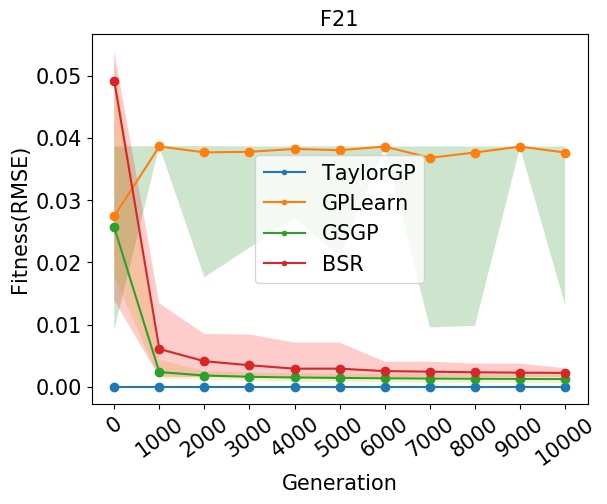}
	}
	\hspace{-0.5cm} \quad
		\subfigure{ \includegraphics[width=3cm]{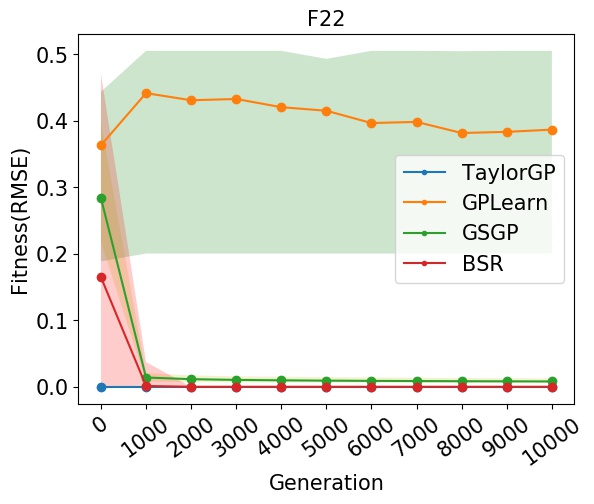}
    }
    \hspace{-0.5cm} \quad
		\subfigure{ \includegraphics[width=3cm]{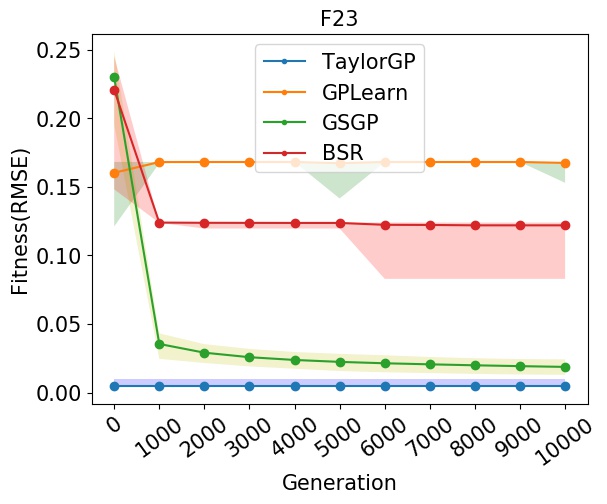}
	}
    \hspace{-0.5cm} \quad
		\subfigure{ \includegraphics[width=3cm]{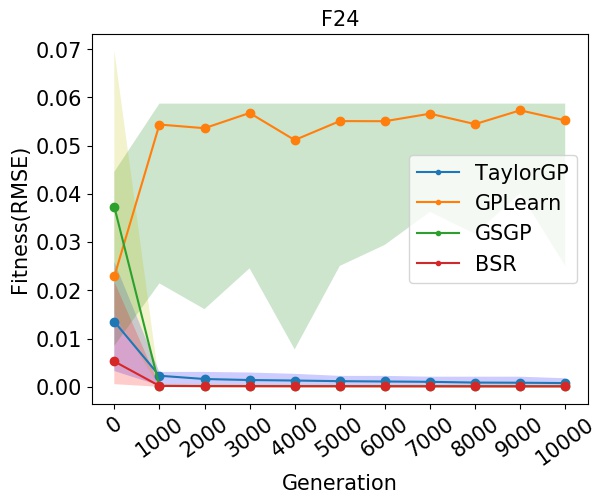}
	}
    \hspace{-0.5cm} \quad
		\subfigure{ \includegraphics[width=3cm]{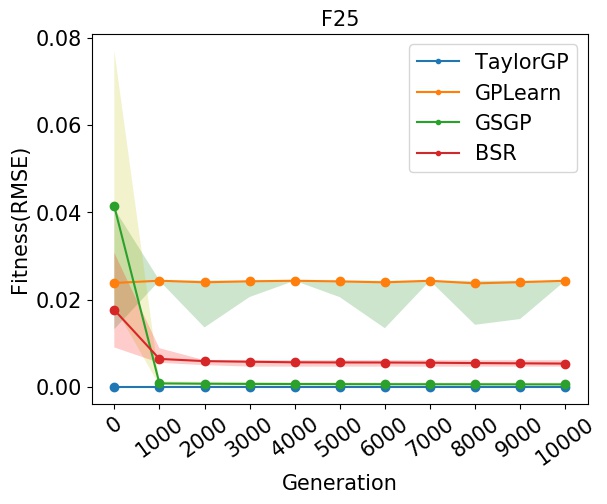}
	}
	    \hspace{-0.5cm} \quad
		\subfigure{ \includegraphics[width=3cm]{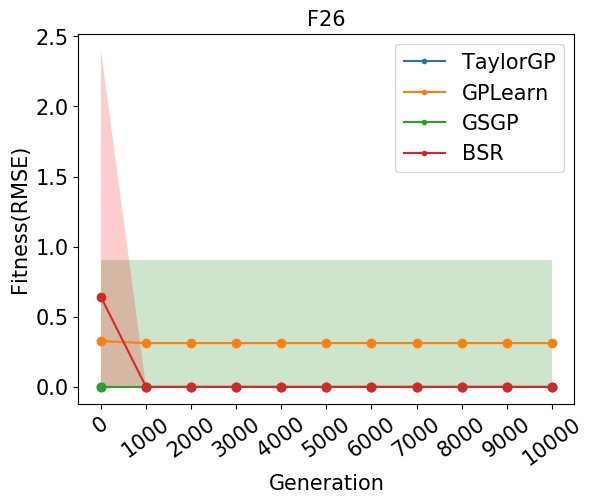}
	}
    \hspace{-0.5cm} \quad
		\subfigure{ \includegraphics[width=3cm]{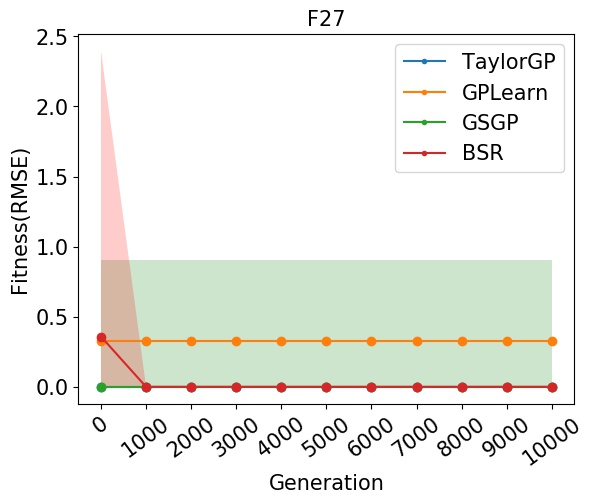}
	}
	    \hspace{-0.5cm} \quad
		\subfigure{ \includegraphics[width=3cm]{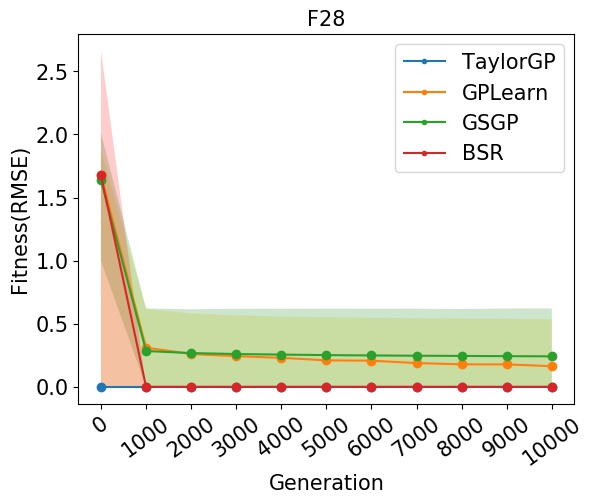}
	}
    \hspace{-0.5cm} \quad
		\subfigure{ \includegraphics[width=3cm]{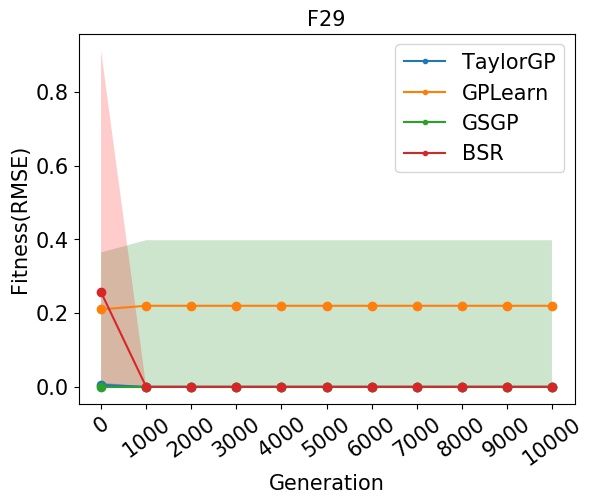}
	}
    \hspace{-0.5cm} \quad
		\subfigure{ \includegraphics[width=3cm]{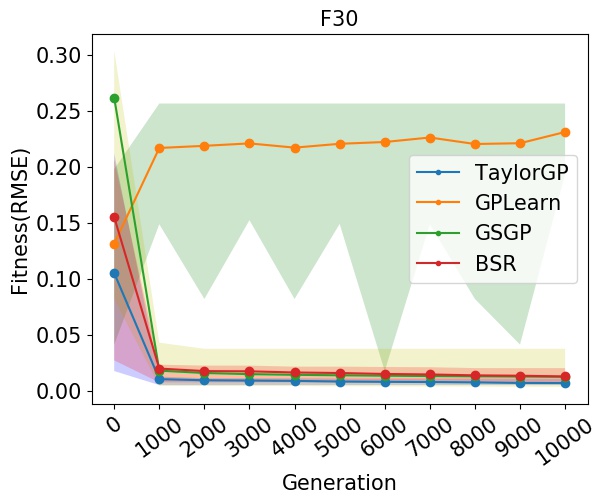}
	}

	\caption{Comparison of convergence. The fitness convergence curve of four algorithms on the benchmarks F1-F30.}
	\label{fig:fitness_converge1}
\end{figure*}

\begin{figure*}[htb]
	\centering

    \hspace{-0.5cm} \quad
		\subfigure{ \includegraphics[width=3cm]{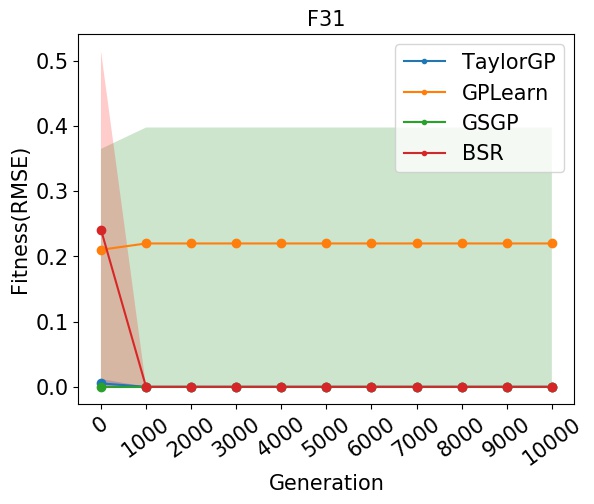}
	}
    \hspace{-0.5cm} \quad
		\subfigure{ \includegraphics[width=3cm]{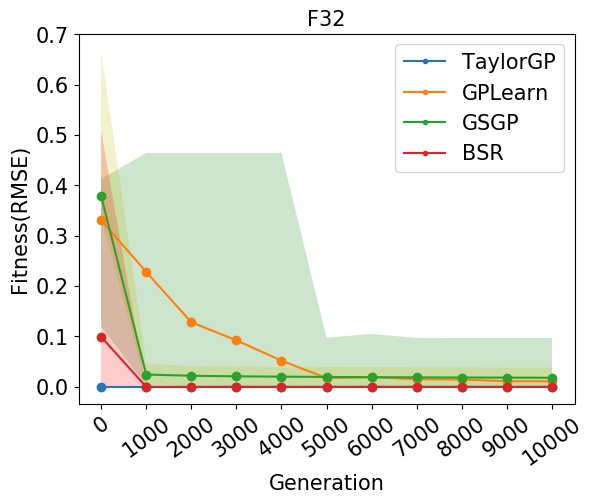}
	}
    \hspace{-0.5cm} \quad
		\subfigure{ \includegraphics[width=3cm]{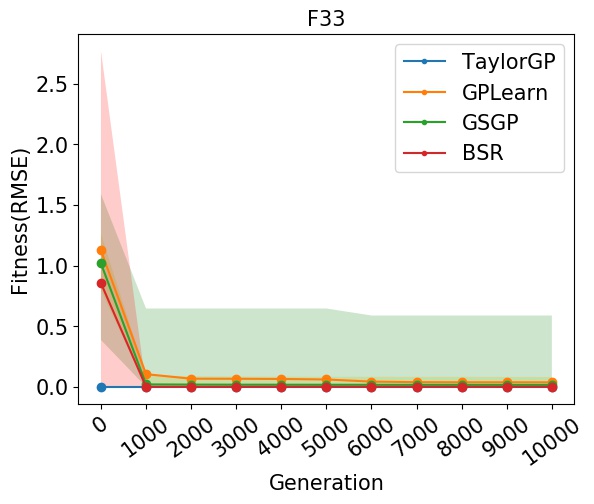}
	}
    \hspace{-0.5cm} \quad
		\subfigure{ \includegraphics[width=3cm]{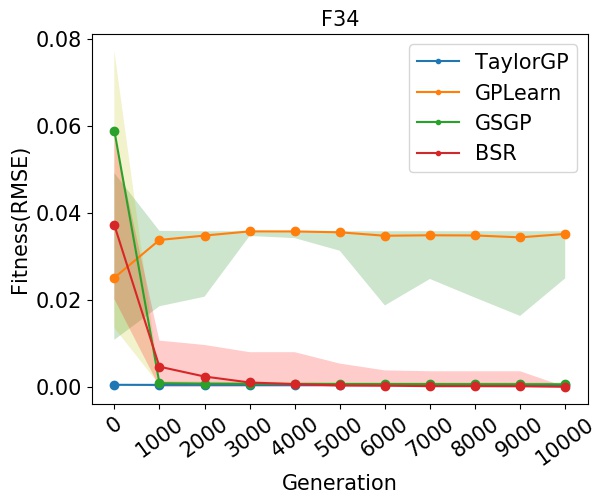}
	}
    \hspace{-0.5cm} \quad
		\subfigure{ \includegraphics[width=3cm]{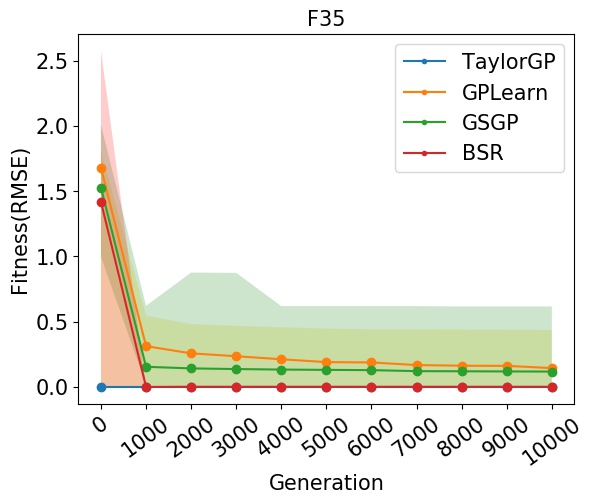}
	}
    \hspace{-0.5cm} \quad
		\subfigure{ \includegraphics[width=3cm]{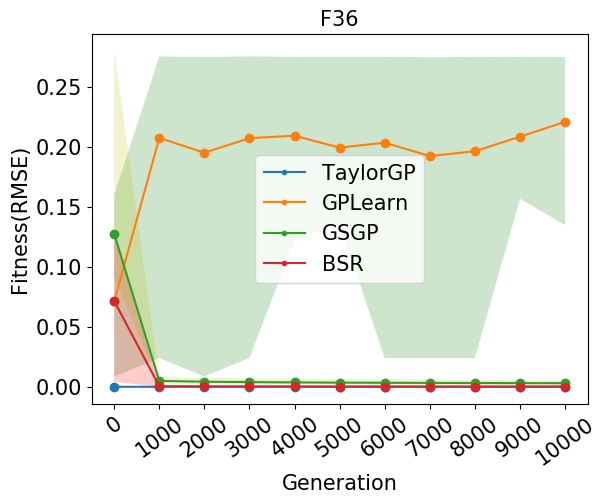}
	}
    \hspace{-0.5cm} \quad
		\subfigure{ \includegraphics[width=3cm]{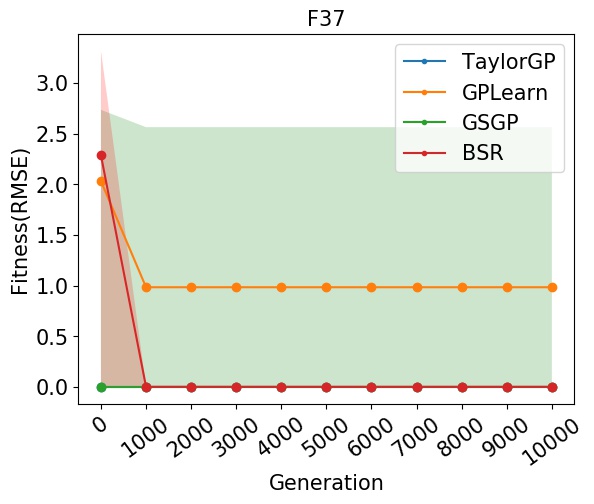}
	}
    \hspace{-0.5cm} \quad
		\subfigure{ \includegraphics[width=3cm]{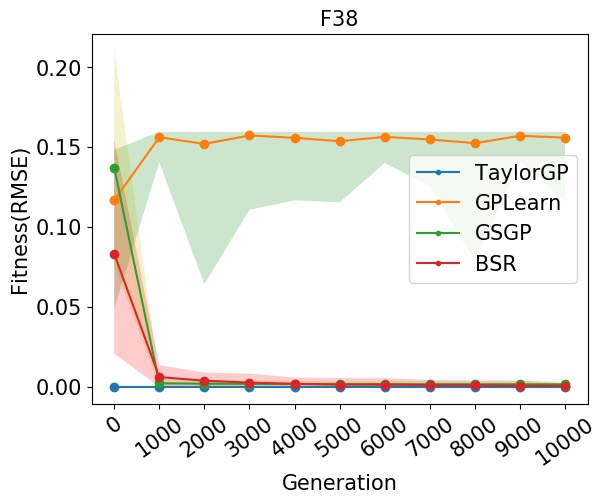}
	}
    \hspace{-0.5cm} \quad
		\subfigure{ \includegraphics[width=3cm]{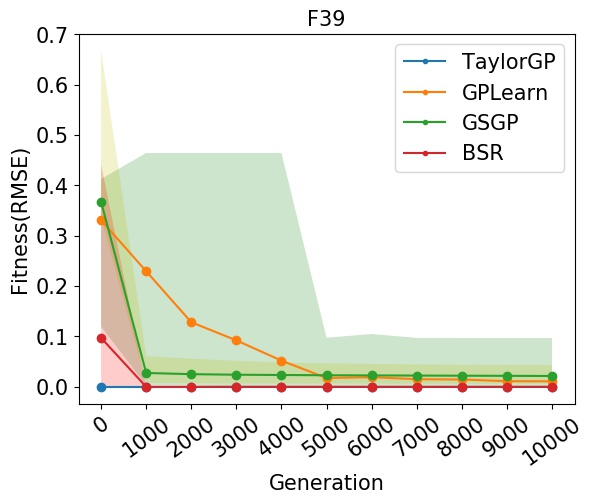}
	}
    \hspace{-0.5cm} \quad
		\subfigure{ \includegraphics[width=3cm]{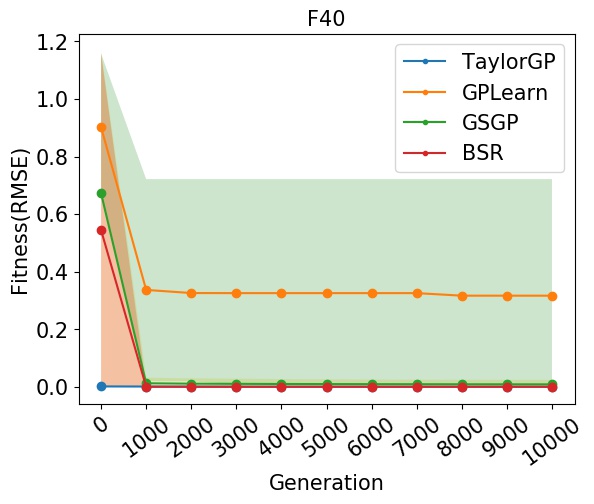}
	}
    \hspace{-0.5cm} \quad
		\subfigure{ \includegraphics[width=3cm]{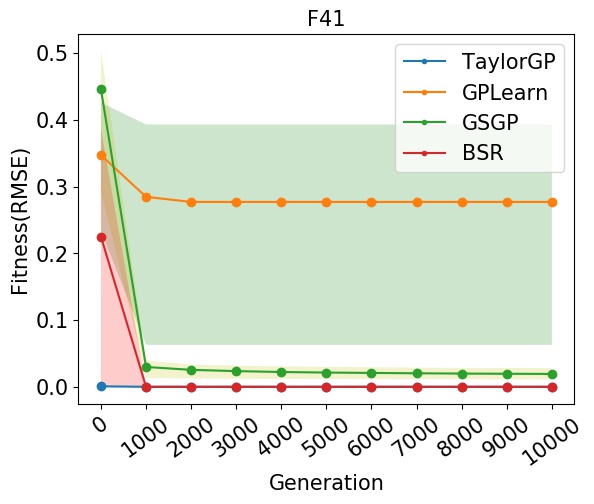}
	}
    \hspace{-0.5cm} \quad
		\subfigure{ \includegraphics[width=3cm]{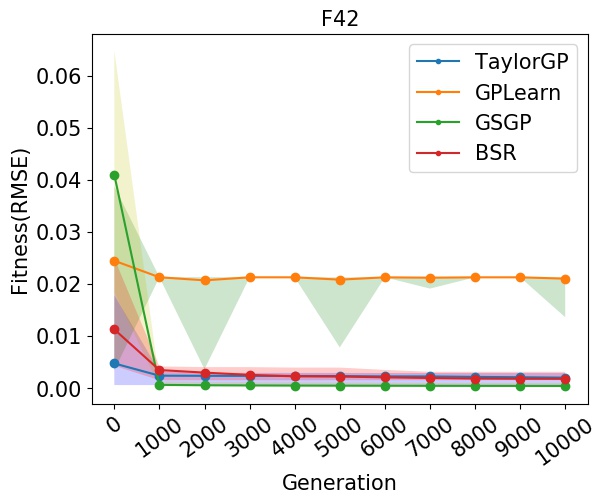}
	}
    \hspace{-0.5cm} \quad
		\subfigure{ \includegraphics[width=3cm]{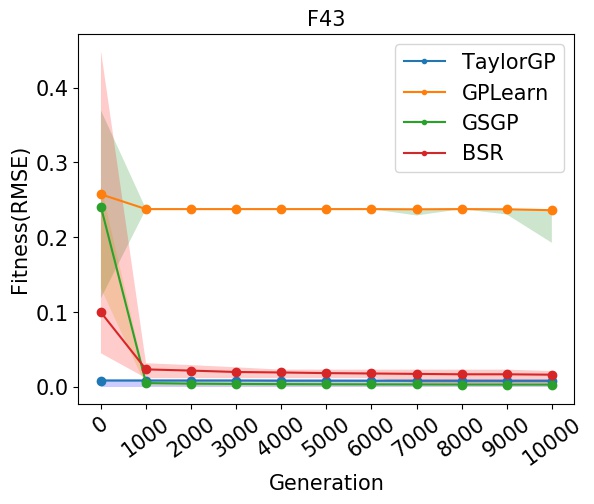}
	}
    \hspace{-0.5cm} \quad
		\subfigure{ \includegraphics[width=3cm]{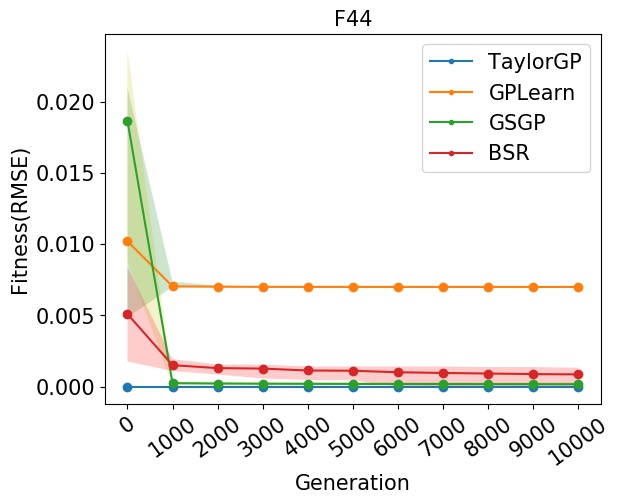}
	}
    \hspace{-0.5cm} \quad
		\subfigure{ \includegraphics[width=3cm]{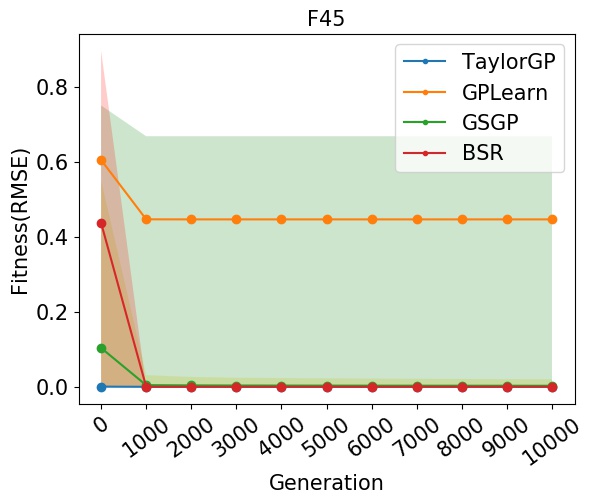}
	}
    \hspace{-0.5cm} \quad
		\subfigure{ \includegraphics[width=3cm]{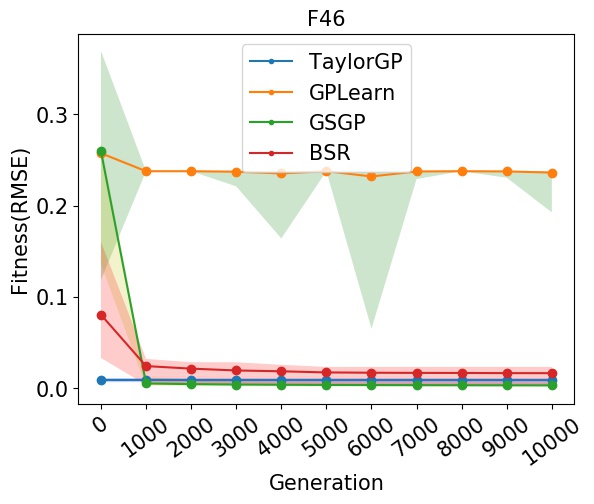}
	}
    \hspace{-0.5cm} \quad
		\subfigure{ \includegraphics[width=3cm]{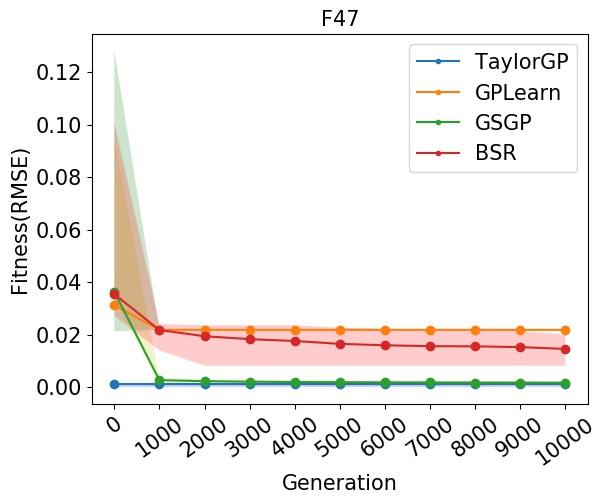}
	}
    \hspace{-0.5cm} \quad
		\subfigure{ \includegraphics[width=3cm]{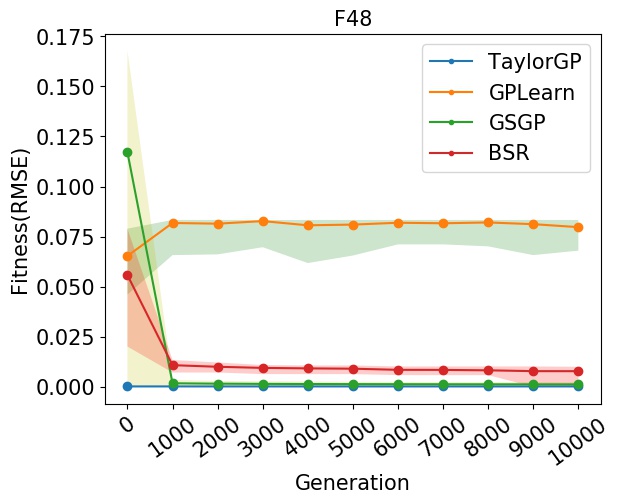}
	}
    \hspace{-0.5cm} \quad
		\subfigure{ \includegraphics[width=3cm]{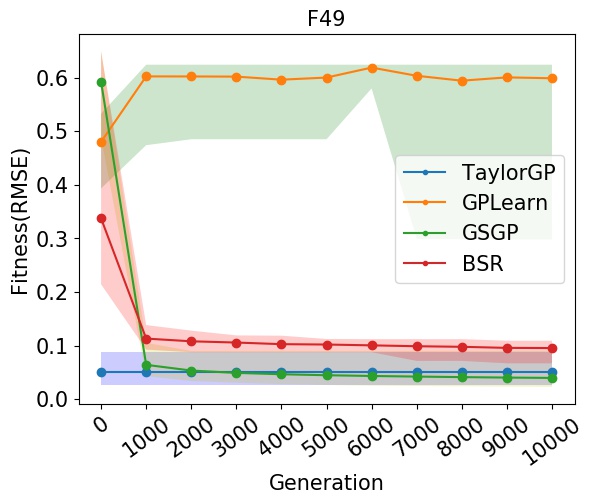}
	}
    \hspace{-0.5cm} \quad
		\subfigure{ \includegraphics[width=3cm]{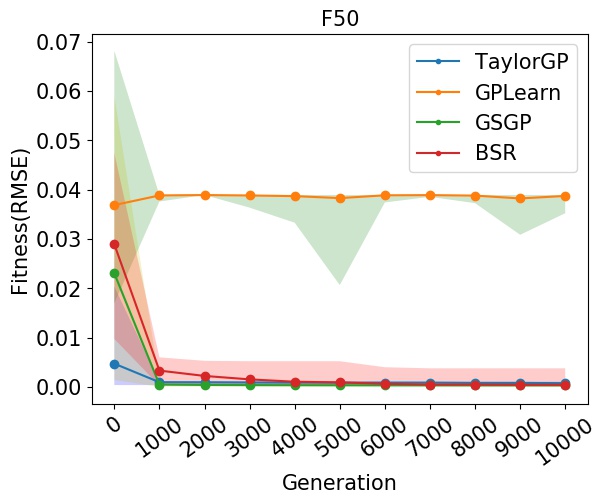}
	}

	    \hspace{-0.5cm} \quad
		\subfigure{ \includegraphics[width=3cm]{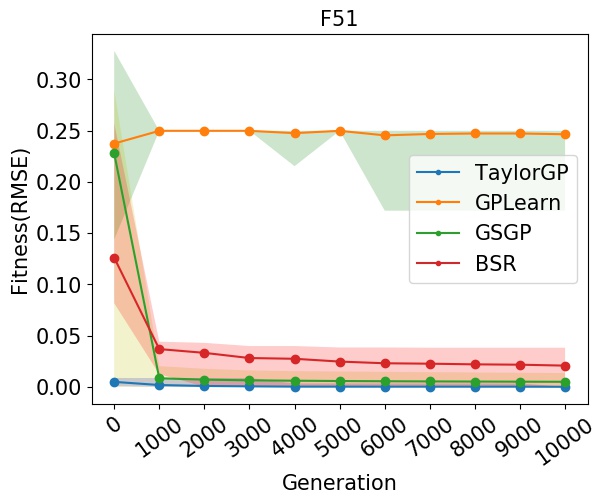}
	}
    \hspace{-0.5cm} \quad
		\subfigure{ \includegraphics[width=3cm]{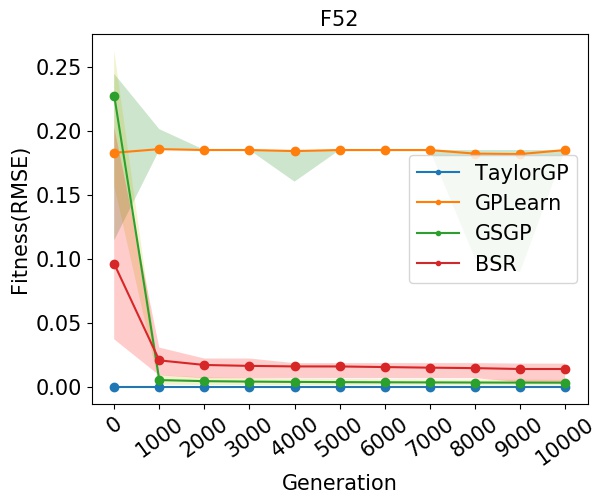}
	}
    \hspace{-0.5cm} \quad
		\subfigure{ \includegraphics[width=3cm]{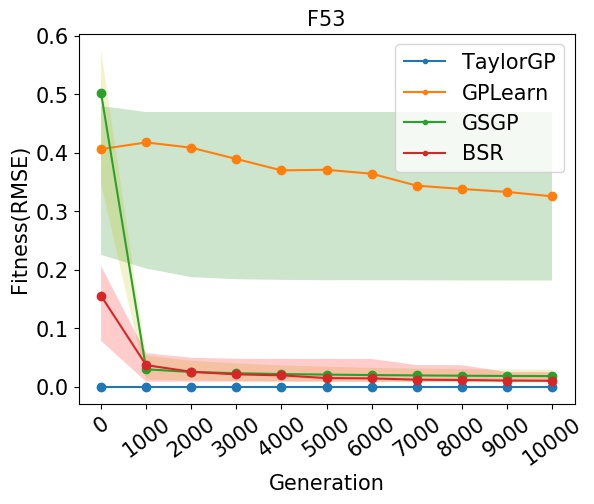}
	}
    \hspace{-0.5cm} \quad
		\subfigure{ \includegraphics[width=3cm]{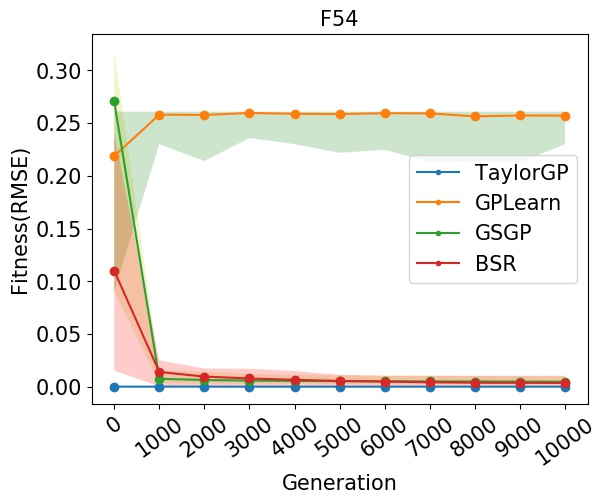}
	}
	    \hspace{-0.5cm} \quad
		\subfigure{ \includegraphics[width=3cm]{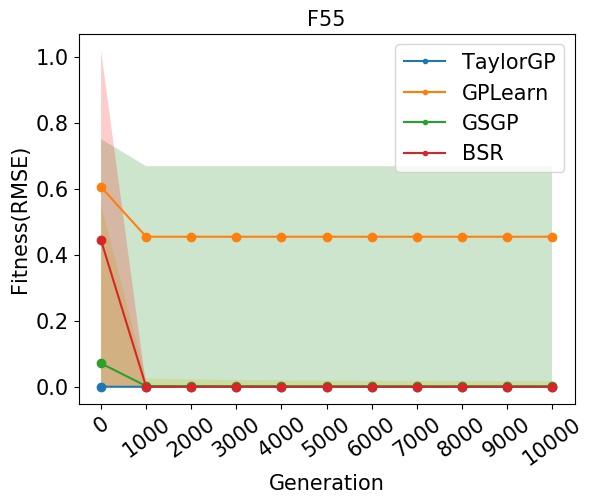}
	}
    \hspace{-0.5cm} \quad
		\subfigure{ \includegraphics[width=3cm]{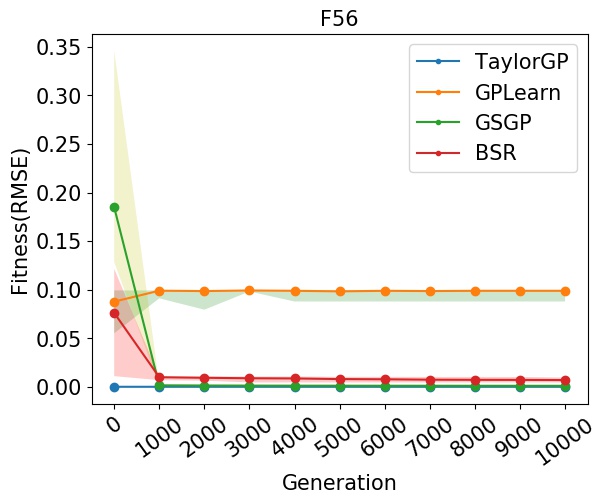}
	}
    \hspace{-0.5cm} \quad
		\subfigure{ \includegraphics[width=3cm]{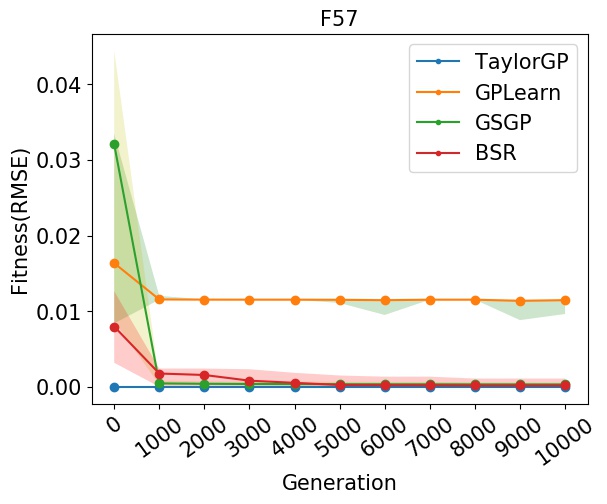}
	}
    \hspace{-0.5cm} \quad
		\subfigure{ \includegraphics[width=3cm]{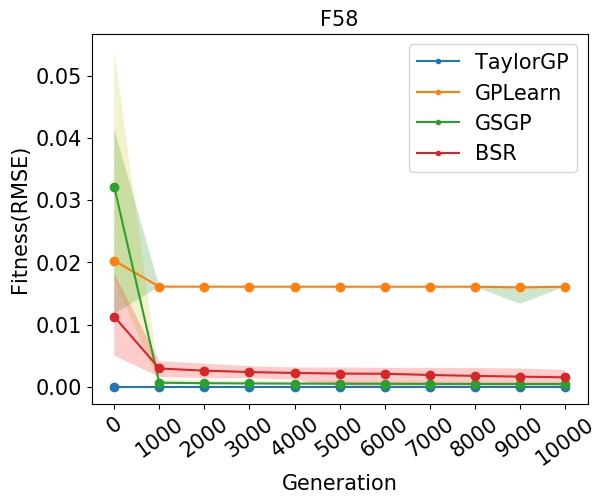}
	}
    \hspace{-0.5cm} \quad
		\subfigure{ \includegraphics[width=3cm]{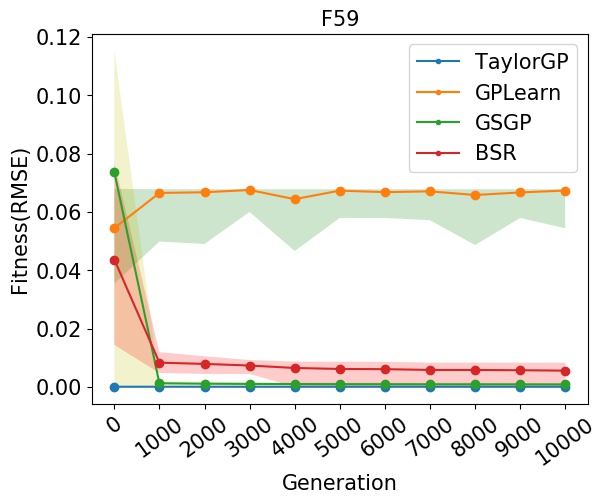}
	}
    \hspace{-0.5cm} \quad
		\subfigure{ \includegraphics[width=3cm]{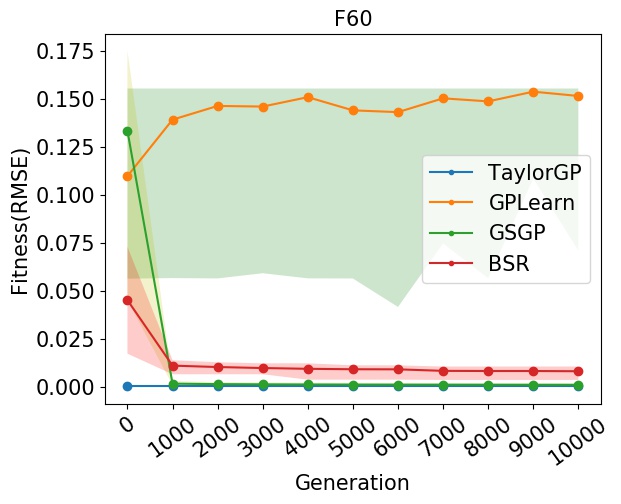}
	}

	\caption{Comparison of convergence. The fitness convergence curve of four algorithms on the benchmarks F31-F60.}
	\label{fig:fitness_converge2}
\end{figure*}

\begin{figure*}[htb]
	\centering

    \hspace{-0.5cm} \quad
		\subfigure{ \includegraphics[width=3cm]{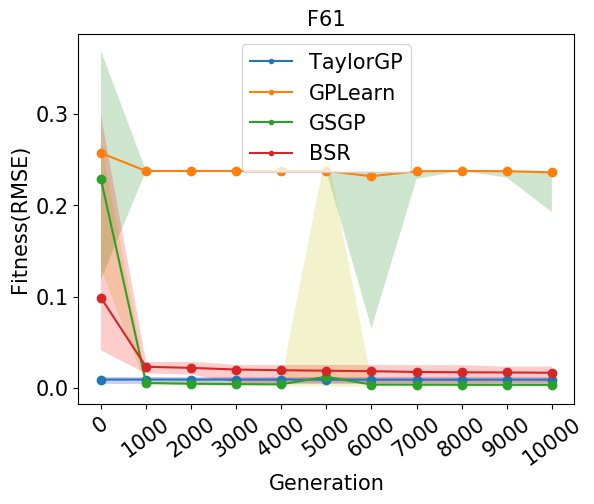}
	}
    \hspace{-0.5cm} \quad
		\subfigure{ \includegraphics[width=3cm]{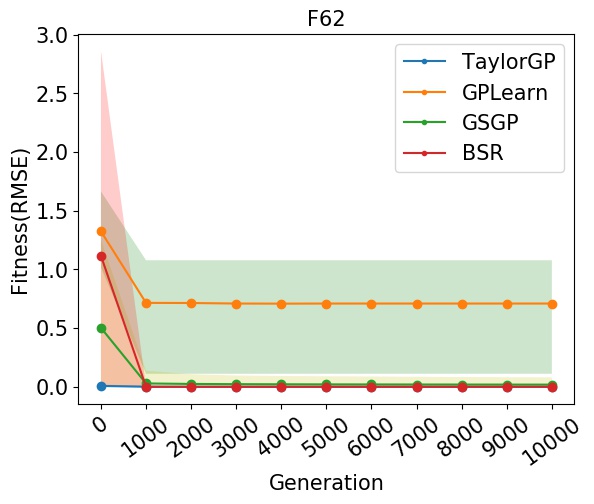}
	}
    \hspace{-0.5cm} \quad
		\subfigure{ \includegraphics[width=3cm]{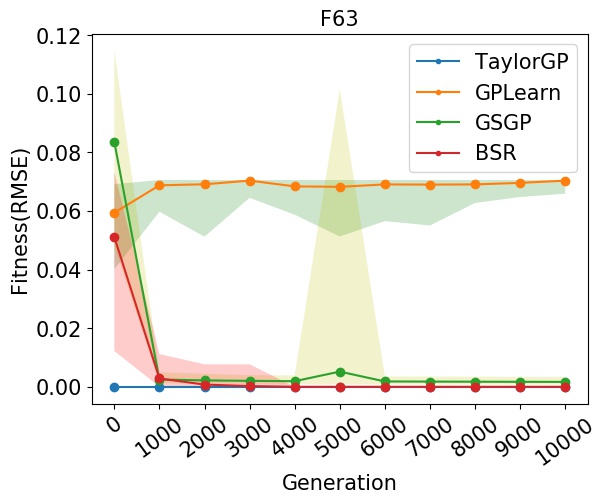}
	}
    \hspace{-0.5cm} \quad
		\subfigure{ \includegraphics[width=3cm]{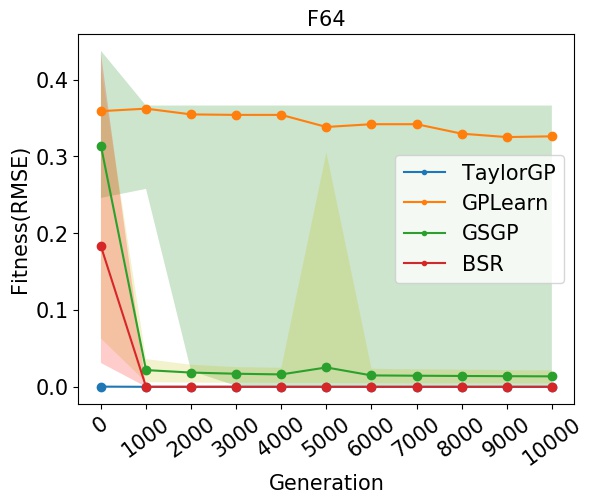}
	}
    \hspace{-0.5cm} \quad
		\subfigure{ \includegraphics[width=3cm]{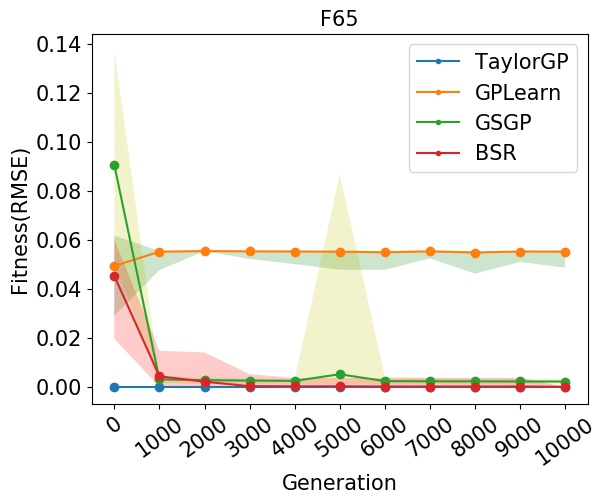}
	}
    \hspace{-0.5cm} \quad
		\subfigure{ \includegraphics[width=3cm]{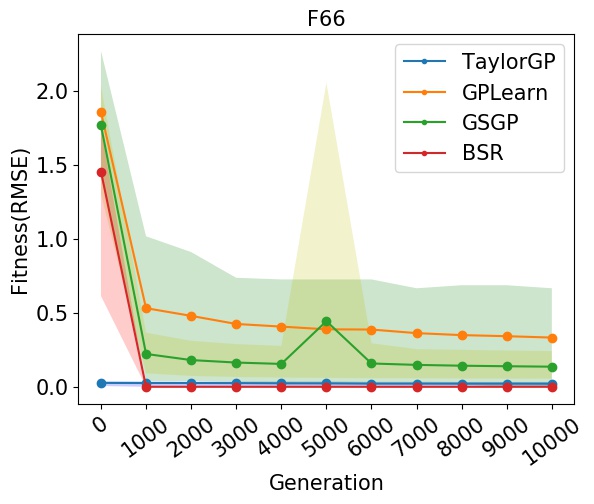}
	}
    \hspace{-0.5cm} \quad
		\subfigure{ \includegraphics[width=3cm]{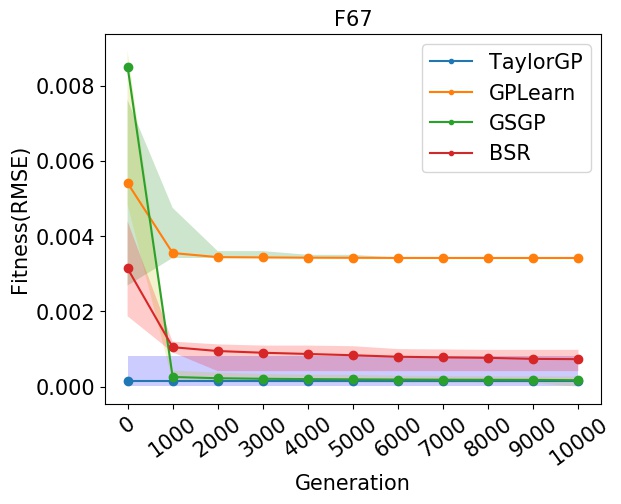}
	}
    \hspace{-0.5cm} \quad
		\subfigure{ \includegraphics[width=3cm]{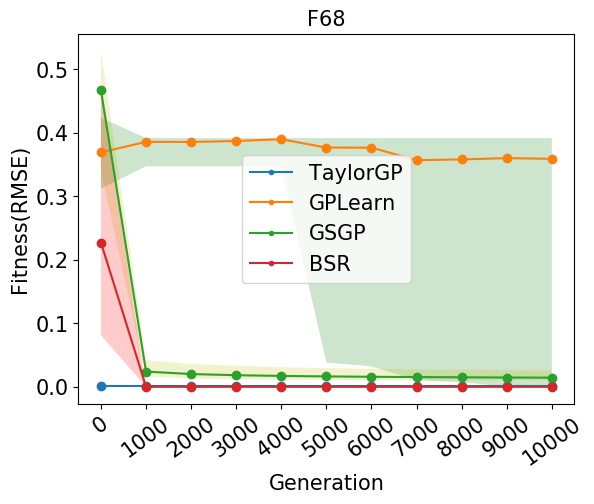}
	}
    \hspace{-0.5cm} \quad
		\subfigure{ \includegraphics[width=3cm]{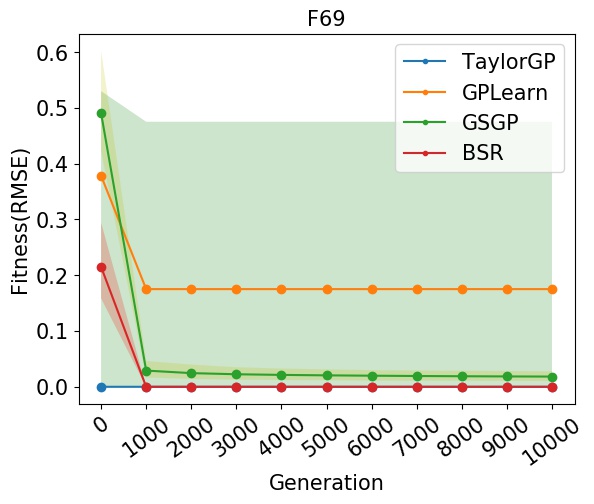}
	}
    \hspace{-0.5cm} \quad
		\subfigure{ \includegraphics[width=3cm]{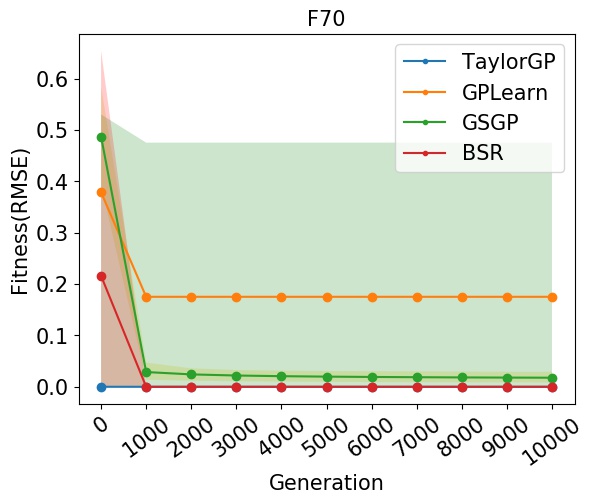}
	}
    \hspace{-0.5cm} \quad
		\subfigure{ \includegraphics[width=3cm]{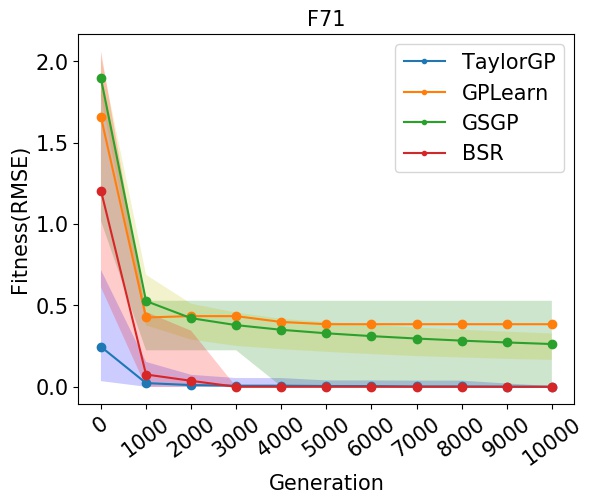}
	}
    \hspace{-0.5cm} \quad
		\subfigure{ \includegraphics[width=3cm]{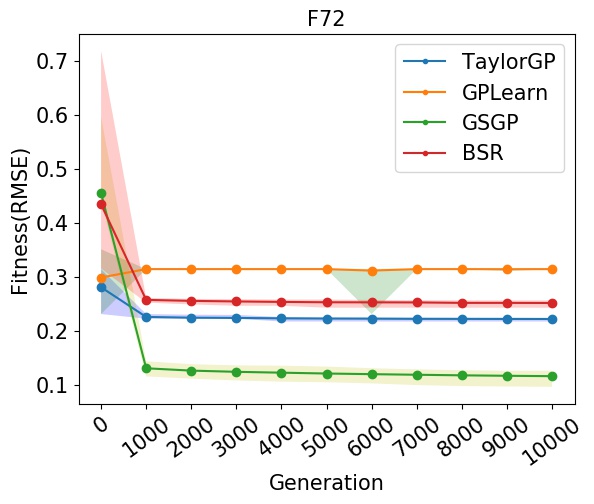}
	}
    \hspace{-0.5cm} \quad
		\subfigure{ \includegraphics[width=3cm]{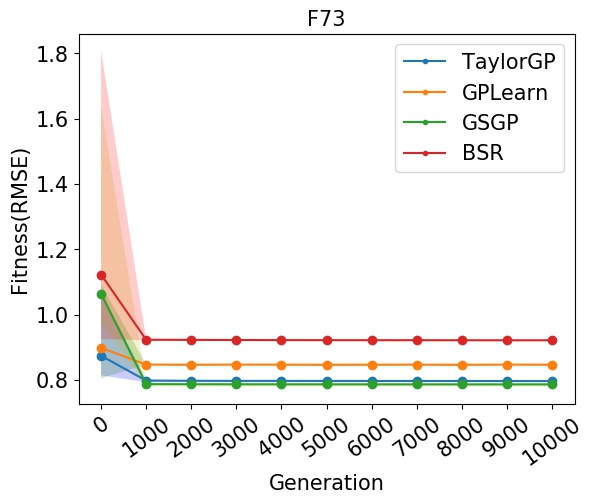}
	}
    \hspace{-0.5cm} \quad
		\subfigure{ \includegraphics[width=3cm]{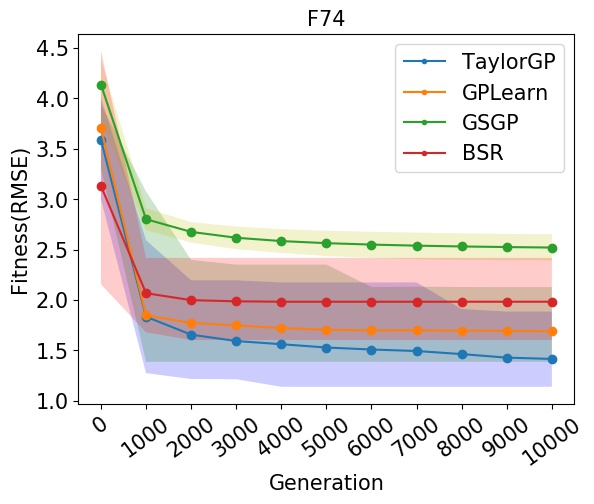}
	}
    \hspace{-0.5cm} \quad
		\subfigure{ \includegraphics[width=3cm]{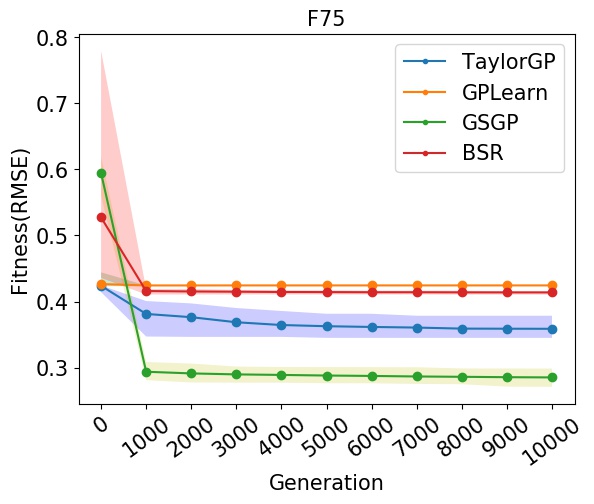}
	}
    \hspace{-0.5cm} \quad
		\subfigure{ \includegraphics[width=3cm]{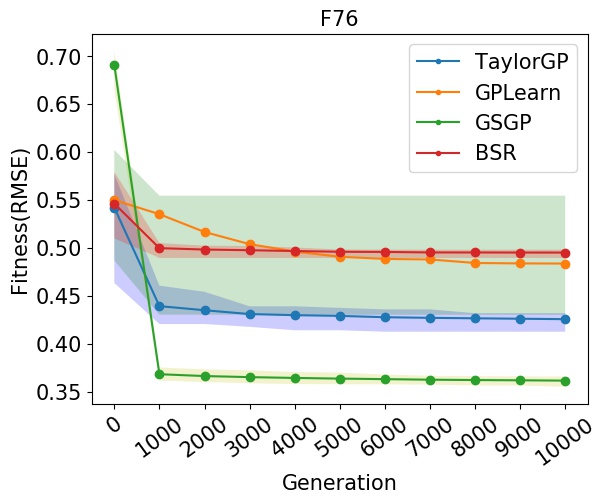}
	}
    \hspace{-0.5cm} \quad
		\subfigure{ \includegraphics[width=3cm]{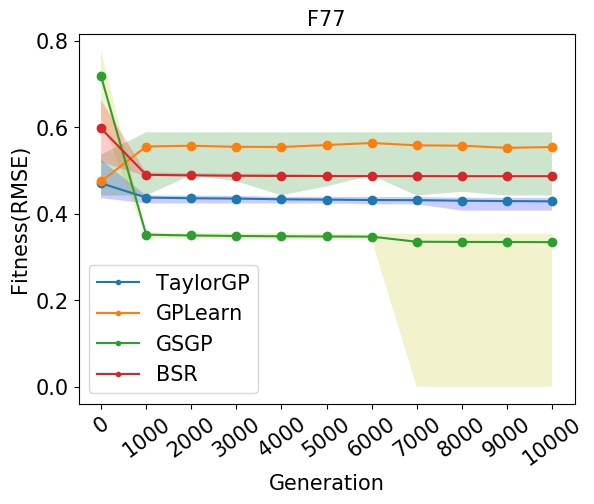}
	}
    \hspace{-0.5cm} \quad
		\subfigure{ \includegraphics[width=3cm]{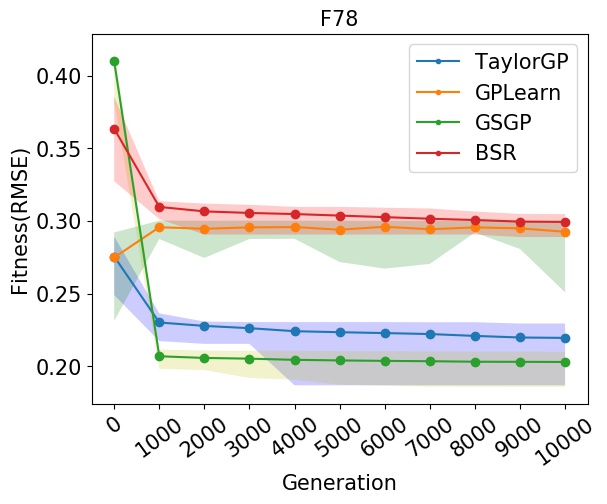}
	}
    \hspace{-0.5cm} \quad
		\subfigure{ \includegraphics[width=3cm]{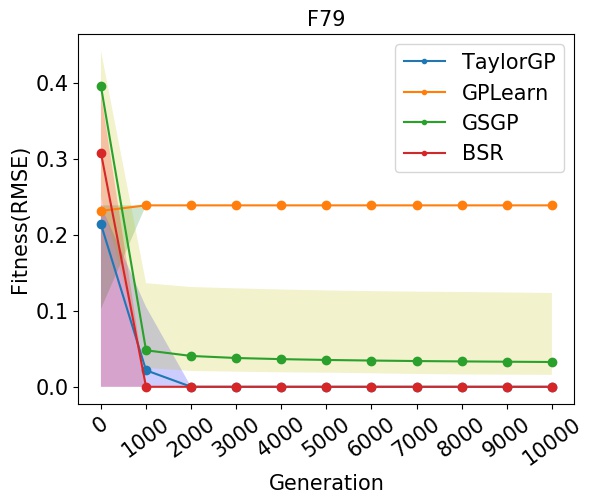}
	}
    \hspace{-0.5cm} \quad
		\subfigure{ \includegraphics[width=3cm]{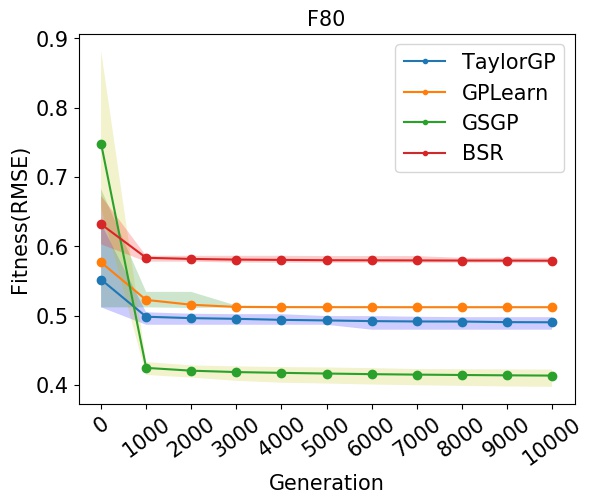}
	}
    \hspace{-0.5cm} \quad
		\subfigure{\includegraphics[width=3cm]{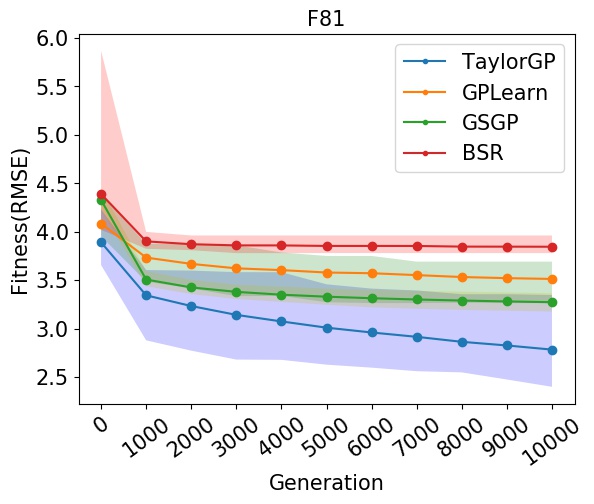}
	}

	\caption{Comparison of convergence. The fitness convergence curve of four algorithms on the benchmarks F61-F81.}
	\label{fig:fitness_converge3}
\end{figure*}